\documentclass{article}

\usepackage[nonatbib,final]{neurips_2022}
\usepackage[numbers,sort&compress]{natbib}

\usepackage[utf8]{inputenc} 
\usepackage[T1]{fontenc}    
\usepackage{hyperref}       
\hypersetup{colorlinks=false, allcolors=black, pdfborder={0 0 0}}
\usepackage{url}            
\usepackage{booktabs}       
\usepackage{nicefrac}      
\usepackage{microtype}     
\usepackage{xcolor}         
\usepackage{amsmath,amssymb,amsthm,bm,bbm,amsfonts,mathtools,thmtools, dsfont} 
\usepackage[capitalize,sort,compress]{cleveref} 
\usepackage{graphicx}
\usepackage{subcaption}
\usepackage{algorithm}
\usepackage[noend]{algorithmic}

\declaretheorem[name=Theorem,numberwithin=section]{theorem}

\newcommand{\norm}[1]{\left\lVert#1\right\rVert}

\title{Fast Bayesian Coresets via Subsampling and Quasi-Newton Refinement}

\author{
   Cian Naik \\
   Department of Statistics \\
   University of Oxford \\
   \texttt{cian.naik@stats.ox.ac.uk}
   \And
   Judith Rousseau \\
   Department of Statistics \\
   University of Oxford \\
   \texttt{judith.rousseau@stats.ox.ac.uk}
   \And
   Trevor Campbell \\
   Department of Statistics \\
   University of British Columbia \\
   \texttt{trevor@stat.ubc.ca}
}

\begin{document}

\maketitle

\begin{abstract}
Bayesian coresets approximate a posterior distribution by building a small
weighted subset of the data points. Any inference procedure that is too
computationally expensive to be run on the full posterior can instead be run
inexpensively on the coreset, with results that approximate those on the full
data. However, current approaches are limited by 
either a significant run-time or the need for the user to specify a low-cost approximation to the 
full posterior. We propose a Bayesian coreset construction algorithm
that first selects a uniformly random subset of data, and then optimizes the weights 
using a novel quasi-Newton method. 
Our algorithm is a simple to implement, black-box method, that
does not require the user to specify a low-cost posterior approximation. 
It is the first to come with a general
high-probability bound on the KL divergence of the output coreset posterior.
Experiments demonstrate that our method 
provides significant improvements in coreset quality 
against alternatives with comparable construction times, with far 
less storage cost and user input required. 

\end{abstract}

\section{Introduction}

Bayesian methods are key tools for parameter estimation and 
uncertainty quantification, but exact inference is rarely 
possible for complex models. Currently, the gold standard
method for approximate inference is 
Markov chain Monte Carlo (MCMC) 
[\citealp{Robert04}; \citealp{Robert11}; \citealp[Ch.~11,12]{Gelman13}], which
involves simulating a Markov chain whose stationary distribution is the 
Bayesian posterior.
However, modern applications are often concerned with very 
large datasets; in this setting, MCMC methods typically have a 
$\Theta(NT)$ complexity---for $T$ samples and dataset size $N$
---which quickly becomes intractable as $N$ increases. 
Motivated by stochastic methods in variational inference \cite{Hoffman13,Ranganath14},
this cost can be reduced by involving
only a random subsample of $M \ll N$ data points in each Markov chain iteration
\cite{bardenet2017markov,Korattikara14,Maclaurin14,Welling11,Ahn12,Quiroz18,bierkens2019zig,pollock2020quasi}.
This reduces the per-iteration computation time of MCMC, but
it can create substantial error in the stationary distribution of the 
resulting Markov chain and cause slow mixing 
\cite{Johndrow20,Nagapetyan17,Betancourt15,Quiroz18,Quiroz19}.

Bayesian coresets \citep{Huggins16,Campbell19JMLR,Campbell18,Campbell19,Manousakas20,Zhang21b} 
provide an alternative for reducing the cost of MCMC and other inference methods. 
The key idea is that in a large-scale data setting, much of the data is often redundant.
In particular, there often exists a \emph{fixed} small, weighted subset of the 
data---a \emph{coreset}---that 
suffices to capture the dataset as a whole in some sense \cite{Campbell19}. 
Thus, if one can find such a coreset, it can be used in place
of the full dataset in the MCMC algorithm, providing the 
simplicity, generality, and per-iteration speed of data-subsampled MCMC
without the statistical drawbacks.
Current state-of-the-art Bayesian coreset approaches 
based on sparse variational inference
\citep{Campbell19} empirically provide high-quality coresets,
but are limited by their significant 
run-time and lack of theoretical convergence guarantees. 
Methods based on sparse regression \citep{Campbell19JMLR,Campbell18,Zhang21b} are 
significantly faster and come with some limited theoretical guarantees,
but require a low-cost approximation to the full 
posterior. This approximation is often 
impractical to find and difficult to tune. Moreover,
it can fundamentally limit the quality of the coreset
obtained.
Aside from \citep{Campbell19}, these methods also require $\Theta(N)$ storage, which is problematic in the large-data setting.
There is a third class of constructions based on importance sampling \cite{Huggins16},
but these do not provide reliable posterior approximations in practice.

In this work, we develop a novel coreset construction algorithm
that is both faster and easier to tune than the current state of the art,
and has theoretical guarantees on the quality of its output.
The key insight in our work is that we can split the construction
of a coreset into two stages: first we select the data in the coreset
via uniform subsampling---an idea developed 
concurrently by \citep{jankowiak2021surrogate,chen2022bayesian}---and then
optimize the weights using a few steps of a novel quasi-Newton method.
Selecting points uniformly randomly avoids the slow inner-outer loop
optimization of \citep{Campbell19}, and guarantees that
the optimally-weighted coreset has a low KL divergence to the posterior (\cref{thm:optimal_kl:finitd,thm:optimal_kl}). 
Weighting these uniformly selected points correctly is crucial, and our quasi-Newton method is guaranteed to converge to a point close
to the optimally-weighted coreset (\cref{thm:convergence}) in significantly fewer iterations than 
the method of \citep{Campbell19}.
Finally, our method is easy to tune and does not require the low-cost posterior approximation
of \citep{Campbell19JMLR,Campbell18}. 
Our experiments show
that the algorithm exhibits significant improvements in coreset quality 
against alternatives with comparable construction times.
\section{Bayesian coresets}
The problem we study is as follows. Suppose we are given a target probability
density $\pi(\theta)$ for variable $\theta \in \Theta$ that is comprised of
$N$ potentials $\left(f_n(\theta)\right)_{n=1}^N$ and base density
$\pi_0(\theta)$,
\begin{align}
    \pi(\theta) \coloneqq \frac{1}{Z}\exp\left(\sum_{n=1}^N f_n(\theta)\right)\pi_0(\theta),
\end{align}
where $Z$ is the normalizing constant. 
This setup corresponds to a Bayesian statistical model with prior $\pi_0$
and i.i.d.~data $X_n$ conditioned on $\theta$, where $f_n(\theta) = \log p(X_n | \theta)$.
Computing expectations under $\pi$ exactly is often intractable. 
MCMC methods can be employed here, but
typically have a computational complexity of $\Theta(NT)$ to obtain $T$
samples, since $\sum_n f_n(\theta)$ needs to be evaluated at each step. Bayesian coresets \cite{Huggins16}
offer an alternative, by constructing a small, weighted subset of the
dataset, on which any MCMC algorithm can then be run.

To do this, we find a set of weights $w \in \mathbb{R}_{\geq 0}^N$
corresponding to each data point, with the constraint that only $M \ll N$ of
these are nonzero, i.e. 
$\norm{w}_0 \coloneqq \sum_{n=1}^N  \mathds{1}_{w_n > 0} \leq M.$
The weighted subset of points corresponding to the strictly positive weights 
is called a \textit{coreset}, and we can then create the coreset posterior approximation,
\begin{align}
    \pi_{w}(\theta) \coloneqq \frac{1}{Z(w)} \exp\left(\sum_{n=1}^N w_n f_n(\theta)\right)\pi_0(\theta),
\end{align}
where $Z(w)$ is the new normalizing constant, and $\pi_1 = \pi$ corresponds to
the full posterior. Running MCMC on this approximation has
complexity $\Theta(MT)$, a considerable speedup if $M \ll N$.

Constructing a sparse set of weights $w$ so that $\pi_w$ is as close to $\pi$ as possible
is the key challenge here. 
Methods based on importance sampling \cite{Huggins16} and sparse regression
\cite{Campbell19JMLR,Campbell18,Zhang21b} have been developed, but these provide poor
posterior approximations or require first finding a low-cost posterior
approximation.  The current state-of-the-art approach resolves both of
the aforementioned issues by formulating the coreset construction
problem as variational inference in the family of coresets \cite{Campbell19}:
\begin{align}\label{eq:sparse_vi_coreset}
    w^{\star}=\underset{w \in \mathbb{R}^{N}}{\arg \min } \quad \mathrm{D}_{\mathrm{KL}}\left(\pi_{w} \| \pi_{1}\right) \quad
    \text {s.t.} \quad w \geq 0,\|w\|_{0} \leq M ,
\end{align}
where $\mathrm{D}_{\mathrm{KL}}(\cdot\|\cdot)$ denotes the KL divergence. Noting that coresets are a
sparse subset of an exponential family---the weights form the
natural parameter $w \in \mathbb{R}_{\geq 0}^{N}$, the component potentials
$\left(f_{n}(\theta)\right)_{n=1}^{N}$ form the sufficient statistic, $\log
Z(w)$ is the log partition function, and $\pi_{0}$ is the base density,
\begin{align}
    \pi_{w}(\theta):=&\exp \left(w^{T} f(\theta)-\log Z(w)\right) \pi_{0}(\theta), \quad f(\theta):=(f_1(\theta), \dots, f_N(\theta))^T \in\mathbb{R}^N,
\end{align}
one can obtain a formula for the gradient
\begin{align}
\nabla_{w} \mathrm{D}_{\mathrm{KL}}\left(\pi_{w} \| \pi_{1}\right)=&-\nabla_{w}^{2} \log Z(w)(1-w) =-\operatorname{Cov}_{w}\left[f, f^{T}(1-w)\right], \label{eq:klgrad}
\end{align}
where $\operatorname{Cov}_{w}$ denotes covariance under $\pi_{w}$. 
Here, and throughout, we take derivatives with
respect to $w$ on the ambient space 
$w \in \mathbb{R}^N$.
The method of \citep{Campbell19} then involves
selecting a data point to add to the coreset,
optimizing the coreset weights using stochastic gradient estimates based on \cref{eq:klgrad}, 
and then iterating. In practice, this method is infeasibly slow outside of small-data problems. 
For every data point selected it requires sampling from $\pi_w$
a number of times equal to the number of optimization steps (set to $100$ in \citep{Campbell19}). 
This can easily lead to tens of thousands of sampling steps in total, each of which typically necessitates the use of an MCMC algorithm.
Conversely, our method requires tens of sampling steps in total.

\section{Construction via subsampling and quasi-Newton refinement}\label{sec:coreset}
In this section, we provide a new Bayesian coreset construction algorithm
(Algorithm \ref{alg:anc_coreset})
to solve the sparse variational inference problem \cref{eq:sparse_vi_coreset}. 
This method involves first uniformly subsampling the data,
and then optimizing the weights on the subsample. Here, we give a
detailed explanation of how the algorithm works.
We defer to \cref{sec:theory} the theoretical analysis of the method, which 
demonstrates that, with high probability, it finds a near-optimal coreset 
in a small number of optimization iterations.

\subsection{Uniform subsampling}\label{sec:subsample}
The first key insight to our approach is that we can select the $M$ data points
that comprise the coreset by simple uniform subsampling.
From an algorithmic standpoint, the benefit of this approach is clear: it
makes the selection of coreset points fast and easy to implement.
It also decouples the subset
selection from the optimizing of the coreset weights, leaving us with a
simpler optimization problem to solve.\footnote{The benefits of this
initial uniformly random selection step 
are also noted by \citep{jankowiak2021surrogate,chen2022bayesian} in concurrent work.}
Intuitively, this is reasonable from a coreset quality standpoint 
because a uniform subsample of the data will, with high
probability, create a highly-expressive ``basis'' of log-likelihood functions
$\left(f_n\right)_{n=1}^M$ with which to approximate the full data log-likelihood. 
Of course, we still need to optimize the coreset weights; but we have 
not limited the flexibility of the coreset family significantly by 
selecting points via uniform subsampling.
\cref{thm:optimal_kl} provides a precise statement of these
ideas, and shows that the optimal weighted coreset posterior built 
using a uniformly random subset of data is a good posterior approximation.

Since the coreset data points are chosen uniformly randomly,
we will w.l.o.g.~re-label the selected subset to have indices $1, \dots, M$, 
and the remaining data points to have indices $M+1, \dots, N$ from now on.
We can specify a new weight constraint set
$\mathcal{W} \subset\mathbb{R}^N$ of vectors $w$ with 
0 entries for every index beyond $M$, and a further
subset $\mathcal{W}_N$ where the weights sum to $N$,
\begin{align*}
\mathcal{W} = \left\{w \in \mathbb{R}^N : w\geq 0, \,\, n > M \implies w_n = 0\right\},\qquad
\mathcal{W}_N = \left\{w\in\mathcal{W} : 1^Tw = N\right\}.
\end{align*}
It is also useful to denote $g(\theta)= (f_1(\theta), \dots, f_M(\theta))^T \in \mathbb{R}^M$ as
the first $M$ components of $f(\theta)$.
\subsection{Quasi-Newton optimization}\label{sec:quasinewton}
Given the choice of $M$ data points to include in the coreset, 
it is crucial to optimize their weights correctly. This is the second
step of our algorithm.
The KL divergence gradient for the $M$ weights is
\begin{align}\label{eq:grad_kl}
\nabla_{w} \mathrm{D}_{\mathrm{KL}}\left(\pi_{w} \| \pi_{1}\right)=&\left(-\nabla_{w}^{2} \log Z(w)(1-w)\right)_{1:M} =-\operatorname{Cov}_{w}\left[g, f^{T}(1-w)\right].
\end{align}
To reduce the number of weight optimization iterations
---each of which generally involves
MCMC sampling from $\pi_w$, where $w$ is the current set of weights---we 
develop a second-order optimization method.
Again using the properties of exponential families, we can derive the Hessian of the KL divergence
with respect to the $M$ weights
\begin{align}
\nabla^2_{w} \mathrm{D}_{\mathrm{KL}}\left(\pi_{w} \| \pi_{1}\right) = \left(\nabla_{w}^{2} \log Z(w) \!-\! \nabla^3_{w} \log Z(w) (1\!-\!w)\right)_{1:M,1:M}.\notag
\end{align}
This matrix is not guaranteed to be positive (semi-)definite. However,
writing out the second term explicitly as
\begin{align*}
    \nabla_{w}^{3} \log Z(w)(1-w) = \mathbb{E}_{\pi_w} \left[ \left( f-\mathbb{E} _{ \pi_w } (f) \right) \left( f-\mathbb{E} _{ \pi_w } (f) \right)^T \left( (f^T (1-w) -\mathbb{E} _{ \pi_w } (f^T (1-w) ) \right) \right],
\end{align*}
we see that this term will be small near the optimum, 
where the coreset approximation is ideally good, i.e., $w^Tf(\theta) \approx 1^Tf(\theta)$.
The first term does not contain this $f^T (1-w)$ term,
and thus should dominate the expression. 

This motivates the use of
\begin{align}
\nabla^2_{w} \mathrm{D}_{\mathrm{KL}}\left(\pi_{w} \| \pi_{1}\right) \approx\left(\nabla_{w}^{2} \log Z(w)\right)_{1:M,1:M} =\operatorname{Cov}_{w}\left[g, g\right] \label{eq:quasi_hess_kl}
\end{align}
to scale gradient steps in the optimization method rather than the true Hessian, thus creating a quasi-Newton method \cite{Broyden72}.
This heuristic motivates our approach, but it is rigorously justified by
\cref{thm:convergence}, which proves that this scaling
results in optimization iterations that are guaranteed to converge exponentially
to a coreset which is nearly globally optimal.
Note that when $M$ is larger than the inherent dimension of the space
of log-likelihood functions, $\operatorname{Cov}_{w}\left[g, g\right]$ will have zero
eigenvalues and be noninvertible. Therefore we add a regularization $\tau > 0$
prior to inversion. \cref{thm:convergence} shows how the regularization influences the optimization,
but in general we want $\tau$ to be as small as possible while still ensuring the numerical
stability of inverting $\operatorname{Cov}_{w}\left[g, g\right]$.

Our optimization method is as follows. First, we initialize
at the uniformly weighted coreset, i.e.,
\begin{align*}
w_0\in\mathcal{W}, \quad w_{0m} = \frac{N}{M}, \quad m=1,\dots, M.
\end{align*}
Next, given a step-size tuning sequence $\gamma_k$, $k\geq 0$, 
we update the weights at each iteration $k\in\mathbb{N}\cup\{0\}$
using the $(\tau > 0)$-regularized quasi-Newton step
\begin{align}
\hat w_{k+1} &= w_k + \gamma_k (G(w_k) + \tau I)^{-1} H(w_k)(1-w_k)\label{eq:qnc_step}\\
G(w) &= \mathrm{Cov}_w\left[g, g\right], \quad H(w) = \mathrm{Cov}_w\left[g, f\right].\label{eq:G_H_def}
\end{align}
\cref{thm:convergence} suggests that the step size $\gamma_k$ should be constant:
$\gamma_k = \gamma \in (0, 1]$; but the analysis in that section assumes
that we can compute $G(w)$ and $H(w)$ exactly, whereas in practice we 
will have to estimate them. Hence we allow $\gamma_k$ to depend 
on the iteration number in general.
Finally, we project $\hat w_{k+1}$ back onto the constraint set to obtain 
next iterate $w_{k+1}$, i.e. for $m=1,\ldots, M$ set
\begin{align}\label{eq:qnc_projection}
w_{k+1,m} =\max\left( \hat w_{k+1,m}, 0\right).
\end{align}
\subsection{Algorithm}\label{sec:algorithm}
The pseudocode for the quasi-Newton coreset 
construction method is shown in \cref{alg:anc_coreset}.
There are a number of practical considerations needed to use this method;
we discuss these here.

The first consideration is that we cannot
calculate $G(w)$ and $H(w)$ in closed form. 
Instead, at step $k$ of the algorithm, we use a Monte
Carlo estimate; first taking $S$ samples
$\left(\theta_{s}\right)_{s=1}^{S} \stackrel{\text { i.i.d. }}{\sim}
\pi_{w_k}$, and calculating 
$\hat{g}_{s}:=g(\theta_s) - \frac{1}{S}\sum_{r=1}^S g(\theta_r)$,
so that we may estimate $G(w_k)$ as
\begin{align}
\hat{G}_k &= \frac{1}{S} \sum_{s=1}^{S} \hat{g}_{s} \hat{g}_{s}^{T} \,\,\in \mathbb{R}^{M\times M}.
\end{align}
Using a Monte Carlo estimate in this way introduces 
one source of error. Furthermore,
we often cannot sample exactly
from $\pi_{w_k}$, and instead use MCMC to do so. This
could lead to additional errors, for example if our samples are from
unconverged chains. However, we find that this method
works well in practice, for a reasonable number of samples $S$.

As $H(w_k)$ is an $M\times N$ matrix, estimating it directly will incur a
 $\Theta(MN)$ storage cost. Instead, we note 
that we only require 
$H(w_k)(1-w_k) = \mathrm{Cov}_{w_k}\left[g, f^T(1-w_k)\right] = \mathrm{Cov}_{w_k}\left[g, h\right]$, 
where $h \coloneqq f^T1 -g^Tw_k = \bar{f} - g^Tw_k$. We can estimate $h$ using 
$\hat{h}_s = \sum_{n=1}^N\left[f_n(\theta_s)- \frac{1}{S}\sum_{r=1}^S f_n(\theta_r)\right] - \hat{g}_s^Tw_k$,
which requires $O(N)$ time, but only $O(S)$ space. We can then 
estimate $H(w_k)(1-w_k)$ as
\begin{align}
\hat{H}_k(1-w_k) &= \frac{1}{S} \sum_{s=1}^{S} \hat{g}_{s} \hat{h}_s.
\end{align}
Thus, we can find $w_{k+1}$ from $w_{k}$ by taking the stochastic Newton step
\begin{align}
\hat{w}_{k+1} = w_k + \gamma_k \left(\hat{G}_k + \tau I \right)^{-1}\hat{H}_k(1-w_k),
\end{align}
and projecting onto the constraint set: 
$w_{k+1,m} = \hat{w}_{k+1,m}\mathds{1}_{\hat{w}_{k+1,m}\geq 0}$. 
We set the regularization parameter $\tau$ by examining the condition number of 
$\hat{G}_k + \tau I$ and keeping it below a reasonable value. 
We can tune $\gamma_k$ using a line search method. As we do not have access to 
the objective function (i.e. the KL-divergence between the coreset and full 
posteriors), we use the curvature part of the Wolfe conditions \citep{wolfe1969convergence}
to tune this. In practice, this line search is expensive. 
Thus, we only tune $\gamma_k$ for $k \leq K_{\text{tune}}$, 
and leave it as a constant thereafter. The intuition here is that we may start with
quite poor coreset weights, and so need to choose the initial steps carefully.
In \cref{app:experiments}, we perform a 
sensitivity analysis for the parameters $S$, $K_{tune}$ and $\tau$.
We see that our results are generally not sensitive to the choice of these parameters, within reasonable ranges.
\begin{algorithm}[t!]
\caption{\sc{QNC (Quasi-Newton Coreset)}}\label{alg:anc_coreset}
\begin{algorithmic}[1]
\REQUIRE $(f_n(\theta))_{n=1}^{N}$ $\pi_0$, $S$, $K$, $K_{tune}$, $\gamma$, $\tau$, $M$.
\STATE Uniformly sample $M$ data points, w.l.o.g. these correspond to indices $1,\ldots,M$.
\STATE Set $w_{0m} = \frac{N}{M}, \quad m=1,\dots, M.$
\FOR{$k=0, \ldots, K-1$}
    \STATE Sample $\left(\theta_{s}\right)_{s=1}^{S} \stackrel{\text { i.i.d. }}{\sim} \pi_{w_k} \propto \exp \left(w_k^{T} f(\theta)\right) \pi_{0}(\theta)$ \label{alg:step:sample_theta}
    \STATE Set $\hat{g}_{s} \leftarrow g(\theta_s) - \frac{1}{S}\sum_{r=1}^S g(\theta_r)$ and $\hat{h}_s \leftarrow \sum_{n=1}^N\left[f_n(\theta_s)- \frac{1}{S}\sum_{r=1}^S f_n(\theta_r)\right] - \hat{g}_s^Tw_k$
    \STATE Set $\hat{H}_k(1-w_k) \leftarrow \frac{1}{S} \sum_{s=1}^{S} \hat{g}_{s} \hat{h}_s$ and $\hat{G}_k \leftarrow \frac{1}{S} \sum_{s=1}^{S} \hat{g}_{s} \hat{g}_{s}^{T}$
    \IF{$k \leq K_\text{tune}$}
        \STATE Choose $\gamma_k$ via line search with starting value $\gamma$
      \ELSE
        \STATE Set $\gamma_k \leftarrow \gamma$
    \ENDIF
    \STATE Take a quasi-Newton step: $\hat{w}_{k+1} \leftarrow w_k + \gamma_k \left(\hat{G}_k + \tau I \right)^{-1}\hat{H}_k(1-w_k)$
    \STATE Project: for all $m\in[M]$, $w_{k+1,m} \leftarrow \max\left( \hat w_{k+1,m}, 0\right) $
\ENDFOR
\STATE \algorithmicreturn $w=w_K$ 
\end{algorithmic}
\end{algorithm}

Computing $\hat{G}_k + \tau I$ involves taking $S$ samples, forming the 
product $\hat{g}_{s} \hat{g}_{s}^{T}$, and then inverting the resulting matrix.
Its time complexity is thus $O(SM^2 + M^3)$. 
Computing $\hat{H}(w_k)(1-w_k)$ similarly has
complexity $O\left(SM + SN\right)$, and computing the product with 
$(\hat{G}_k + \tau I)^{-1}$ has complexity
$O\left(M^2\right)$. For $K$ Newton steps in Algorithm
\ref{alg:anc_coreset}, the overall time complexity is thus $O(K(M^3 +SM^2
+ SN))$, which is linear in $N$. The space complexity is 
$O(M^2)$ (to store $\hat{G}_k$), which is sublinear in $N$, unlike in 
\citep{Campbell18,Zhang21b}.

In theory, we can reduce the time complexity by using a further stochastic estimate
of $\hat H_k(1-w_k)$. In particular, we can instead
calculate $\hat{h}_{s}$ only for indices in 
$\mathcal{I} \coloneqq \{1,\ldots,M\} \cup \mathcal{T}$, where $\mathcal{T}$ 
is a uniformly selected sample $\mathcal{T} \subseteq \{1,\ldots,N\}$. 
Calculating $\hat{H}(w_k)(1-w_k)$ using this subsampled vector 
$\hat{h}_{s,\mathcal{I}}$ then has complexity $O(SM+ST)$ in the worst
case, where $T \coloneqq |\mathcal{T}|$. The overall time complexity 
is then $O(K(M^3 +SM^2 + ST))$, which is sublinear 
in $N$ if $T = o(N)$. This could give significant 
improvements in coreset construction time in the large-data
regime. However, this approach 
requires further study, as we find that it leads
to a degradation in the performance of our
algorithm. We do not use it in our experiments.

The final practical consideration in the design of
\cref{alg:anc_coreset} is that we cannot actually
calculate the objective function that we are trying to
minimize, namely the KL divergence
$\mathrm{D}_{\mathrm{KL}}\left(\pi_{w} \| \pi_{1}\right)$ 
between the coreset and full posteriors. Thus, it may
be hard to tell if the optimization is actually making
progress. In practice, we monitor the norm of the
gradient of the KL divergence (which we do have access
to, as given by \cref{eq:grad_kl}). We can terminate
our algorithm early if we do not see a significant
enough decrease in this measure.

\section{Theoretical analysis}\label{sec:theory}

In this section we provide a theoretical analysis of the proposed method.  We
demonstrate 
(in \cref{thm:optimal_kl:finitd,thm:optimal_kl}) that the optimal coreset posterior built
using a uniformly random subset of data is, with high probability, an exact or near-exact
approximation of the true posterior.  
Second, we show (in \cref{thm:convergence})
that the proposed optimization algorithm converges exponentially quickly to a
point near the optimal coreset, thus realizing the guarantee in
\cref{thm:optimal_kl} in practice, and theoretically justifying our proposed algorithm. 
The assumptions required for our theory are somewhat technical.
We give some intuition here on settings where they hold, but discuss this 
further in \cref{app:theory}.
Proofs of all results may be found in \cref{sec:proofs}. 

\subsection{Subset selection via uniform subsampling}\label{sec:theory-sel}
\cref{thm:optimal_kl:finitd} and \cref{thm:optimal_kl} provide a theoretical 
foundation for selection via
uniform subsampling. Both recommend setting $M \gtrsim d\log N$, 
where $d$ is essentially the ``dimension'' of the span of the log-likelihood functions.
\cref{thm:optimal_kl:finitd} states that 
if the span of the log-likelihood functions is indeed finite-dimensional with dimension $d$---e.g., in a usual
exponential family model---then with high probability, the optimal coreset
using a uniformly random selection of $M$ points is exact.
\cref{thm:optimal_kl} extends this
to the general setting where $\Theta\subseteq \mathbb{R}^D$, $N$ is large relative to $D$,
and where we assume the posterior concentrates as $N\to\infty$.
In particular, \cref{thm:optimal_kl} 
states that with high probability, 
the optimal coreset using a uniformly
random selection of $M$ data points provides low error.

In what follows, we assume that $f_n(\cdot) $ are i.i.d. 
from some true data-generating distribution  $p$, and 
we define $\bar f(\cdot)  = \mathbb{E}_p( f_n(\cdot))$.  
We also introduce the new notation $f(\cdot,\cdot)$, where we consider
the potentials $f$ as functions of both $x$ and $\theta$.
This enables us to make the data-dependence explicit when needed, i.e., $f_n(\cdot) = f(x_n,\cdot)$;
for instance, in the standard context of Bayesian 
inference $f_n(\cdot) = f(x_n, \cdot) =  \log p(x_n | \cdot)$, 
with $x_n \overset{\text{i.i.d.}}{\sim} p_0$
for $n \in [N]$. 
\begin{theorem}\label{thm:optimal_kl:finitd}
Let $S_0$ be the vector 
space of functions on $\Theta$ spanned by 
$\{ f(x, \cdot); x \in \mathcal X\}$. Assume that 
$S_0$ is finite dimensional, with dimension $d$. 
Let $\mu $ be any measure on $\Theta $ such that 
$S_0\subset L^2(\mu)$ and denote  
$r^2_\mu= \mathbb{E}_{\mu} \left(\mathbb{E}_p ( [f_1-\bar f]^2 )\right)$.  
For $\delta>0$ define
\begin{equation}\label{cond:scalarS}
J(\delta) = \!\!\!\!\!\inf_{a \in S_0; \|a\|=1} \!\!\!\Pr \left(\!\! \langle a,\! f_1 \!-\! \bar f\rangle_{L^2(\mu)} \!>\! \frac{2r_\mu}{\sqrt{ N\delta}} \right)\!.
\end{equation}
Then for $\delta > 0$ such that $J(\delta)+\delta < 1$,
a universal constant $C_1$,
and $M\in\mathbb{N}$ such that
$$
M \geq \frac{ 2d }{ J(\delta) } \left(\log N +  \log \left(\frac{4\delta}{J(\delta)}\right) + C_1\right), 
$$
we have that with probability $\geq 1- \delta- e^{-MJ(\delta)/4}$,
\begin{align*}
\min_{w\in\mathcal{W}_N}\mathrm{KL}(\pi_{w}||\pi) &=0.
\end{align*}
\end{theorem}
Note that the condition that $S_0$ has finite dimension $d<\infty$ is  
satisfied for all $d$-dimensional exponential 
families when $f_n (\theta ) = \log p(x_n|\theta)$. 
Further, note that \cref{thm:optimal_kl:finitd} still applies 
in situations where $d$ increases with $N$. 
To understand the behaviour of $J(\delta)$, 
we consider in \cref{app:theory} a 
full rank exponential family. We can show that,
under certain conditions, $J(\delta)$ converges
to a strictly positive value as $\delta N \to \infty$.
In this case, we can set $\delta = \omega(1/N)$, 
such that with probability $\geq 1 - \omega(1/N)$, we indeed have that
$\min_{w\in\mathcal{W}_N}\mathrm{KL}(\pi_{w}||\pi) =0 $ for $M\gtrsim d\log N$. 

In the interest of obtaining a more general approximation result, we now 
consider the case where $S_0$ is infinite dimensional,
but where we can find a finite-dimensional space $S_1$ which 
approximates $f_n(\theta)-\bar f(\theta)$ reasonably well. 
A typical setup where this is valid is when $\Theta \subseteq \mathbb{R}^D$,
the posterior concentrates (as $N\to\infty$) 
at a point $\theta_0 \in \Theta$ that maximizes $\bar f(\theta)$,
and each $f_n(\cdot)$ is smooth near $\theta_0$ and globally well-behaved 
in some sense. 
\cref{thm:optimal_kl} below states, roughly, that for a model satisfying these assumptions,
$\min_{w\in\mathcal{W}_N}\mathrm{KL}(\pi_{w}||\pi)$ will be small with high probability.
We give the exact conditions in \cref{app:theory}. Although they are perhaps strong,
the result is still indicative of when a good posterior approximation is obtained by our method.

Intuitively, we require that $f(x, \theta)$ be differentiable, strongly
concave in $\theta$, and ``smooth enough''. Letting $\theta_0$ be the 
unique value such that $\bar f(\theta) \leq \bar f(\theta_0)$ 
$\forall \theta \in \Theta$, we also require that we can
split  $f(x; \theta)$ into two parts, 
\begin{align*}
f(x, \theta) =f^{(1)}(x, \theta) + f^{(2)}(x, \theta), 
\end{align*}
where the span of 
$\left\{f^{(1)}(x, \cdot) - f^{(1)}(x, \theta_0) : x\in\mathcal{X}\right\}$
is a finite-dimensional vector space $S_1$.
This split needs to be made such that $f^{(2)}$ is negligible, in the sense that 
$f(x,\theta)\approx f^{(1)}(x,\theta)$ locally around $\theta_0$.
A usual case where these conditions hold
is when $f(x; \theta)$ is $C^3$ in $\theta$. In this case, we can obtain
$f^{(1)}(x;\theta)$ by a second-order Taylor expansion,
\begin{align*}
    f^{(1)}(x;\theta) = f(x;\theta_0) + (\theta-\theta_0)^T\nabla_\theta f(x;\theta_0) + (\theta- \theta_0)^T\nabla^2f(x; \theta_0) (\theta- \theta_0)/2,
\end{align*}
which has dimension bounded by $1 + D+ D(D+1)/2$. $f^{(2)}(x;\theta)$ is 
then the remainder term, which is indeed negligible around $\theta_0$ 
in the desired sense.
In \cref{app:theory}, we provide further discussion on the conditions of \cref{thm:optimal_kl}.
Recalling that a quantity is $o(1)$ as $N\to\infty$ if and only if it converges to 0,
we can then state the result as follows.

\begin{theorem}\label{thm:optimal_kl}
Suppose the assumptions A1-A4 in \cref{app:theory} hold, and we set
$M\gtrsim  D ( \log N + 1 ) $.
Then as $N\to\infty$, with probability at least $1 - o(1)$,
we have that
\begin{align*}
\min_{w\in\mathcal{W}_N}\mathrm{KL}(\pi_{w}||\pi) &= o(1).
\end{align*}
\end{theorem}

In the above example, we can examine the proof to obtain
a bound on $\mathrm{KL}(\pi_{w}||\pi)$ of order 
$1/\sqrt{N}$ with probability converging to 1, showing that this result does
indeed lead to a useful bound.

\subsection{Quasi-Newton weight optimization}\label{sec:theory-opt}
\cref{thm:optimal_kl:finitd,thm:optimal_kl} guarantee the existence of a high-quality coreset,
but do not provide any insight into whether it is possible to find it tractably. 
\cref{thm:convergence} addresses this remaining gap. 
In particular, intuitively, it asserts that as long as 
(1) the
regularization parameter $\tau>0$ is nonzero but small enough that it does
not interfere with the optimization (in particular, the minimum eigenvalue 
of $G(w)$ over the space), and
(2) the \emph{optimal} coreset is a good approximation to the full dataset
(which is already guaranteed by \cref{thm:optimal_kl}),
then
the weights $w_k$ at iteration $k$ of the approximate Newton method 
converge exponentially to a point close to a global optimum. 
In the result below, let $G(w)$ and $H(w)$ 
be defined as in \cref{eq:G_H_def}.

\begin{theorem}\label{thm:convergence}
Let $W\subseteq \mathcal{W}$ be a closed convex set,
$W^\star \subseteq \arg\!\min_{w\in W} \mathrm{KL}(\pi_w||\pi)$
be a maximal closed convex subset, and fix the regularization parameter $\tau > 0$.
Define
\begin{align*}
\xi = \inf_{w\in W} \min\left\{ \frac{\lambda}{\lambda+\tau} : \lambda \in \mathrm{eigvals}\,G(w), \, \lambda > 0\right\}.
\end{align*}
Suppose $\exists$ $\epsilon\in [0,1)$ and $\delta \geq 0$ such that
for all $w\in W$ 
\begin{align}
\hspace{-.3cm}\|\left(G(w)\!+\!\tau I\right)^{-1}H(w)(1\!-\!w^\star) \|\leq \epsilon \|w\!-\!w^\star\|\!+\!\delta, \label{eq:optw_error}
\end{align}
where $w^\star = \mathrm{proj}_{W^\star}(w)$.
Then the $k^\text{th}$ iterate $w_k\in W$ and its projection $w^\star_k = \mathrm{proj}_{W^\star}(w_k)$
of the projected Newton method 
defined in \cref{sec:quasinewton} by \cref{eq:qnc_step,,eq:G_H_def,,eq:qnc_projection},
initialized at $w_0\in W$ with a fixed step size $\gamma \in [0, 1]$
and regularization $\tau$
satisfies
\begin{align*}
\left\|w_{k} - w^\star_k\right\|
&\leq \eta^{k}\left\|w_0 - w^\star_0\right\| + \gamma\delta \left(\frac{1-\eta^k}{1-\eta}\right), \qquad \eta = 1-\gamma (\xi - \epsilon).
\end{align*}
\end{theorem}
For this result to make sense, we require that $\eta \in [0,1]$. We have that $\eta \geq 0$, since $\gamma \in [0,1]$, $\epsilon \geq 0$ and $\xi \leq 1 $ by definition. It may be the case that $\eta \geq 1$, in which case our result still holds, but is vacuous.

In order to use this result, the
minimum positive eigenvalue of $G(w)$ for $w\in W\subseteq \mathcal{W}$
needs to be bounded, as does the error of the optimal coreset weights $w^\star$ in the sense of \cref{eq:optw_error}. A notable corollary of \cref{thm:convergence} occurs when
there exists a $w^\star$ such that
\begin{align*}
\forall \theta\in\Theta, \quad \sum_{m=1}^M w^\star_m f_m(\theta) &= \sum_n f_n(\theta),
\end{align*}
i.e., the optimal coreset is identical to the true posterior distribution, as 
is guaranteed by \cref{thm:optimal_kl:finitd}.
For any model where the conditions of \cref{thm:optimal_kl:finitd} hold,
such as any exponential family model, \cref{eq:optw_error} holds with
$\epsilon = \delta = 0$. We can thus take $\gamma = 1$ to find that
\begin{align*}
\left\|w_{k} - w^\star_k\right\| &\leq (1-\xi)^{k}\left\|w_0 - w^\star_0\right\|.
\end{align*}
This provides some intuition on \cref{thm:convergence}; as long as 
the optimal coreset posterior is a reasonable approximation to $\pi$, and we use a 
small regularization $\tau>0$ (such that $\xi\approx 1$), then we generally
need only a small number of optimization steps to find the optimal coreset. 
The conditions of \cref{thm:optimal_kl:finitd} are sufficient, but not necessary,
for \cref{eq:optw_error} to hold. We present
further discussion in \cref{app:theory}. 

We note that if $\xi = 0$, then there is no
convergence, even in the setting where
\cref{thm:optimal_kl:finitd} holds.
However, it is important to emphasises that, for any given dataset, we are free to choose $\tau$. We can (and should) choose $\tau$ to be sufficiently small that $\xi$ is roughly $1$. In particular, this means setting $\tau$ significantly smaller than the minimum positive eigenvalue of $G(w)$.

\section{Experiments}\label{sec:experiments}
In this section, we compare our proposed quasi-Newton coreset (QNC) construction
against existing constructions---uniform subsampling (UNIF),
greedy iterative geodesic ascent (GIGA) \citep{Campbell18}
and iterative hard thresholding (IHT) \citep{Zhang21b}---as well as
the Laplace approximation (LAP), which represents what
one obtains by assuming posterior normality in the large-data setting. 
Experiments were performed on a machine with a 2.6GHz 6-Core 
Intel Core i7 processor, and 16GB memory;
code is available at  \url{https://github.com/trevorcampbell/quasi-newton-coresets-experiments}.

In each case, we use $S=500$ Monte Carlo samples during coreset construction.
We compute error metrics using 1000 samples from each method's approximate posterior.
We also compare to the baseline of sampling from the full posterior (FULL)
to establish a noise floor for the given comparison sample size of 1000.
In the synthetic Gaussian experiment, which is the simplest
one we consider, we see that sparse variational inference 
(SVI) \citep{Campbell19} is prohibitively slow, taking 
$4.2 \times 10^3$s to construct a coreset of size 1
(and scaling at best linearly with coreset size).
Thus, we do not compare against it here.
In \cref{app:experiments} we run a smaller,
lower dimensional, synthetic Gaussian experiment. 
Here, we compare against SVI, and confirm that it
is prohibitively slow for the size of datasets
that we consider.

For GIGA and IHT, we need to supply a low-cost approximation $\hat{\pi}$
to the posterior. To ensure these methods apply as generally as QNC and SVI,
we use a uniformly sampled coreset approximation of
size $M$ with weights $N/M$ (where $M$ is the same as the desired coreset size we are
constructing). We also use these weights for UNIF. 
In \cref{app:experiments} we provide additional results
for GIGA and IHT with a Laplace approximation used for $\hat{\pi}$. 
This does not lead to a significant improvement, and
we provide further discussion there.

Experiments are performed in settings
standard in the Bayesian coresets literature,
but with larger dataset sizes than those that could be considered previously.
For each experiment we plot the approximate reverse KL divergence
obtained by assuming posterior normality, and the build time.
In \cref{app:experiments} we provide comparisons on the additional metrics
of relative mean and covariance error, forward KL 
divergence and per sample time to sample from the 
respective posteriors. For the experiments in \cref{sec:sub:sparsereg,sec:sub:logreg}
with heavy tailed priors, we also compare using the maximum mean discrepancy \citep{gretton2012kernel} and kernel Stein discrepancy
\citep{liu2016kernelized,korba2021kernel},
with inverse multi-quadratic (IMQ) kernel.

Throughout the experiments, we see that QNC outperforms the other subsampling methods for 
most coreset sizes we consider. GIGA and IHT in particular are limited by a fixed choice of $\hat{\pi}$ and finite projection dimension defined
by our choice of $S=500$---we discuss this further in \cref{app:experiments}. Furthermore, outside of the synthetic Gaussian
experiment, we see that QNC outperforms LAP for coreset sizes above certain thresholds, which represent a small fraction of the full dataset. 

In \cref{app:experiments}, we also perform a 
sensitivity analysis for the parameters $S$, $K_{tune}$ and $\tau$ that we use 
in \cref{alg:anc_coreset}. We see that our results are generally not sensitive to the choice of these parameters, within reasonable ranges.
\begin{figure}[t!]
\begin{center}
\begin{subfigure}{0.48\columnwidth}
\includegraphics[width=0.48\columnwidth]{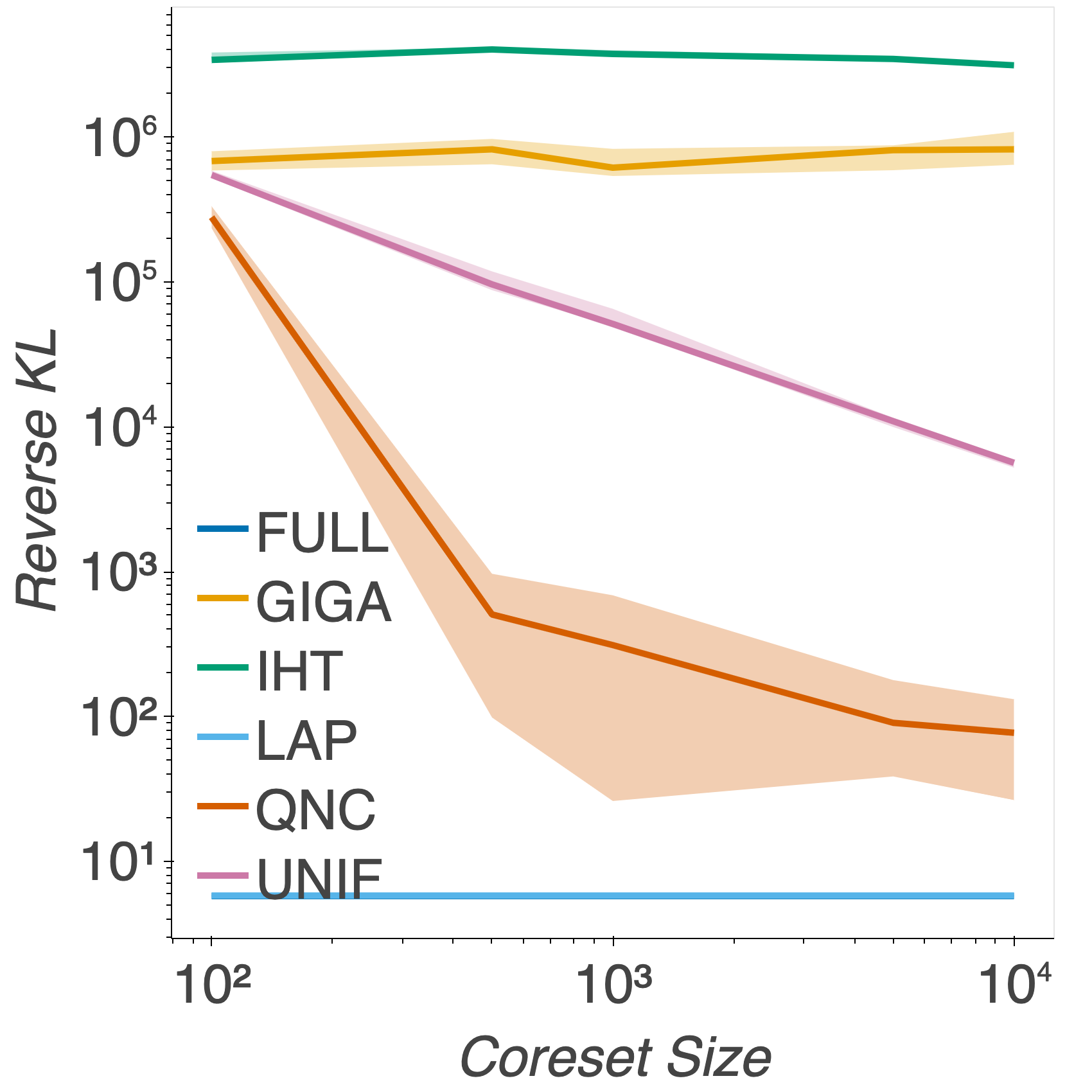}
\includegraphics[width=0.48\columnwidth]{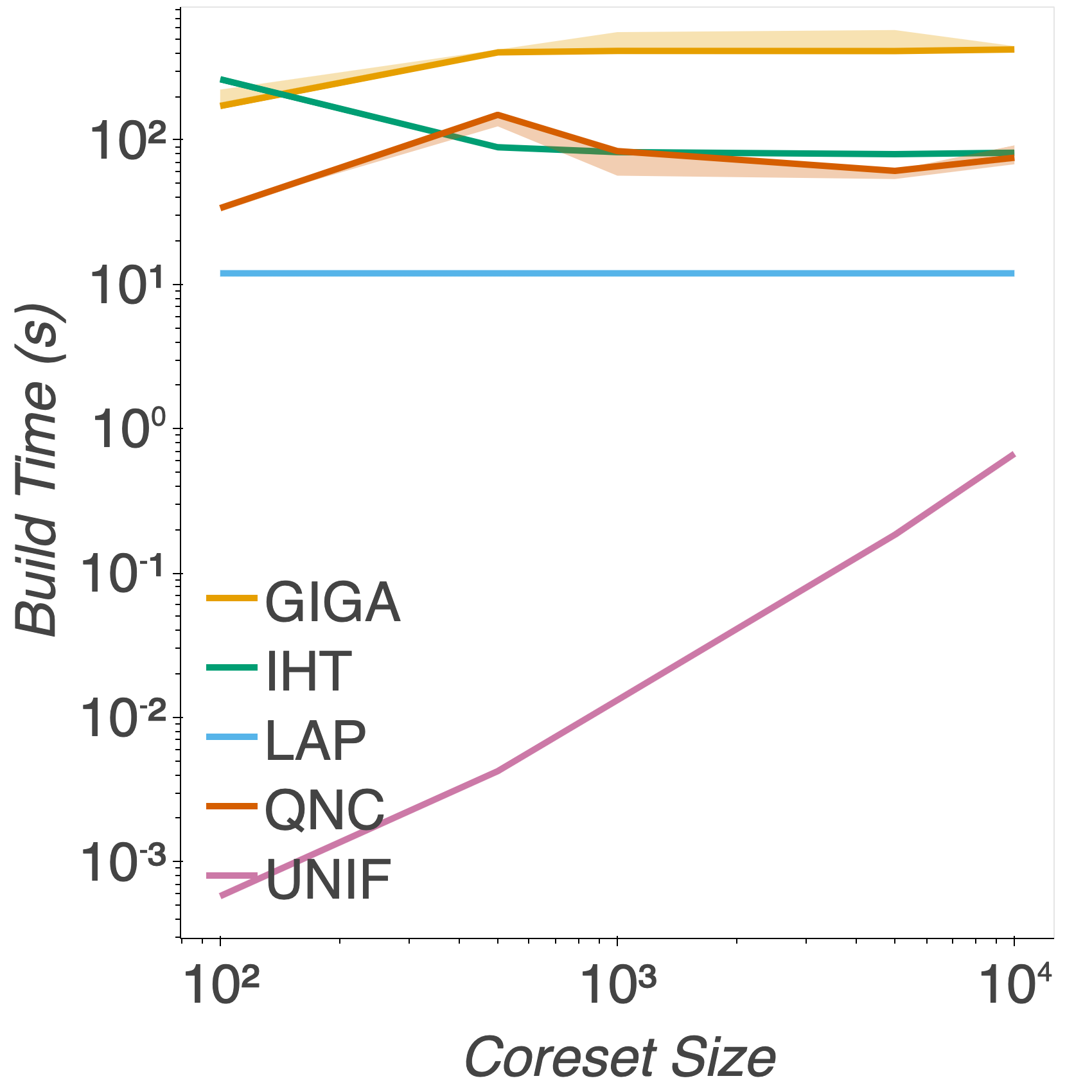}
\caption{Gaussian}
\label{fig:synth_gauss_kl_build_time}
\end{subfigure}
\begin{subfigure}{0.48\columnwidth}
\includegraphics[width=0.48\columnwidth]{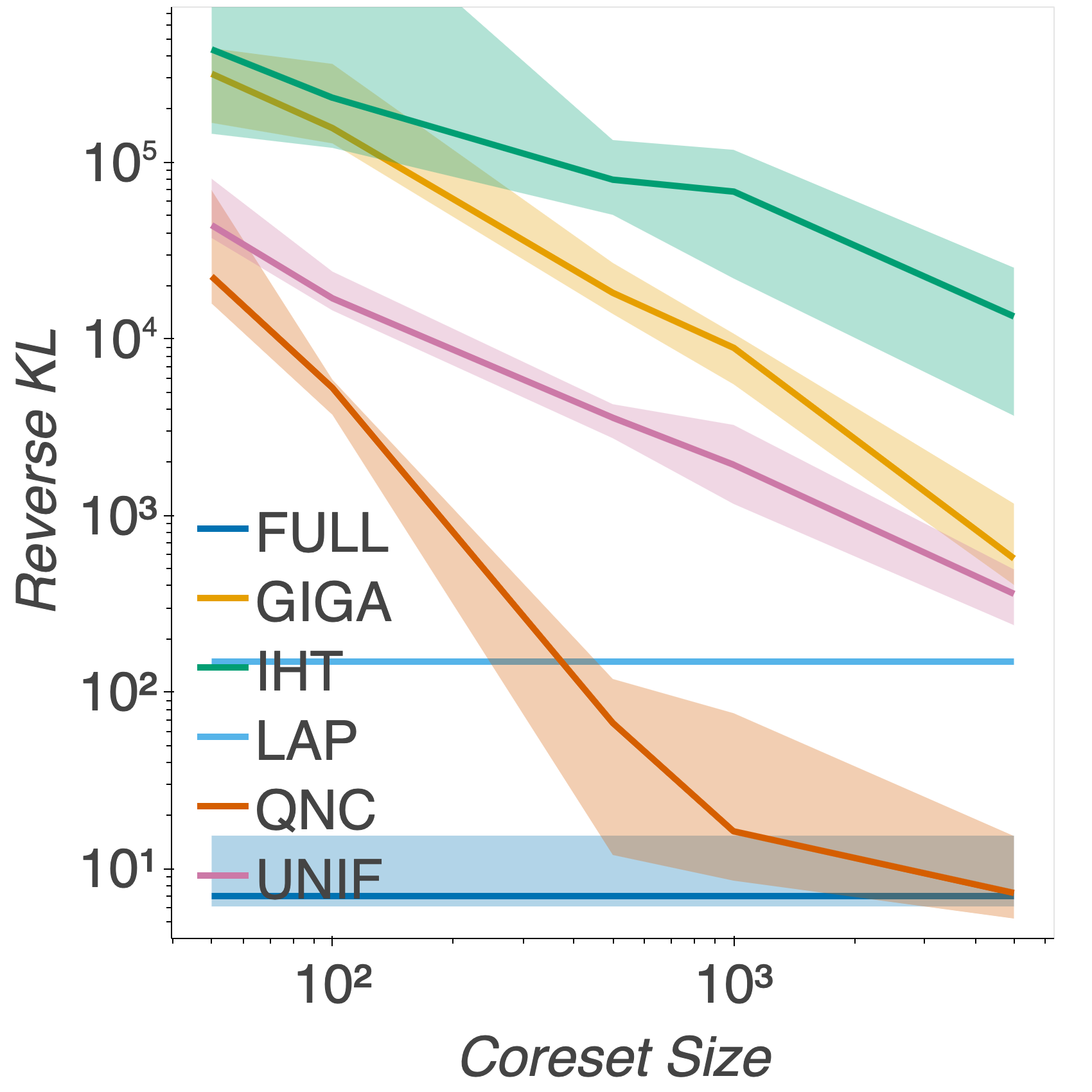}
\includegraphics[width=0.48\columnwidth]{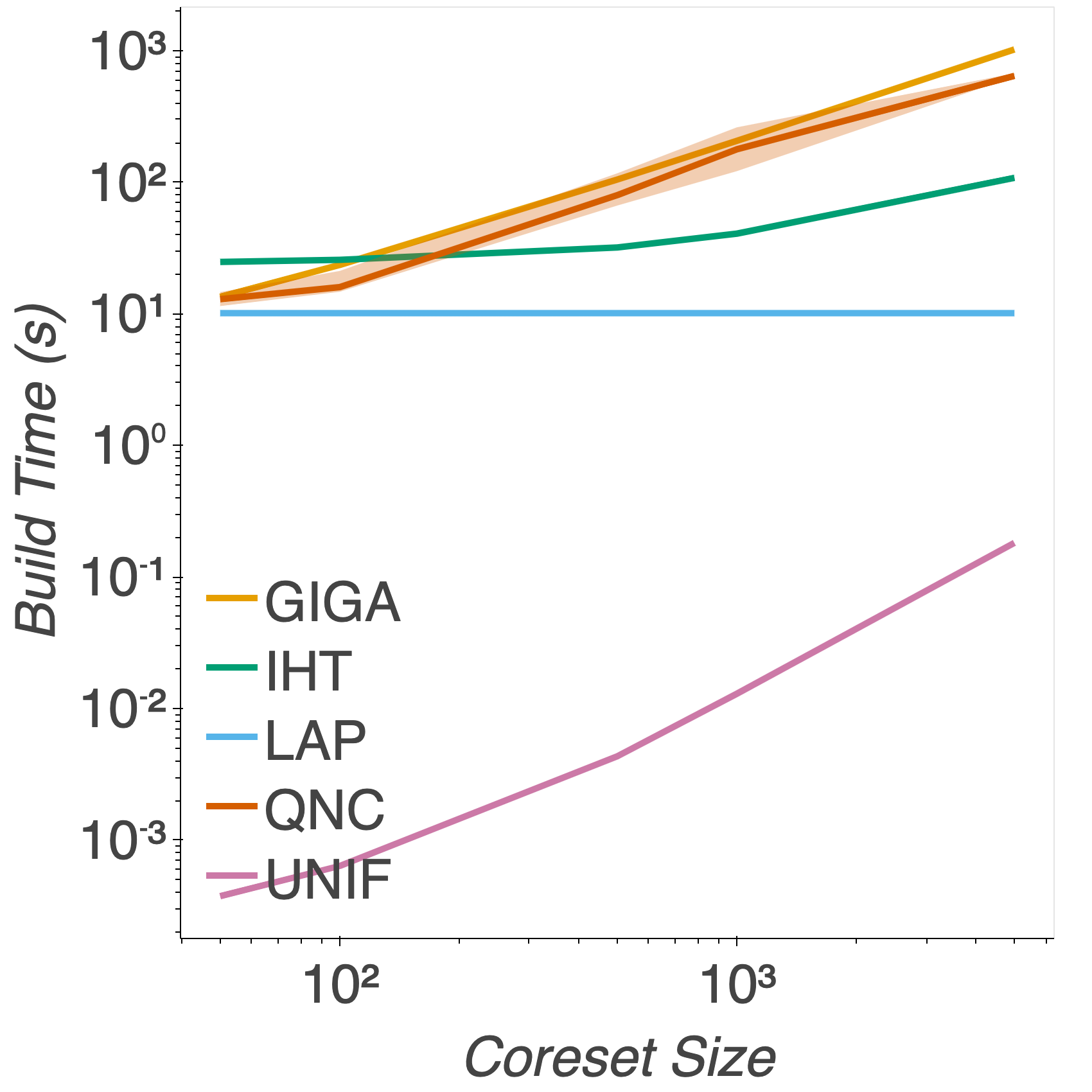}
\caption{Sparse linear regression}
\label{fig:delays_sparsereg_stein_kl_build_time}
\end{subfigure}

\begin{subfigure}{0.48\columnwidth}
\includegraphics[width=0.48\columnwidth]{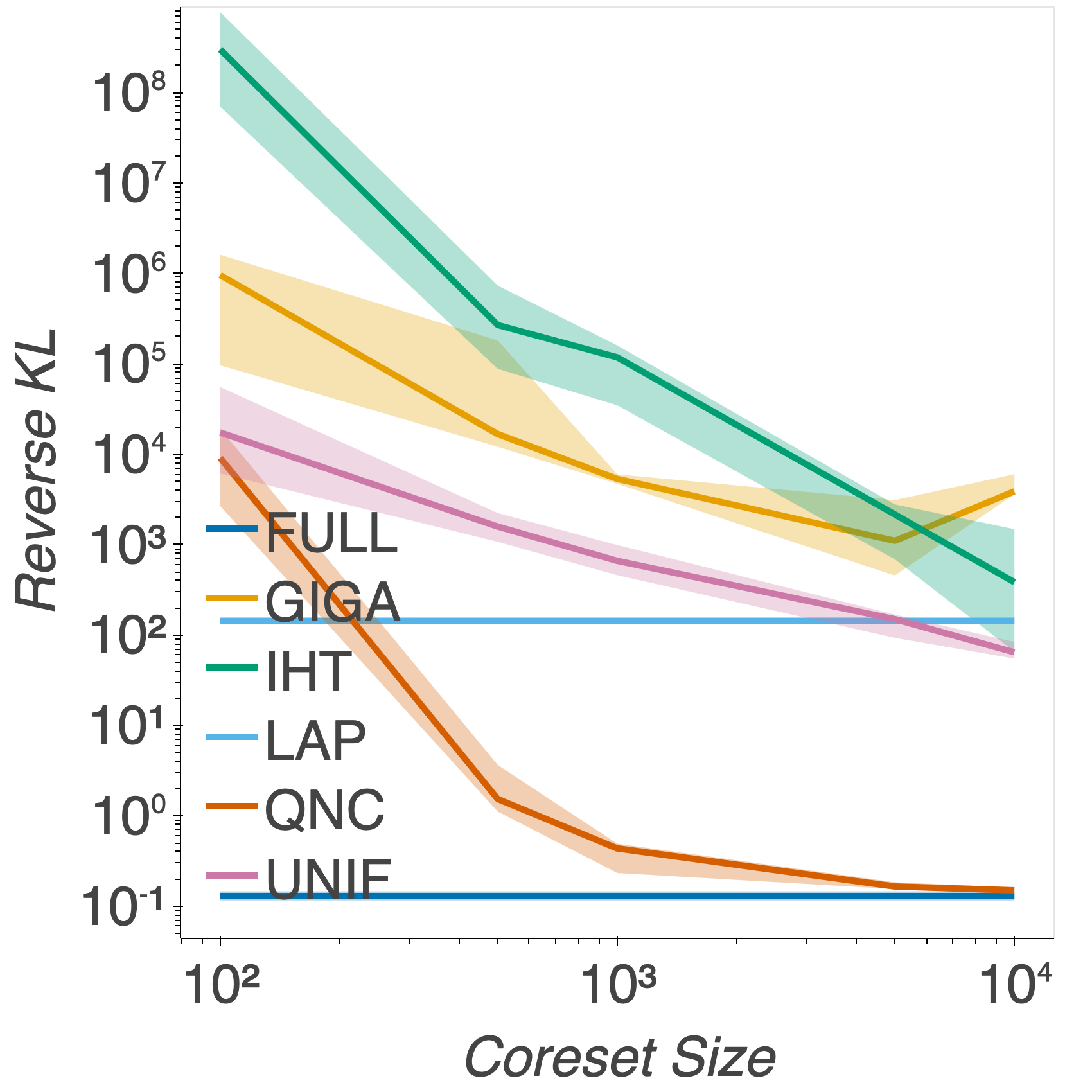}
\includegraphics[width=0.48\columnwidth]{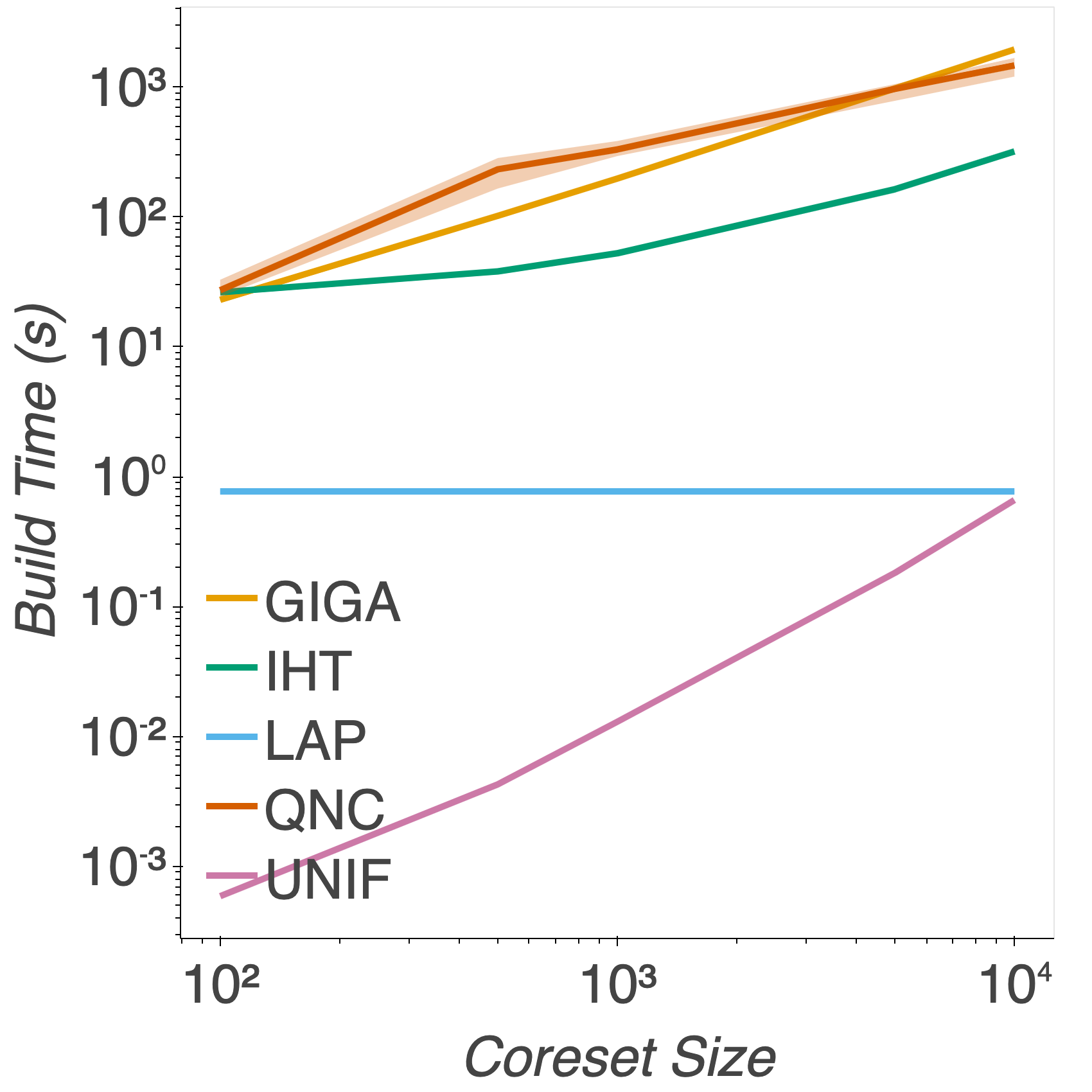}
\caption{Logistic regression}
\label{fig:delays_log_stein_kl_build_time}
\end{subfigure}
\begin{subfigure}{0.48\columnwidth}
\includegraphics[width=0.48\columnwidth]{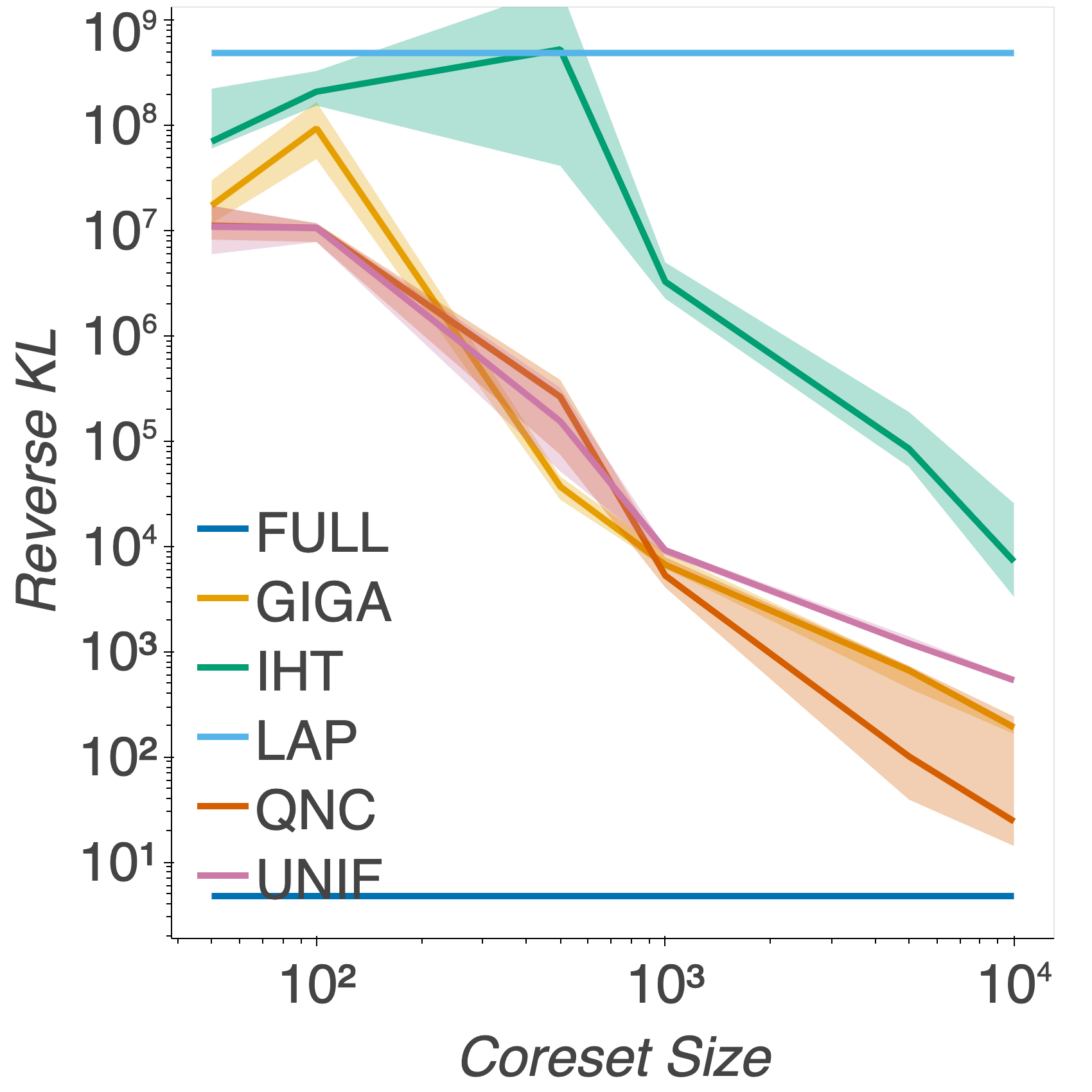}
\includegraphics[width=0.48\columnwidth]{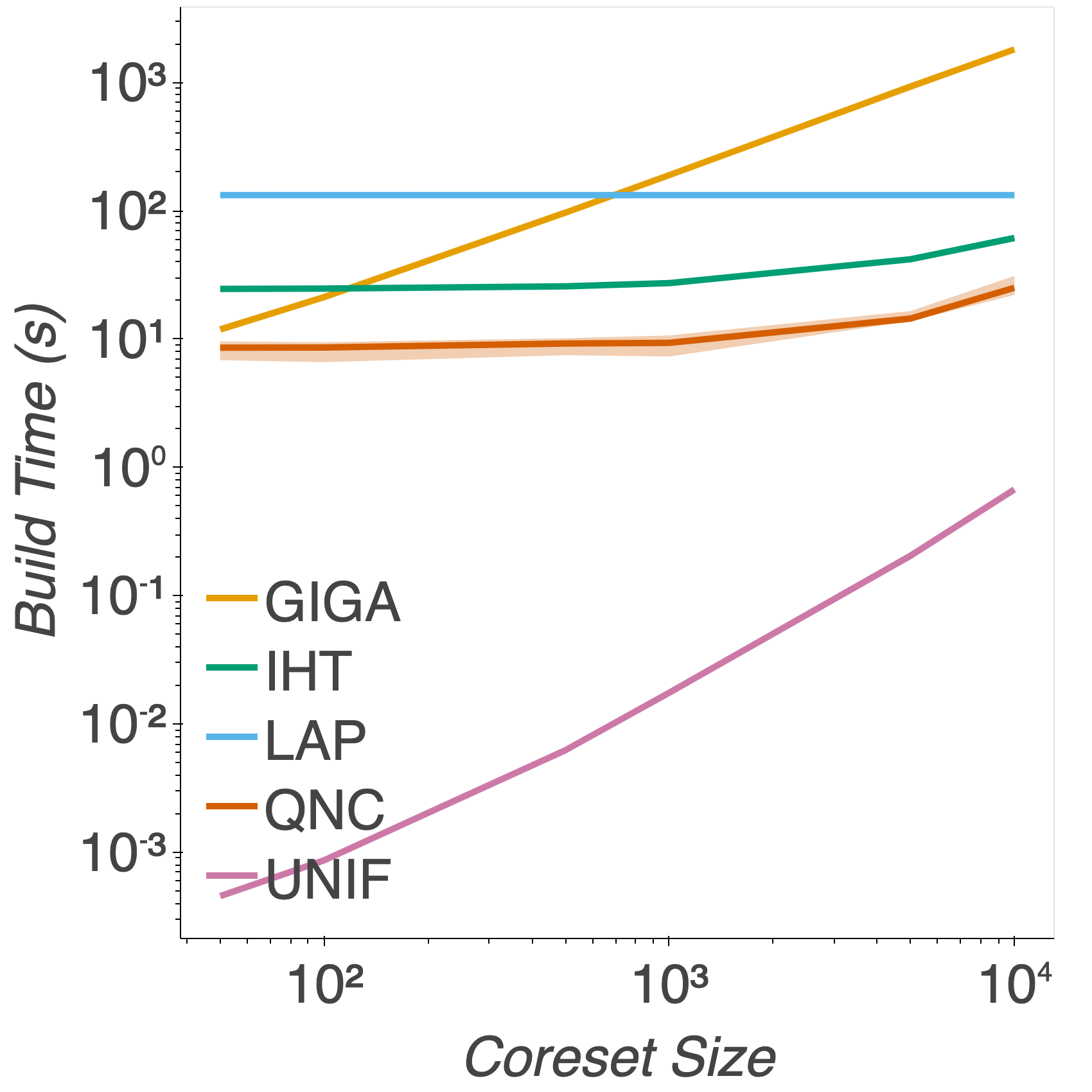}
\caption{Basis function regression}
\label{fig:housing_log_stein_kl_build_time}
\end{subfigure}
\caption{Reverse KL divergence (left) and build time 
in seconds (right) for each experiment. We plot the median and a shaded area 
between the 25$^\text{th}$/75$^\text{th}$ percentiles 
over 10 random trials. Our algorithm (QNC) provides an improvement in coreset quality, with a comparable run-time and less user input.}
\label{fig:all_kl_build_time}
\end{center}
\vskip -0.2in
\end{figure}

\subsection{Synthetic Gaussian location model}\label{sec:sub:synth_gauss}
Our first comparison is on a Gaussian location model,  with prior
$\theta \sim \mathcal{N}(\mu_0,\sigma_0 I)$, 
and likelihood $(X_n)_{n=1}^N \overset{\text{i.i.d.}}{\sim} \mathcal{N}(\theta, \sigma I)$
in $D$ dimensions.
Here, we take $\mu_0=0$ and $\sigma_0=1$. 
Closed form expressions are available for the subsampled posterior distributions 
\citep[Appendix B]{Campbell19}, and we can sample from them without MCMC.
We compare the methods on a synthetic dataset with $N=1,000,000$, $D=100$, 
where we generate the 
$(X_n)_{n=1}^N \overset{\text{i.i.d.}}{\sim} \mathcal{N}(\mu, \sigma I)$ 
and set $(\mu_i)_{i=1}^D \overset{\text{i.i.d.}}{\sim} \mathcal{N}(0,100)$ 
and $\sigma=100$. 
From \cref{fig:synth_gauss_kl_build_time} we see that 
our method outperforms the other subsampling methods for 
all coreset sizes. This example is largely illustrative;
we expect LAP to provide the exact posterior here by design, and this is indeed what we find (the reverse KL plots for LAP and FULL overlap). 
\subsection{Bayesian sparse linear regression}\label{sec:sub:sparsereg}
Next, we study a Bayesian sparse linear regression problem, where the data 
$\left(x_{n}, y_{n}\right)_{n=1}^{N}$ consists of a feature $x_{n} \in
\mathbb{R}^{D}$ and an outcome $y_{n} \in \mathbb{R}^N$. The posterior distribution
is that of $\theta \in \mathbb{R}^{D}$ in the model
\begin{align}
y_{n} = x_{n}^T\theta + \epsilon_{n}, \quad n=1,\ldots,N
\end{align}
where $\epsilon_n \overset{\text{i.i.d.}}{\sim} \mathcal{N}(0,\sigma),\ n=1,\ldots,N$. 
We place independent Cauchy priors on the coefficients $\theta$:
\begin{align*}
    \theta_i \overset{\text{i.i.d.}}{\sim} \text{Cauchy}(0,\tau \lambda_i), \quad i=1,\ldots,D,
\end{align*}
where the hyperpriors are $\lambda_i  \overset{\text{i.i.d.}}{\sim} \text{Half-Cauchy}(0,1)$, $\tau  \sim \text{Half-Cauchy}(0,\sigma_0)$,
with $\sigma_0=2$, and $\sigma \sim \Gamma(a_0,b_0)$ with $a_0=1,b_0=1$. We perform posterior inference on the 
$2D+2-$dimensional set of parameters 
$(\theta_1,\ldots,\theta_D,\lambda_1,\ldots,\lambda_D,\sigma,\tau)$. 
Sampling is performed using STAN \citep{carpenter2017stan}.
The dataset we study is a flight delays dataset,\footnote{This dataset was constructed by merging
airport on-time data from the US Bureau of Transportation Statistics
\url{https://www.transtats.bts.gov/DL_SelectFields.asp?gnoyr_VQ=FGJ} 
with historical weather records from
\url{https://wunderground.com}.} with $N = 100,000$ and $D=13$ (so that the overall inference problem is $28-$dimensional). The response variable $y_n$ is the delay in the departure time of a flight, and the features are meteorological and flight-specific information. 
We see in \cref{fig:delays_sparsereg_stein_kl_build_time} that 
our method outperforms the other subsampling methods for all coreset sizes,
and LAP for sizes above $500$---representing $0.5\%$ 
of the data. In \cref{app:experiments} we see that the target posterior in this
case has heavy tails, which makes the Laplace approximation particularly unsuited to this problem. We can see the effect this has even more clearly in the additional results presented there.

\subsection{Heavy-tailed Bayesian logistic regression}\label{sec:sub:logreg}
For this comparison, we perform Bayesian logistic regression with
parameter $\theta\in\mathbb{R}^{D}$ 
having a heavy-tailed Cauchy prior $\theta_i \overset{\text{i.i.d.}}{\sim}\text{Cauchy}(0,\sigma)$, $i=1,\dots,D$.
The data $\left(x_{n}, y_{n}\right)_{n=1}^{N}$ consists of a feature $x_{n} \in
\mathbb{R}^{D}$ and a label $y_{n} \in\{-1,1\}$. The relevant posterior distribution
is that of $\theta \in \mathbb{R}^{D}$ which governs the generation
of $y_{n}$ given $x_{n}$ via
\begin{align}
y_{n} \mid x_{n}, \theta \stackrel{\text { indep }}{\sim} \operatorname{Bern}\left(\frac{1}{1+e^{-x_{n}^{T} \theta}}\right) .
\end{align}
Again, sampling is performed using STAN.
As in \cref{sec:sub:sparsereg}, we use the flight delays dataset, so we have 
that $N = 100,000$ and $D=13$. The response variables $y_n$ are binarized, so 
that $y_n=1$ if a flight was cancelled or delayed by more than an hour, and $y_n=-1$ if it was not.
In \cref{fig:delays_log_stein_kl_build_time} we see that our method 
outperforms other subsampling methods for all coreset sizes,
and LAP for sizes above $500$---representing $0.5\%$ 
of the data.
\subsection{Bayesian radial basis function regression}\label{sec:sub:linreg}
Our final comparison is on a Bayesian basis function regression example.
This is a larger version of the same experiment performed by \citep{Campbell19},
with $D=301$ as before, but $N=100,000$. Full details are given in \cref{app:experiments}.
From \cref{fig:housing_log_stein_kl_build_time} we see that 
our method provides a significant improvement over LAP
for all coreset sizes, which performs particularly poorly in this experiment, despite the Gaussianity of the target posterior \citep[Appendix B]{Campbell19}.
It also outperforms the other subsampling methods for 
coreset sizes of 1000 and above (representing $1\%$ of the data) but this difference 
is not as marked as in the other examples.
The fact that QNC requires a large coreset to start providing a benefit
beyond UNIF
stems from the fact that, in this higher dimensional 
example, there are sometimes data points that are very 
influential for a certain basis, and both UNIF and QNC can miss these. 
This indicates the need (in some settings) for 
a secondary method that can check for and include 
these, without the significant cost that e.g. SVI entails.
We defer this study to future work.
However, we do note that for this example QNC has the lowest
build time, even outperforming LAP. This is because we only 
perform a small number of quasi-Newton steps before no further 
improvement is possible.
\section{Conclusion}
This paper introduces a novel method for data summarization
prior to Bayesian inference. In particular, the method first selects
a small subset of data uniformly randomly, and then optimizes the 
weights on those data points using a novel quasi-Newton method.
Theoretical results demonstrate that the method is guaranteed
to find a near-optimal coreset, and that the optimal coreset 
has a low KL divergence to the posterior with high probability.
Future work includes investigating the performance of the method
in more complex models, and studying
conditions under which the method can provide
compression in time sublinear in dataset size.

\begin{ack}
T.~Campbell was supported by a National Sciences and Engineering
Research Council of Canada (NSERC) Discovery Grant and Discovery Launch Supplement. C.~Naik was supported by the Engineering and Physical Sciences Research Council and Medical Research Council [award reference 1930478].
J.~Rousseau was supported by the  European Research Council (ERC) under the European Union’s Horizon 2020 research and innovation programme (grant agreement No 834175).
\end{ack}

\bibliographystyle{unsrtnat}
\bibliography{sources}

\section*{Checklist}

\begin{enumerate}

\item For all authors...
\begin{enumerate}
  \item Do the main claims made in the abstract and introduction accurately reflect the paper's contributions and scope?
    \answerYes{The claims made in the abstract are supported by the theoretical and experimental results in \cref{sec:theory} and \cref{sec:experiments}}
  \item Did you describe the limitations of your work?
    \answerYes{Methodological limitations are discussed in \cref{sec:algorithm} - mainly the fact that we need to used Monte Carlo estimates of some quantities of interest. The theoretical results in \cref{sec:theory} show the setting in which our method can be expected to work. We give some intuition for what settings these are in the main text, with further discussion in \cref{app:theory}. In our fourth experiment in \cref{sec:sub:linreg} we also discuss some of the limitations of our work in high dimensions.}
  \item Did you discuss any potential negative societal impacts of your work?
    \answerNo{Our work provides a generic preprocessing algorithm for sampling from probability distributions whose log density involves many summed terms. Our method is generic and foundational in nature: it has many downstream applications, including Bayesian inference, which may have a societal impact depending on the particular data and model under consideration. But we do not speculate on the impacts of potential downstream applications in this work.}
  \item Have you read the ethics review guidelines and ensured that your paper conforms to them?
    \answerYes{}
\end{enumerate}

\item If you are including theoretical results...
\begin{enumerate}
  \item Did you state the full set of assumptions of all theoretical results?
    \answerYes{The assumptions for \cref{thm:optimal_kl:finitd} and \cref{thm:convergence} are given in the theorem statements. Before the statement of \cref{thm:optimal_kl} we give intuitive explanations of its assumptions, but the full assumptions are explicitly laid out in \cref{app:theory}.}
        \item Did you include complete proofs of all theoretical results?
    \answerYes{Proofs of all theoretical results are included in \cref{sec:proofs}.}
\end{enumerate}

\item If you ran experiments...
\begin{enumerate}
  \item Did you include the code, data, and instructions needed to reproduce the main experimental results (either in the supplemental material or as a URL)?
    \answerYes{Code, data, and instructions needed to reproduce all experimental results is provided in the supplemental material.}
  \item Did you specify all the training details (e.g., data splits, hyperparameters, how they were chosen)?
    \answerYes{Important details are provided in \cref{sec:experiments}, and all further details are included in the code provided as part of the supplemental material.}
        \item Did you report error bars (e.g., with respect to the random seed after running experiments multiple times)?
    \answerYes{Error bars representing the 25$^\text{th}$/75$^\text{th}$ percentiles 
over 10 random trials are presented for each plot of the results.}
        \item Did you include the total amount of compute and the type of resources used (e.g., type of GPUs, internal cluster, or cloud provider)?
    \answerYes{The details of the machine used to perform all experiments are given in \cref{sec:experiments}.}
\end{enumerate}

\item If you are using existing assets (e.g., code, data, models) or curating/releasing new assets...
\begin{enumerate}
  \item If your work uses existing assets, did you cite the creators?
    \answerYes{Creators of all existing assets are cited.}
  \item Did you mention the license of the assets?
    \answerYes{All the datasets we use are publicly available, and some do not include a license. Where relevant, the license is mentioned.}
  \item Did you include any new assets either in the supplemental material or as a URL?
    \answerYes{We include new code in the supplemental material.}
  \item Did you discuss whether and how consent was obtained from people whose data you're using/curating?
    \answerNo{The datasets we use are publicly available, and do not pertain to identifiable individuals. We obey the license terms of the data where applicable, but do not discuss whether consent was obtained from the data creators.}
  \item Did you discuss whether the data you are using/curating contains personally identifiable information or offensive content?
    \answerNo{None of the datasets we use contain personally identifiable information or offensive content.}
\end{enumerate}

\item If you used crowdsourcing or conducted research with human subjects...
\begin{enumerate}
  \item Did you include the full text of instructions given to participants and screenshots, if applicable?
    \answerNA{}
  \item Did you describe any potential participant risks, with links to Institutional Review Board (IRB) approvals, if applicable?
    \answerNA{}
  \item Did you include the estimated hourly wage paid to participants and the total amount spent on participant compensation?
    \answerNA{}
\end{enumerate}

\end{enumerate}

\newpage
\appendix 

\section{Theoretical analysis}\label{app:theory}
In this section we give further details on the theoretical results
presented in \cref{sec:theory}. We start by giving the exact conditions
needed for \cref{thm:optimal_kl} to hold.
\setcounter{equation}{17}
\subsection{Full conditions for \cref{thm:optimal_kl}}
In order to state \cref{thm:optimal_kl}, 
we define $\theta_0$ as the 
unique value such that $\bar f(\theta) \leq \bar f(\theta_0)$ 
$\forall \theta \in \Theta$, where 
$\bar f(\cdot)  = \mathbb{E}_p( f_n(\cdot))$.
If we split  $f(x; \theta)$ into two parts, 
\begin{align*}
f(x, \theta) =f^{(1)}(x, \theta) + f^{(2)}(x, \theta), 
\end{align*}
then we require that
there exist constants $\epsilon_0, L_0, L_1, L_2>0$ such that:

$\bullet$ \textbf{A1}: For all 
    $x$, $f(x, \theta)$ is differentiable and strongly
    concave in $\theta$ with constant $L_0$. \\
$\bullet$ \textbf{A2}:
For all $0 < \epsilon < \epsilon_0$  and $\| \theta - \theta_0\|\leq \epsilon$,
 \begin{align*}
            &|f^{(2)}(x, \theta)| \leq R(x)r(\epsilon) \|\theta -     \theta_0\|^2, 
 \end{align*}
where $R(x), r(\epsilon)$ are positive functions, $r$ is monotone increasing, $\lim_{\epsilon\to 0} r(\epsilon) = 0$, and
\begin{align*}
4r(\epsilon_0)\left[\mathbb{E}_p(R(X))+1\right] \leq L_0, \quad\mathbb{E}_p(R(X)^2)<\infty.
\end{align*}
$\bullet$ \textbf{A3}: For all $\|\theta-\theta_0\|\leq \epsilon_0$,
 \begin{align*}
     \bar f(\theta_0) - \bar f(\theta) \leq L_1 \|\theta-\theta_0\|^2, \quad \mathrm{Var}_p(f(X; \theta_0)-f(X; \theta)) \leq L_2\|\theta-\theta_0\|^2 .
 \end{align*}
$\bullet$ \textbf{A4}: $\inf_{\|\theta-\theta_0\|\leq \epsilon_0}\pi_0(\theta) > 0$,
and $\sup_{\theta} \pi_0(\theta) < \infty$.

Here, the functions $R(x), r(\epsilon)$ 
control the size of the ``discarded part'' 
$f^{(2)}$ locally around $\theta_0$; 
these should be small to ensure that 
$f(x,\theta)\approx f^{(1)}(x,\theta)$
locally. 
Note also that when $r(\epsilon)\asymp\epsilon$ as $\epsilon\to 0$,
the result of \cref{thm:optimal_kl} holds even if $D$ is increasing
such that $D = o(\log N)$.

\subsection{Discussion on conditions of \cref{thm:optimal_kl:finitd,thm:optimal_kl,thm:convergence}}
In order for \cref{thm:optimal_kl:finitd} to hold, we 
require that $S_0$ is finite dimensional 
(say with dimension $d$),
where $S_0$ is the vector 
space of functions on $\Theta$ spanned by 
$\{ f(x, \cdot); x \in \mathcal X\}$.
For this to hold, we need to find a set of $d$
basis functions for the space, even though $\mathcal X$
may be potentially uncountable. There are two settings
in which we can clearly see that this holds. 
The first of these is when $f(x_n,\theta) = \log p(x_n|\theta)$
and we can write the likelihood as a $d$-dimensional exponential 
family. The standard exponential family form gives us the 
desired basis for $S_0$. The second setting is when $\mathcal X$
is in fact a finite set. If $x \in \mathcal X$ can
only take $d$ values, then we can easily find a basis as before.

In the exponential family case, we can gain
some more intuition on the behaviour of $J(\delta)$.
Consider a full rank, $d$-dimensional exponential 
family with $f_n (\theta ) = \log p(x_n|\theta)$. 
We can write
\begin{align*}
    (f_1 - \bar f )(\theta) = A(\theta)^T (B(X_1) - \mathbb{E}_p(B(X_1))) - K(X_1),
\end{align*}
where $X_1$ is the first observed data point 
(corresponding to the potential $f_1$), 
$B(\cdot)$ is the (full rank) sufficient statistic and
$A(\theta)$ is the natural parameter. $K(X_1)$ arises from the 
log-base density, and is a function of $X$ only.
If there exists $\mu$ such that 
$R_A \coloneqq \int A(\theta) A(\theta)^T d\mu(\theta) $ has rank $d$, then  
\begin{align*}
    J(\delta) \xrightarrow{\delta N \to \infty}  \inf_{a \in S_0; \|a\|=1} \Pr \left( \langle a, f_1 - \bar f\rangle_{L^2(\mu)} > 0 \right)>0.
\end{align*}
We can see this as follows. Firstly, $J(\delta)$ is 
monotonically increasing as $\delta N \to \infty$.
Hence, we can interchange the limit with the infimum to see that 
\begin{align*}
    \lim_{\delta N \to \infty} J(\delta) = \inf_{a \in S_0; \|a\|=1} \Pr \left( \langle a, f_1 - \bar f\rangle_{L^2(\mu)} > \lim_{\delta N \to \infty} \frac{2r_{\mu}}{\sqrt{N\delta}} \right),
\end{align*}
and $\lim_{\delta N \to \infty} \frac{2r_{\mu}}{\sqrt{N\delta}} = 0$.
To see that this limit is strictly positive, 
note that the rank condition means that 
$S_0 = \text{span}\left(A_1(\theta), \ldots, A_d(\theta)\right)$.
Thus, $a\in S_0 \iff a = V^TA$ for some $V$. Choosing $\mu$ such that 
$\mathbb{E}_{\mu}(A) = 0$, 
\begin{align*}
    \langle a, f_1 - \bar f\rangle_{L^2(\mu)}  &= V^T\left( \int A(\theta) A(\theta)^T d\mu(\theta) \right)(B(X_1) - \mathbb{E}_p(B(X_1))) - V^T\mathbb{E}_{\mu}(A)K(X_1) \\
    &= V^TR_A(B(X_1) - \mathbb{E}_p(B(X_1))) \qquad \text{using our choice of $\mu$}.
\end{align*}
Thus
\begin{align*}
     \Pr \left( \langle a, f_1 - \bar f\rangle_{L^2(\mu)} \geq 0 \right) =  \Pr \left( V^TR_A(B(X_1) - \mathbb{E}_p(B(X_1)))  \geq 0 \right).
\end{align*}
Furthermore, $\|a\|=1 \iff V^TR_AV = 1$. 
Since $\mathbb{E}_p\left(V^TR_A(B(X_1) - \mathbb{E}_p(B(X_1))) \right) = 0$,
\begin{align*}
     &\Pr \left( V^TR_A(B(X_1) - \mathbb{E}_p(B(X_1)))  \geq 0 \right) = 0 \\
     &\iff V^TR_A(B(X_1) - \mathbb{E}_p(B(X_1))) = 0 \quad \text{almost surely}.
\end{align*}
However, this is impossible since both $R_A$ and $B$ are of full rank.
Thus, $\forall V$, $\Pr (V) \coloneqq \Pr \left( V^TR_A(B(X_1) - \mathbb{E}_p(B(X_1)))  \geq 0 \right) >0$.
Now, if $\exists V_n$ such that $\Pr(V_n) \to 0$,
the fact that we are on the compact space $\|a\|=1$ means that 
there exists a subsequence $V_{n_k} \to V_*$ with $V_*^TR_AV_* = 1$.
We would then have 
\begin{align*}
     V_{n_k}^TR_A(B(X_1) - \mathbb{E}_p(B(X_1))) \xrightarrow{d} V_{*}^TR_A(B(X_1) - \mathbb{E}_p(B(X_1))),
\end{align*}
where $\xrightarrow{d}$ indicates convergence in distribution.
Then at the limit, $\Pr (V_*) = 0$, which we have shown is impossible.
Thus, we do indeed have that 
\begin{align*}
    \lim_{\delta N \to \infty} J(\delta) =  \inf \Pr (V) > 0.
\end{align*}
In this case, we can set $\delta = \omega(1/N)$, 
such that with probability $\geq 1 - \omega(1/N)$, we indeed have that
$\min_{w\in\mathcal{W}_N}\mathrm{KL}(\pi_{w}||\pi) =0 $ for $M\gtrsim d\log N$. 

In order for \cref{thm:optimal_kl} to hold, we require
additional smoothness and concavity conditions on $f(x,\theta)$. The
smoothness conditions are satisfied if, for example, 
$f(x; \theta)$ is $C^3$ in $\theta$, as discussed in \cref{sec:theory}. For example, we can 
consider a logistic regression problem.
Here, we are not in the exponential family setting.
However, the logistic regression  
log-likelihood is concave and $C^3$ 
in $\theta$. Hence we can find $f^{(1)}$
and $f^{(2)}$ via a second-order Taylor
expansion. Finally, A4 is satisfied by any
prior that is strictly positive and finite, 
such as the Cauchy prior we use in our logistic
regression experiment. 

However, the logistic
log-likelihood is concave but not strictly concave.
Thus A1 does not hold, and our logistic regression
experiment is not covered by the assumptions of \cref{thm:optimal_kl}. 
It is therefore reassuring to see that our method
obtains favourable empirical results in this
experiment, and we conjecture that the strong concavity
condition is sufficient but not necessary. We leave the
relaxing of this condition to future work.

In the statement of \cref{thm:convergence}, we 
suppose that $\exists$ $\epsilon\in [0,1)$ and $\delta \geq 0$ such that
for all $w\in W$ 
\begin{align}
\hspace{-.3cm}\|\left(G(w)\!+\!\tau I\right)^{-1}H(w)(1\!-\!w^\star) \|\leq \epsilon \|w\!-\!w^\star\|\!+\!\delta, \label{eq:optw_error_app}
\end{align}
where $w^\star = \mathrm{proj}_{W^\star}(w)$ and 
$W^\star \subseteq \arg\!\min_{w\in W} \mathrm{KL}(\pi_w||\pi)$.

If we can show that
\begin{align}\label{eq:w_star_exact}
\forall \theta\in\Theta, \quad \sum_{m=1}^M w^\star_m f_m(\theta) &= \sum_n f_n(\theta),
\end{align}
then
\begin{align}
H(w)(1-w^\star) &= \mathrm{Cov}_{w}\left[g, f^T(1-w^\star)\right] \nonumber \\
&= 0.\label{eq:H_zero}
\end{align}
and so thus the assumption
will hold with $\epsilon = \delta = 0$. 

\cref{thm:optimal_kl:finitd} tells us
that $\pi_{w^\star}(\theta) = \pi(\theta)$ for almost 
all values of $\theta$. This implies that \cref{eq:w_star_exact} 
also holds for almost all values of $\theta$, 
and so \cref{eq:H_zero} holds. 
This means that the assumption in \cref{thm:convergence}
does in fact hold with $\epsilon = \delta = 0$. 

Thus, in any setting where the conditions 
of \cref{thm:optimal_kl:finitd} hold (such as the two 
discussed above), \cref{thm:convergence} holds as well.
This reasoning also gives us some intuition into how we can
generalise this. Intuitively, we require that 
\begin{align}\label{eq:w_star_approx}
\forall \theta\in\Theta, \quad \sum_{m=1}^M w^\star_m f_m(\theta) &\approx \sum_n f_n(\theta),
\end{align}
so that $H(w)(1-w^\star) \approx 0$. 
Essentially, we see that the better the quality of the 
optimal coreset $w^\star$, the tighter the bound
we can find for \cref{eq:optw_error_app}. 
We conjecture that if $ \exists\ \xi$ such that 
we can uniformly bound
\begin{align*}
    \sup_{\theta\in \Theta} \left| \sum_{m=1}^M w^\star_m f_m(\theta) - \sum_n f_n(\theta) \right| \leq \xi,    
\end{align*}
then we can find $\epsilon$ and $\delta$ 
such that \cref{eq:optw_error_app} holds.
We leave the theoretical examination of this
claim for future work. 

\section{Proofs}\label{sec:proofs}
In this section we give the full proofs of \cref{thm:optimal_kl:finitd,thm:optimal_kl,thm:convergence}. 
The proofs of \cref{thm:optimal_kl:finitd,thm:optimal_kl} are quite technical,
so in each case we give start by giving a roadmap of the proof ideas.
\subsection{Proof of \cref{thm:optimal_kl:finitd}}
\textit{Roadmap.}
\begin{enumerate}
    \item We start the proof by giving in \cref{convHull} a sufficient condition for our desired result to hold. Proving that this condition holds will be the target of the rest of the proof. 
    \item The next step is to show that this is indeed a sufficient condition. We do this by deriving the form for the KL divergence found in \cref{KL:expression}. 
    \item We then proceed by lower bounding the probability that the quantity on the right side of \cref{convHull} lies within a ball of a given radius.
    \item Next, we find a lower bound for the probability that we can get the left hand side of \cref{convHull} to be equal to any value within the same ball as above.
    \item If the events introduced in 2. and 3. hold, then we will be able to satisfy our sufficient condition \cref{convHull}. This implies that our desired result will also hold. Thus we can combine the two relevant lower bounds detailed in 2. and 3. to get a lower bound on the probability of our desired result holding.
    \item Obtaining this final lower bound requires setting a condition on the coreset size $M$. We conclude the proof by rewriting condition to make it more interpretable.
\end{enumerate}
\begin{proof}
We select $M$ indices uniformly at random from $[N]$, 
corresponding to $M$ potential functions $f_n(\theta)$.
Because the functions $f_n$ are generated i.i.d.~and 
the indices chosen uniformly, the 
particular indices selected are unimportant for this 
proof; without loss of generality,
we can assume the $M$ selected indices are $n=1,\dots, M$.

Let $\mu $ be a probability measure such that $S_0$ is 
the associated $d$ dimensional subspace of $L^2(\mu)$ and let $M\in[N]$. In what follows, we assume that all
norms and inner products are the weighted $L^2(\mu)$ norms and 
inner products respectively, unless otherwise stated.
Further, define $\widetilde{f}_N (\cdot ) = \sum_{n=1}^N [f_n - \bar f](\cdot ) /N$.
We show below  that with 
probability $ \geq 1-\delta$, 
there exists $v_1, \cdots , v_M \geq 0$ with 
$\sum_m v_m =1$  such that 
\begin{equation}\label{convHull}
 \sum_{m=1}^M (f_m - \bar f) v_m = \widetilde{f}_N.
 \end{equation}
For such a $v = (v_1, \cdots, v_M)$, set 
$w(v) = Nv$ (with the implicit convention that 
$w(v)_j=0$ for $j>M$). 
Our goal is to prove that $\mathrm{KL}(\pi_{w(v)}||\pi) = 0$. We do this by starting with the form for the KL divergence derived in the proof of Lemma 3 of \citet{Campbell19}:
\begin{align*}
    \mathrm{KL}(\pi_{w(v)}||\pi) = 2(1-w)^T\mathbb{E}\left[C(\gamma(T))\right] (1-w),
\end{align*}
where:
\begin{itemize}
    \item $T \sim \text{Beta}(1,2)$.
    \item $\gamma(t)$ is the path defined by $\gamma(t) = (1-t)w + t1$ for $t \in [0,1]$, coreset weights $w \in \mathbb{R}^N$, and unit vector $1 \in \mathbb{R}^N$.
    \item $C(w)$ is the Fisher information metric \citet[p. 33,34]{amari2016information} defined as:
    \begin{align*}
        C(w)  &\coloneqq \operatorname{Cov}_{w} [f,f]  \\
        &\coloneqq \mathbb{E}_{\pi_w} \left[ \left( f-\mathbb{E} _{ \pi_w } (f) \right) \left( f-\mathbb{E} _{ \pi_w } (f) \right) ^T \right].
    \end{align*}

    Note that we have changed the notation from that of \cite{Campbell19}, to avoid a clash of notation with terms already defined in our work.
\end{itemize}
Using the fact that the p.d.f. of a $\text{Beta}(1,2)$ distribution is given by $f_T(t) = 2(1-t)$, we therefore have that
\begin{align}
    \mathrm{KL} (\pi_{w(v)}||\pi) &= 2(1-w)^T \mathbb{E} \left[ C(\gamma (T))\right] (1-w) \notag \\
    &= 2(1-w)^T \left[ \int_0^1 2(1-t) C ( \gamma (t)) \mathrm{d}t \right] (1-w) \notag \\
 &= 4 (1-w)^T \left[ \int_0^1 (1-t) \mathbb{E} _{w(v),t} \left[ \left( f- \mathbb{E} _{\pi _{ \gamma (t)}} (f) \right) \left( f-\mathbb{E} _{\pi _{ \gamma (t) }} (f) \right)^T \right] \mathrm{d}t \right] (1-w) \notag \\
&= 4\int_0^1 (1-t)\mathbb{E} _{w(v),t} \left[\left((1-w)^T\left(f-\mathbb{E} _{w(v),t} (f)\right)\right)^2 \right]\mathrm{d}t \notag \\
&= 4\int_0^1 (1-t)\mathbb{E} _{w(v),t} \left[\left(\sum _{n=1}^N (1-w_n) f_n - \mathbb{E} _{w(v),t} \left(\sum _{n=1}^N(1-w_n)f_n\right)\right)^2 \right]\mathrm{d}t, \label{KL:old}
\end{align}
where $\mathbb{E} _{w(v),t}$ refers to taking expectations under $\pi _{w(v),t}$, the coreset posterior corresponding to weights given by $\gamma(t) = (1-t) w + t1$. To be explicit:
\begin{align}
    \pi _{w(v),t} (\theta)=  \frac{  e^{t\sum _n f_n + (1-t) \sum _{n} w_n f_n}\pi _0 (\theta)  }{ \int_\Theta e^{t\sum _n f_n + (1-t) \sum _{n}w_n f_n}\pi_0(\theta)d\theta }. \label{pi_t}
\end{align}

We now note that 
\begin{align*}
    \frac{1}{N}\sum_{n=1}^N (1-w_n)f_n &= \frac{1}{N}\sum_{n=1}^N (f_n - \bar f) - \frac{1}{N}\sum_{n=1}^N w_n(f_n - \bar f) \\
    &= \widetilde{f}_N- \sum_m v_m (f_m-\bar f),
\end{align*}
and thus 
\begin{align*}
    t\sum_n f_n + (1-t) \sum_{n}w_nf_n &= \sum_n f_n - (1-t) \sum_{n}(1-w_n)f_n \\
    &= \sum_n f_n - N(1-t)  [ \widetilde{f}_N - \sum_m v_m (f_m-\bar f) ].
\end{align*}
Thus, by substituting into \cref{pi_t}, we can write $\mathbb{E} _{w(v),t}$ as 
$$ \mathbb{E}_{w(v),t} (  h ) =  \frac{ \int_\Theta h(\theta) e^{ \sum_n f_n(\theta) - N(1-t)  [ \widetilde{f}_N - \sum_m v_m (f_m-\bar f) ]}\pi_0(\theta)d\theta   }{ \int_\Theta e^{ \sum_n f_n(\theta) - N(1-t)  [ \widetilde{f}_N- \sum_m v_m (f_m-\bar f) ]}\pi_0(\theta)d\theta }.
$$ 
Furthermore, substituting into \cref{KL:old} we have that
\begin{align}  \label{KL:expression}
 \mathrm{KL}(\pi_{w(v)}||\pi) &=4 N^2\int_0^1 (1-t)  \mathbb{E}_{w(v),t} \left[ \left(\widetilde{f}_N- \sum_m v_m (f_m-\bar f) - \mathbb{E}_{w(v),t} \left(\widetilde{f}_N- \sum_m v_m (f_m-\bar f) \right) \right)^2\right] \mathrm{d}t .
\end{align} 
If we can prove \cref{convHull}, \cref{KL:expression} 
would give us that  $\mathrm{KL}(\pi_{w(v)}||\pi)=0$, 
thus completing the proof. Of course, if \cref{convHull} holds
then the coreset and full posteriors are equal, so
$\mathrm{KL}(\pi_{w(v)}||\pi)$ is trivially zero. However, targeting this expression
will also be helpful for the proof of \cref{thm:optimal_kl}.
In order to prove \cref{convHull}, first note that
$\| \widetilde{f}_N\|_{L^2(\mu) }^2 = \mathbb{E}_\mu\left(\left(\frac{1}{N}\sum_n(f_n-\bar f)\right)^2\right)$. 
With $r^2_\mu= \mathbb{E}_{\mu} \left(\mathbb{E}_p ( [f_1-\bar f]^2 )\right)$, we have by Markov's inequality that, $\forall \delta>0$, 
\begin{align*}
\Pr\left(\| \widetilde{f}_N\|_{L^2(\mu) }^2 > \frac{r_\mu^2} {N\delta}  \right)  &\leq \frac{N\delta}{r_\mu^2}  \mathbb{E}_p\left(\mathbb{E}_\mu\left(\left(\frac{1}{N}\sum_n(f_n-\bar f)\right)^2\right)\right) \\
&= \frac{N\delta}{r_\mu^2}  \mathbb{E}_\mu\left(\mathbb{E}_p\left(\left(\frac{1}{N}\sum_n(f_n-\bar f)\right)^2\right)\right) \\
&= \frac{N\delta}{r_\mu^2} \frac{1}{N} \mathbb{E}_\mu\left(\mathbb{E}_p\left(\left(f_1-\bar f\right)^2\right)\right) \\
&= \delta.
\end{align*}
Defining $t_N \coloneqq \sqrt{r_\mu^2 /(N\delta)}$,
it is enough to show that, with high probability, the 
convex hull of  $ (f_m - \bar f), m\in [M]$ contains 
the ball (in $S_0$) centered at 0 and with radius $t_N$.
As in the theorem statement,  $S_0$ is defined
as the vector 
space of functions on $\Theta$ spanned by 
$\{ f(x, \cdot); x \in \mathcal X\}$.
This boils down to bounding
\begin{align*}
     \Pr\left( \inf_{a \in S_0; \|a\|=1} \max_m \langle a, f_m - \bar f \rangle \geq t_N\right)
\end{align*} 
from below, since on this event any point in the ball will be in the convex hull. Moreover, with probability $\geq 1- \delta$, $\widetilde{f}_N$ is in this ball, and thus \cref{convHull} will be satisfied.

From \citet[Corollary 1.2]{Boroczky03},  the unit sphere in 
$d$-dimensions can be covered by
\begin{align*}
N_d(\phi) = \frac{C\cdot \cos\phi}{\sin^d\phi} d^{\frac{3}{2}}\log(1+d\cos^2\phi) \leq \phi^{-d}A_d, \quad A_d = C e^{\frac{d}{2}}d^{\frac{3}{2}}\log(1+d)
\end{align*}
balls of radius $0 < \phi \leq \arccos\frac{1}{\sqrt{d+1}}$ 
and centers $(a_i)_{i\leq N_d(\phi)}$, where $C$ is a universal constant. Moreover,
\begin{align*}
 &\inf_{a \in S_0; \|a\|=1} \max_m \langle a, f_m - \bar f \rangle
 \geq 
\min_{i\leq N_d(\phi)} \max_{m=1,\dots, M}  \left[ \langle a_i, f_m - \bar f\rangle - \phi\|f_m - \bar f\|_{L^2(\mu)} \right].
\end{align*}
For any $a \in S_0$ with $\|a\|=1$, by independence, 
\begin{align*}
\Pr &\left(\max_{m=1,\dots, M}  \left[ \langle a, f_m - \bar f\rangle - \phi\|f_m - \bar f \|_{L^2(\mu)} \right] \leq  t_N \right) \\
& =\Pr\left(  \langle a, f_1- \bar f\rangle - \phi\|f_1 - \bar f \|_{L^2(\mu)}\leq  t_N \right)^M. 
\end{align*}
For any $\phi>0$, if $\langle a, f_1- \bar f \rangle  \leq 2 t_N$ and $\|f_1 - \bar f \|_{L^2(\mu)} > t_N/\phi$ then necessarily $  \langle a, f_1- \bar f\rangle - \phi\|f_1 - \bar f \|_{L^2(\mu)}\leq  t_N $ by the triangle inequality. Thus, we can use the union bound to bound the above by
\begin{align*}
\Pr\left(  \langle a, f_1- \bar f\rangle - \phi\|f_1 - \bar f \|_{L^2(\mu)}\leq  t_N \right)^M \leq \left[ \Pr\left(   \langle a, f_1- \bar f \rangle  \leq 2 t_N \right) + \Pr\left(\|f_1 - \bar f \| > t_N/\phi \right) \right]^M  
\end{align*}
We have, for all $a\in S_0$ with $\|a\|=1$,
\begin{align*}
    \Pr( \langle a, f_1- \bar f \rangle  > 2t_N ) &\geq \inf_{a \in S_0; \|a\|=1}\Pr \left(\langle a, f_1 -\bar f\rangle > 2t_N \right) \\
    &= J(\delta),
\end{align*}
using the definition of $J(\delta)$ as given in the statement of \cref{thm:optimal_kl:finitd}. Thus,
\begin{align*}
    \Pr( \langle a, f_1- \bar f \rangle  \leq 2t_N ) \leq 1-J(\delta).
\end{align*}
Furthermore, choosing $\phi^2 = J(\delta)/(4 N\delta)$,  we have by Chebychev's inequality that
\begin{align*}
    \Pr\left(\|f_1 - \bar f \| > t_N/\phi \right) \leq J(\delta)/2.
\end{align*}
Finally we obtain, using the union bound and the above results, that 
\begin{align*}
\Pr\left( \inf_{a \in S_0; \|a\|=1} \max_m \langle a, f_m - \bar f\rangle  < t_N  \right) &\leq \Pr\left( \min_{i\leq N_d(\phi)} \max_{m=1,\dots, M}  \left[ \langle a_i, f_m - \bar f\rangle - \phi\|f_m - \bar f\|_{L^2(\mu)} \right]< t_N\right) \\
&\leq \sum_{i\leq N_d(\phi)} \Pr\left( \max_{m=1,\dots, M}  \left[ \langle a_i, f_m - \bar f\rangle - \phi\|f_m - \bar f\|_{L^2(\mu)} \right]< t_N\right)  \\
&\leq \sum_{i\leq N_d(\phi)} \left[ \Pr\left(   \langle a_i, f_1- \bar f \rangle  \leq 2 t_N \right) + \Pr\left(\|f_1 - \bar f \| > t_N/\phi \right) \right]^M   \\
&\leq \sum_{i\leq N_d(\phi)}  (1 - J(\delta) + J(\delta)/2)^M   \\
&= N_d(\phi) (1 - J(\delta)/2)^M .
\end{align*}
For $M$ such that $ M J(\delta) \geq  4\log\left(N_d(\phi)\right) $, we then have that
\begin{align*}
    N_d(\phi) (1 - J(\delta)/2)^M &\leq e^{M J(\delta)/4}e^{M\log(1 - J(\delta)/2)} \\
    &\leq e^{M J(\delta)/4}e^{-MJ(\delta)/2} = e^{-M J(\delta)/4}.
\end{align*}
Using the union bound, we can therefore see that 
\cref{convHull} holds with probability 
$\geq 1 - \delta - e^{-M J(\delta)/4}$.

Noting that $N_d(\phi) \leq \phi^{-d}A_d$,
a sufficient condition on $M$ for this to 
  hold is that 
\begin{align*}
    M &\geq  \frac{4}{J(\delta) }\log\left(\phi^{-d}A_d\right) \\
    &=  \frac{4}{J(\delta) }\log\left(\left(\frac{J(\delta)}{4 N\delta}\right)^{-d/2}A_d\right) \\
    &=  \frac{2d}{J(\delta) }\left[\log N + \log\left(\frac{4 \delta}{J(\delta)}\right) + \log\left(A_d^{2/d}\right)\right] \\
    &\geq  \frac{2d}{J(\delta) }\left[\log N + \log\left(\frac{4 \delta}{J(\delta)}\right) + C_1\right],
\end{align*}
where $C_1$ is a constant such that $C_1 \leq\log\left(A_d^{2/d}\right)$. 
This arrives at 
the condition on $M$ in the theorem statement, 
and thus completes the proof.
\end{proof}

\subsection{Proof of \cref{thm:optimal_kl}}
\label{app:optimal_kl_proof}
\textit{Roadmap.}
\begin{enumerate}
    \item The goal of this proof is to upper bound \cref{KL:expression}. We start by using a substitution to find an initial upper bound in the form of an integral.
    \item Next, we split this integral up into two different terms, which we call $I_{\epsilon,1}$ and $I_{\epsilon,2}$, that together sum to the full integral as shown in \cref{KL:split}. These terms can both be expressed as fractions, and share a common denominator. We will then find upper bound for these two terms separately.
    \item We start by targeting $I_{\epsilon,1}$, and in particular upper bounding its numerator. We do this across a series of steps which provide sequential upper bounds. The aim is to find a final upper bound which is simpler, while retaining the necessary asymptotic properties. The final upper bound is given in \cref{I1_bound}.
    \item As part of this process, we assume that a certain set of events occur, and lower bound the probability that they do. These probability bounds will be incorporated into the high probability bound for the final result.
    \item Obtaining this upper bound on the numerator of $I_{\epsilon,1}$ also requires the introduction a condition similar to the one that was central to the proof of \cref{thm:optimal_kl:finitd}. This condition is given in \cref{convexf0}.
    \item Next, we lower bound the common denominator of $I_{\epsilon,1}$ and $I_{\epsilon,2}$. This involves a series of steps similar to those involved in upper bounding the numerator of $I_{\epsilon,1}$, and again we need to assume a set of events occurring. As before, we lower bound the probabilities that they do occur, and these probability bounds will be incorporated into the high probability bound for the final result. The final lower bound is given in \cref{denom_lower}.
    \item Similarly, we need to introduce two more important conditions that are required for our lower bound on the denominator to hold. These are given BY \cref{convexf1,convexf2}.
    \item So far, we have obtained in \cref{I1} an upper bound on $I_{\epsilon,1}$, which holds with a probability that we have obtained a lower bound for, under certain specific conditions that we specify in \cref{,eq:conditions1,eq:conditions2,eq:conditions3}.
    \item The next step is to upper bound $I_{\epsilon,2}$, and in particular its numerator (since we have already lower bounded its denominator). We do this using similar techniques to before, which introduces additional terms into the final high probability bounds. The final upper bound is given in \cref{I2}.
    \item This gives us an upper bound on our original target (i.e. \cref{KL:expression}), which holds with a probability that we have lower bounded, and assuming that the conditions \cref{,eq:conditions1,eq:conditions2,eq:conditions3} hold. 
    \item The final step is to lower bound the probability that \cref{,eq:conditions1,eq:conditions2,eq:conditions3} all hold. This can be done in the same way as in the proof for \cref{thm:optimal_kl:finitd}, and we omit the full details.
    \item Incorporating the above probability lower bounds, we finally have our desired upper bound for the original target, and a lower bound for the probability that this bound holds. This is what we refer to as our ``high probability bound'' in this roadmap.
\end{enumerate}
\begin{proof}
By the concavity and differentiability of $f (x, \cdot)$ we have 
$f(x; \theta) \leq f(x; \theta_0) + \nabla_\theta f (x; \theta_0)^T(\theta - \theta_0) - L_0 \|\theta - \theta_0\|^2 $.
Set $S_N(\theta_0) =\sum_n \nabla_\theta f_n (\theta_0)/\sqrt{N}$ 
and recall that  by definition of 
$\theta_0$ as the maximizer of 
$\bar{f} = \mathbb{E}_p(f_1(\theta))$,
$\mathbb{E}_p\left( \nabla_\theta f_1 (\theta_0)\right)=0$.

Starting from \cref{KL:expression}, we first use the 
fact that $\widetilde{f}_N (\cdot ) = \sum_{n=1}^N [f_n - \bar f](\cdot ) /N$
to rewrite
\begin{align*}
&\widetilde{f}_N - \sum_m v_m (f_m-\bar f) - \mathbb{E}_{w(v),t}[\widetilde{f}_N - \sum_m v_m (f_m-\bar f)] \\
=& 
\frac{\sum_n f_n(\theta) -f_n(\theta_0) }{N}   - \sum_m v_m (f_m(\theta) -f_m(\theta_0))
 -
 \mathbb{E}_{w(v),t}\left[\frac{\sum_n f_n(\theta) -f_n(\theta_0) }{N}   - \sum_m v_m (f_m(\theta) -f_m(\theta_0))\right].
\end{align*}
Substituting this into \cref{KL:expression}, and using
the fact that, for any random variable $X$,
$\mathbb{E}\left[\left(X - \mathbb{E}(X)\right)^2\right] = \mathbb{E}\left(X^2\right) - \mathbb{E}\left(X\right)^2 \leq  \mathbb{E}\left(X^2\right) $,
we can upper bound \cref{KL:expression} by

\begin{align*}
 \mathrm{KL}(\pi_{w(v)}||\pi) \leq 4 N^2\int_0^1 (1-t)  \mathbb{E}_{w(v),t} \left[ \left(\frac{1}{N}\sum_n f_n(\theta) -f_n(\theta_0)   - \sum_m v_m (f_m(\theta) -f_m(\theta_0))\right)^2\right] \mathrm{d}t .
\end{align*} 

We now decompose, for each $t\in (0,1)$,  
the integral over $\Theta$ in 
the above into an integral over
$B_\epsilon = \{ \|\theta-\theta_0\|\leq \epsilon \}$
and an integral over $B_\epsilon^c$. Here, 
$0< \epsilon <\epsilon_0$, where $\epsilon_0$ is
as defined in the assumptions of \cref{thm:optimal_kl}. 
We can rewrite $\mathbb{E}_{w(v),t}(h)$ as 

\begin{align*}
    \mathbb{E}_{w(v),t} (  h ) &=  \frac{ \int_\Theta h(\theta) e^{ \sum_n f_n(\theta) - N(1-t)  [ \widetilde{f}_N(\theta) - \sum_m v_m (f_m(\theta)-\bar{f}(\theta)) ]}\pi_0(\theta)d\theta   }{ \int_\Theta e^{ \sum_n f_n(\theta) - N(1-t)  [ \widetilde{f}_N(\theta) - \sum_m v_m (f_m(\theta)-\bar{f}(\theta)) ]}\pi_0(\theta)d\theta } \\
    &=  \frac{ \int_\Theta h(\theta) e^{ t\sum_n f_n(\theta) + N(1-t)\sum_m v_m f_m(\theta)}\pi_0(\theta)d\theta   }{ \int_\Theta e^{ t\sum_n f_n(\theta) + N(1-t)\sum_m v_m f_m(\theta)}\pi_0(\theta)d\theta } \times \frac{e^{ -t\sum_n f_n(\theta_0) - N(1-t)\sum_m v_m f_m(\theta_0)}  }{e^{ -t\sum_n f_n(\theta_0) - N(1-t)\sum_m v_m f_m(\theta_0)} } \\
    &= \frac{ \int_\Theta h(\theta) e^{ t \sum_n( f_n(\theta)- f_n(\theta_0) )  + N(1-t)  \sum_m v_m (f_m(\theta)-f_m(\theta_0)) ]}\pi_0(\theta)d\theta   }{ \int_\Theta e^{ t \sum_n( f_n(\theta)- f_n(\theta_0) )  + N(1-t)  \sum_m v_m (f_m(\theta)-f_m(\theta_0)) ]} \pi_0(\theta)d\theta } \\
    &= \frac{ \int_{B_{\epsilon}^c} h(\theta) e^{ t \sum_n( f_n(\theta)- f_n(\theta_0) )  + N(1-t)  \sum_m v_m (f_m(\theta)-f_m(\theta_0)) ]}\pi_0(\theta)d\theta}{ \int_\Theta e^{ t \sum_n( f_n(\theta)- f_n(\theta_0) )  + N(1-t)  \sum_m v_m (f_m(\theta)-f_m(\theta_0)) ]} \pi_0(\theta)d\theta } \\
    &\qquad + \frac{\int_{B_{\epsilon}} h(\theta) e^{ t \sum_n( f_n(\theta)- f_n(\theta_0) )  + N(1-t)  \sum_m v_m (f_m(\theta)-f_m(\theta_0)) ]}\pi_0(\theta)d\theta }{ \int_\Theta e^{ t \sum_n( f_n(\theta)- f_n(\theta_0) )  + N(1-t)  \sum_m v_m (f_m(\theta)-f_m(\theta_0)) ]} \pi_0(\theta)d\theta } \\
    &\coloneqq I_{\epsilon,1}(h) + I_{\epsilon,2}(h) 
\end{align*}
Writing $I_{\epsilon,1} = I_{\epsilon,1}\left( \left(\frac{1}{N}\sum_n f_n(\theta) -f_n(\theta_0)   - \sum_m v_m (f_m(\theta) -f_m(\theta_0))\right)^2\right)$, 
and similarly for $I_{\epsilon,2}$, we 
can thus upper bound \cref{KL:expression} by 
\begin{align}\label{KL:split} 
KL(\pi_{w(v)}||\pi) &\leq 4N^2\int_0^1 (1-t) \left(I_{\epsilon,1} + I_{\epsilon,2} \right) \mathrm{d}t 
\end{align}
We first prove that the integral $I_{\epsilon,1}$
over $B_\epsilon^c $ is small. 
Using the concavity bound for $f_n$, and 
recalling that $\sum_m v_m = 1$, we have that 
\begin{align*}
t \sum_n( f_n(\theta)&- f_n(\theta_0) )  + N(1-t)  \sum_m v_m (f_m(\theta)-f_m(\theta_0)) \\
 &\leq t \sum_n\left( \nabla_\theta f_n(\theta_0)^T(\theta - \theta_0) - L_0 \|\theta - \theta_0\|^2\right)  + N(1-t)  \sum_m v_m \left(\nabla_\theta f_m(\theta_0)^T(\theta - \theta_0) - L_0 \|\theta - \theta_0\|^2\right) \\
  &=
t \left[\sqrt{N}S_N(\theta_0)^T(\theta - \theta_0) -NL_0\|\theta - \theta_0\|^2 \right] + N(1-t) \left[ \sum_m v_m \nabla_\theta f_m(\theta_0)^T(\theta-\theta_0) - L_0\|\theta - \theta_0\|^2 \right].
 \end{align*}
If
\begin{equation}\label{convexf0} 
\sum_m v_m \nabla_\theta f_m(\theta_0) = \frac{ S_N(\theta_0)  }{ \sqrt{N} }
\end{equation}
then 
\begin{align*}
t \sum_n( f_n(\theta)- f_n(\theta_0) )  &+ N(1-t)  \sum_m v_m (f_m(\theta)-f_m(\theta_0)) \leq 
 \sqrt{N}S_N(\theta_0)^T(\theta - \theta_0) -NL_0\|\theta - \theta_0\|^2. 
 \end{align*}
Using this bound, we can bound $I_{\epsilon, 1}$ by
\begin{align*}
&I_{\epsilon, 1} \\
& : = \frac{ \int_{B_\epsilon^c}   \left[\frac{1}{N}\sum_n f_n(\theta) -f_n(\theta_0)   - \sum_m v_m (f_m(\theta) -f_m(\theta_0))\right]^2 e^{ t \sum_n( f_n(\theta)- f_n(\theta_0) )  + N(1-t)  \sum_m v_m (f_m(\theta)-f_m(\theta_0)) ]}\pi_0(\theta)d\theta   }{ \int_\Theta e^{ t \sum_n( f_n(\theta)- f_n(\theta_0) )  + N(1-t)  \sum_m v_m (f_m(\theta)-f_m(\theta_0)) ]} \pi_0(\theta)d\theta } \\
&\leq \frac{\int_{B_\epsilon^c}  \left[\frac{1}{N}\sum_n f_n(\theta) -f_n(\theta_0)   - \sum_m v_m (f_m(\theta) -f_m(\theta_0))\right]^2 e^{  \sqrt{N}S_N(\theta_0)^T(\theta - \theta_0) -NL_0\|\theta - \theta_0\|^2 }\pi_0(\theta)d\theta   }{ \int_\Theta e^{ t \sum_n( f_n(\theta)- f_n(\theta_0) )  + N(1-t)  \sum_m v_m (f_m(\theta)-f_m(\theta_0)) ]} \pi_0(\theta)d\theta  }.
\end{align*}
Completing the square in the exponent of the
numerator, we can rewrite it as 
\begin{align*}
    &\sqrt{N}S_N(\theta_0)^T(\theta - \theta_0) -NL_0\|\theta - \theta_0\|^2 \\
    &= -NL_0\left(\|\theta - \theta_0\|^2 - \frac{S_N(\theta_0)^T(\theta - \theta_0)}{\sqrt{N}L_0} \right) \\
    &= -NL_0\left\|(\theta - \theta_0) - \frac{S_N(\theta_0)}{2\sqrt{N}L_0} \right\|^2 +  \frac{\|S_N(\theta_0)\|^2}{ 4L_0}
\end{align*}
We now assume that 
$\|S_N(\theta_0)\| \leq \left(2 - \sqrt{2}\right) \epsilon L_0\sqrt{N}$. 
By Markov's inequality, 
\begin{align*}
    \Pr\left(\|S_N(\theta_0)\| \geq \left(2 - \sqrt{2}\right) \epsilon L_0\sqrt{N}\right) &\leq \frac{\mathbb{E}_p\left(\|S_N(\theta_0)\|^2\right)}{\left(2 - \sqrt{2}\right)^2\epsilon^2NL_0^2} \\
    &\leq3\frac{\mathbb{E}_p\left(\|\sum_n \nabla_\theta f_n (\theta_0)\|^2\right)}{\epsilon^2N^2L_0^2} \\
    &=3\frac{\sum_n\mathbb{E}_p\left(\| \nabla_\theta f_n (\theta_0)\|^2\right)}{\epsilon^2N^2L_0^2} \\
    &=3\frac{\mathbb{E}_p\left(\| \nabla_\theta f_1 (\theta_0)\|^2\right)}{\epsilon^2NL_0^2},
\end{align*}
and thus the probability that 
$\|S_N(\theta_0)\| \leq \left(2 - \sqrt{2}\right)\epsilon L_0\sqrt{N}$ 
is bounded from below by 
$1- 3\mathbb{E}_p( \|\nabla_\theta f_1(\theta_0)\|^2)/(L_0^2N \epsilon^2)$. 
Here, and throughout, we assume that $N$
is large enough such that the relevant 
probabilities we use for our high-probability
bounds are $\geq 0$.
When this event occurs, 
\begin{align*}
    \left\|(\theta - \theta_0) - \frac{S_N(\theta_0)}{2\sqrt{N}L_0} \right\| &\geq \left\|\theta - \theta_0 \right\| - \left\| \frac{S_N(\theta_0)}{2\sqrt{N}L_0} \right\|\\
    &\geq \left\|\theta - \theta_0 \right\| - \frac{\left(2 - \sqrt{2}\right)\epsilon L_0\sqrt{N}}{2\sqrt{N}L_0} \\
    &\geq \left\|\theta - \theta_0 \right\| -  \left(\frac{2 - \sqrt{2}}{2}\right)\epsilon\\\
     &\geq \left\|\theta - \theta_0 \right\| -  \left(\frac{2 - \sqrt{2}}{2}\right)\left\|\theta - \theta_0\right\| \qquad \text{ on $B_{\epsilon}^c$}\\
    &= \frac{1}{\sqrt{2}}\|\theta - \theta_0\|.
\end{align*}
Thus
\begin{align*}
    \left\|(\theta - \theta_0) - \frac{S_N(\theta_0)}{2\sqrt{N}L_0} \right\|^2 \geq  \frac{1}{2}\|\theta - \theta_0\|^2,
\end{align*}
since both sides are positive numbers.

For ease of notation, we define $\Delta_n \coloneqq f_n(\theta) - f_n(\theta_0)$. 
We can then bound $I_{\epsilon,1}$ by
\begin{align*}
I_{\epsilon,1} &\leq \frac{e^{  \frac{\|S_N(\theta_0)\|^2}{ 4L_0} } \int_{B_\epsilon^c}  \left[ \frac{1}{N}\sum_n \Delta_n - \sum_m v_m \Delta_m
 \right]^2 e^{ -NL_0\left\|(\theta - \theta_0) - \frac{S_N(\theta_0)}{2\sqrt{N}L_0} \right\|^2}\pi_0(\theta)d\theta   }{ \int_\Theta e^{ t \sum_n( f_n(\theta)- f_n(\theta_0) )  + N(1-t)  \sum_m v_m (f_m(\theta)-f_m(\theta_0)) ]} \pi_0(\theta)d\theta  }\\
&\leq \frac{e^{\epsilon^2NL_0/8} \int_{B_\epsilon^c}  \left[ \frac{1}{N}\sum_n \Delta_n - \sum_m v_m \Delta_m
 \right]^2 e^{ -NL_0\|\theta - \theta_0\|^2/2 }\pi_0(\theta)d\theta   }{ \int_\Theta e^{ t \sum_n( f_n(\theta)- f_n(\theta_0) )  + N(1-t)  \sum_m v_m (f_m(\theta)-f_m(\theta_0)) ]} \pi_0(\theta)d\theta  }.
\end{align*}
To further simplify the upper bound on the numerator,
we define $\bar{\Delta} \coloneqq \mathbb{E}_p(\Delta_n)$ and
consider the event where
\begin{align}\label{eq:I_1_numerator_bound}
     \int_{B_\epsilon^c}  \left[ \frac{1}{N}\sum_n \Delta_n - \sum_m v_m \Delta_m
 \right]^2 e^{   -NL_0\|\theta - \theta_0\|^2/2 }\pi_0(\theta)d\theta \leq e^{-NL_0\epsilon^2/4}  \int_\Theta \mathbb{E}_p(\Delta_1 - \bar \Delta)^2(\theta)\pi_0(\theta)d\theta.
\end{align}
The probability of \cref{eq:I_1_numerator_bound} holding can be bounded from below
by using Markov's inequality, 
\begin{align}
\Pr  &\left( \int_{B_\epsilon^c}  \left[ \frac{1}{N}\sum_n \Delta_n - \sum_m v_m \Delta_m
 \right]^2 e^{   -NL_0\|\theta - \theta_0\|^2/2 }\pi_0(\theta)d\theta > e^{-NL_0\epsilon^2/4}  \int_\Theta \mathbb{E}_p(\Delta_1 - \bar \Delta)^2(\theta)\pi_0(\theta)d\theta\right)   \notag\\
&\leq \frac{e^{NL_0\epsilon^2/4} \mathbb{E}_p\left(\int_{B_\epsilon^c}  \left[ \frac{1}{N}\sum_n \Delta_n - \sum_m v_m \Delta_m
 \right]^2 e^{   -NL_0\|\theta - \theta_0\|^2/2 }\pi_0(\theta)d\theta \right)}{ \int_\Theta \mathbb{E}_p(\Delta_1 - \bar \Delta)^2(\theta)\pi_0(\theta)d\theta}.\label{eq:pnumbound}
\end{align}
We now use the fact that $(a-b)^2 \leq 2a^2 + 2b^2$ to bound
\begin{align*}
\left[ \frac{1}{N}\sum_n \Delta_n - \sum_m v_m \Delta_m
 \right]^2 & = \left[ \frac{ \sum_n (\Delta_n - \bar \Delta) }{ N} - \sum_m v_m ( \Delta_m-\bar \Delta) \right]^2\\
&\leq 2 \left(\frac{ \sum_n (\Delta_n - \bar \Delta) }{ N} \right)^2 + 2\left( \sum_m v_m ( \Delta_m-\bar \Delta) \right)^2\\
&\leq 2 \left(\frac{ \sum_n (\Delta_n - \bar \Delta) }{ N} \right)^2 + 2\left( \max_M( \Delta_m-\bar \Delta)\sum_m v_m \right)^2\\
& \leq 2 \left(\frac{ \sum_n (\Delta_n - \bar \Delta) }{ N} \right)^2 + 2\max_M( \Delta_m-\bar \Delta)^2.
\end{align*}
Combined with the fact that $\|\theta - \theta_0\| \geq \epsilon$ on $B_{\epsilon}^c$,
we can bound \cref{eq:pnumbound} above by
\begin{align}\label{eq:I_1_numerator_bound2}
    \leq \frac{2\left(\int_{B_\epsilon^c} \mathbb{E}_p \left[ \left(\frac{ \sum_n (\Delta_n - \bar \Delta) }{ N} \right)^2 + \max_M( \Delta_m-\bar \Delta)^2 \right] e^{   -NL_0\|\theta - \theta_0\|^2/4 }\pi_0(\theta)d\theta \right)}{ \int_\Theta \mathbb{E}_p(\Delta_1 - \bar \Delta)^2(\theta)\pi_0(\theta)d\theta}.
\end{align} 
Then, by the definition of $\bar{\Delta}$, and the
independence of the $f_n$ and therefore $\Delta_n$, 
\begin{align*}
    \mathbb{E}_p \left[ \left(\frac{ \sum_n (\Delta_n - \bar \Delta) }{ N} \right)^2 + \max_M( \Delta_m-\bar \Delta)^2 \right] 
    &= \frac{1}{N^2} \sum_n \mathbb{E}_p\left((\Delta_n - \bar \Delta)^2\right) + \mathbb{E}_p \left[\max_M( \Delta_m-\bar \Delta)^2 \right] \\
    &\leq \frac{1}{N} \mathbb{E}_p\left((\Delta_1 - \bar \Delta)^2\right) + \mathbb{E}_p \left[\sum_{m=1}^M( \Delta_m-\bar \Delta)^2 \right] \\
    &= \frac{1}{N} \mathbb{E}_p\left((\Delta_1 - \bar \Delta)^2\right) + M \mathbb{E}_p\left((\Delta_1 - \bar \Delta)^2\right).
\end{align*}
Thus, \cref{eq:I_1_numerator_bound2} can be upper
bounded by
\begin{align*}
&\leq \frac{2 \int_{B_\epsilon^c}e^{   -NL_0\|\theta - \theta_0\|^2/4}  \mathbb{E}_p(\Delta_1 - \bar \Delta)^2(\theta)\left(\frac{1}{N} +M\right) \pi_0(\theta)d\theta}{\int_\Theta \mathbb{E}_p(\Delta_1 - \bar \Delta)^2(\theta)\pi_0(\theta)d\theta} \\
&\leq \frac{2 \int_{B_\epsilon^c}e^{   -NL_0\epsilon^2/4}  \mathbb{E}_p(\Delta_1 - \bar \Delta)^2(\theta)\left(2M\right) \pi_0(\theta)d\theta}{\int_\Theta \mathbb{E}_p(\Delta_1 - \bar \Delta)^2(\theta)\pi_0(\theta)d\theta}\\
&\leq 4Me^{-NL_0\epsilon^2/4}.
\end{align*}
This tells us that the probability of \cref{eq:I_1_numerator_bound} holding can be bounded from below by $1 - 4Me^{-NL_0\epsilon^2/4}$.
We have now obtained an upper bound on the 
numerator of $I_{\epsilon,1}$ that holds with high probability. 
To summarize the result so far:
if
$\sum_m v_m \nabla_\theta f_m(\theta_0) = \frac{ S_N(\theta_0)  }{ \sqrt{N} }$,
we have that for any $0 < \epsilon < \epsilon_0$,
\begin{align}\label{I1_bound} 
I_{\epsilon,1} \leq \frac{e^{ -\epsilon^2NL_0/8}  \int_\Theta \mathbb{E}_p(\Delta_1 - \bar \Delta)^2(\theta)\pi_0(\theta)d\theta}{ \int_\Theta e^{ t \sum_n( f_n(\theta)- f_n(\theta_0) )  + N(1-t)  \sum_m v_m (f_m(\theta)-f_m(\theta_0)) } \pi_0(\theta)d\theta  }.
\end{align}
with probability at least
\begin{align*}
1 - 4Me^{-NL_0\epsilon^2/4} - 3\frac{\mathbb{E}_p\left(\| \nabla_\theta f_1 (\theta_0)\|^2\right)}{\epsilon^2NL_0^2}.
\end{align*}

We continue 
by finding a lower bound for the denominator of $I_{\epsilon,1}$. 
We decompose $f_n $ into $f_n^{(1)}$ 
and $f_n^{(2)}$ and assume that
 \begin{equation}\label{convexf1}
    \sum_m v_m (f_m^{(1)}(\theta)-f_m^{(1)}(\theta_0)) - \frac{ \sum_n( f_n^{(1)}(\theta)- f_n^{(1)}(\theta_0) )}{N}  =0.
\end{equation}
Then
\begin{align} 
 &t \sum_n( f_n(\theta)- f_n(\theta_0) )  + N(1-t) \left[ \sum_m v_m (f_m(\theta)-f_m(\theta_0)) \right] \notag\\
 &= \sum_n( f_n(\theta)- f_n(\theta_0) )  + N(1-t) \left[ \sum_m v_m (f_m(\theta)-f_m(\theta_0)) - \frac{\sum_n( f_n(\theta)- f_n(\theta_0) ) }{N} \right] \notag\\
 &= \sum_n( f_n(\theta)- f_n(\theta_0) )  + N(1-t) \left[ \sum_m v_m (f_m^{(2)}(\theta)-f_m^{(2)}(\theta_0)) - \frac{\sum_n( f_n^{(2)}(\theta)- f_n^{(2)}(\theta_0) ) }{N} \right] \notag\\
 &= \sum_n( f_n(\theta)- f_n(\theta_0) )  + N(1-t)\left[  \sum_m v_m f_m^{(2)}(\theta)  -  \frac{ \sum_n f_n^{(2)}(\theta) }{N} \right], \label{eq:fn2lower}
\end{align}
using \cref{convexf1} and the fact that
$f_n^{(2)}(\theta_0) = 0$ by assumption \textbf{A2}. 
For any $0 < \epsilon' < \epsilon_0$, 
and $\|\theta - \theta_0\|\leq \epsilon'$,
we use assumption \textbf{A2} again, along 
with the fact that 
$a - b \geq -|a| -|b|$, 
to bound \cref{eq:fn2lower} from below by
\begin{align}
 & \geq \sum_n( f_n(\theta)- f_n(\theta_0) )  - N(1-t)\left|  \sum_m v_m f_m^{(2)}(\theta)  -  \frac{ \sum_n f_n^{(2)}(\theta) }{N} \right| \notag\\
 & \geq \sum_n( f_n(\theta)- f_n(\theta_0) )  - N(1-t)r(\epsilon')\|\theta - \theta_0\|^2
\left[ \left|  \sum_m v_m R(x_m) \right| + \left| \frac{ \sum_n R(x_n)}{N} \right|\right] \notag\\
 &\geq \sum_n( f_n(\theta)- f_n(\theta_0) )  \notag\\
 &~~~~~~~~ -  N(1-t)r(\epsilon')\|\theta - \theta_0\|^2
\left[ 2 \mathbb{E}_p(R(X)) + \left|\sum_m v_m (R(x_m)-\mathbb{E}_p(R(X))) \right|
  + \frac{ \left|\sum_n (R(x_n)-\mathbb{E}_p(R(X))) \right| }{N} \right] \label{eq:epRetc}
\end{align}
using the fact that $R(x)>0$ by assumption. 
Next, we assume that   
\begin{equation}\label{convexf2} 
\left|\sum_m v_m (R(x_m)-\mathbb{E}_p(R(X)) \right| \leq 1 \qquad \text{and}\qquad \frac{ \left|\sum_n (R(x_n)-\mathbb{E}_p(R(X)) \right| }{N}  \leq 1.
\end{equation}
We can bound the probability of the second inequality holding from below by 
$1 - \mathrm{Var}_p(R(X))/N$ via Chebychev's inequality:
\begin{align*}
    \Pr\left(\left|\frac{1}{N}\sum_n R(x_n)-\mathbb{E}_p(R(X)) \right|  \geq \frac{\sqrt{\mathrm{Var}_p\left(\frac{1}{N}\sum_n R(x_n)\right)}}{\sqrt{\mathrm{Var}_p\left(\frac{1}{N}\sum_n R(x_n)\right)}}\right) \leq \mathrm{Var}_p\left(\frac{1}{N}\sum_n R(x_n)\right) = \frac{1}{N}\mathrm{Var}_p(R(X)).
\end{align*}     
Substituting the two assumptions from \cref{convexf2} into \cref{eq:epRetc} above, we obtain that
\begin{align*} 
 t \sum_n( f_n(\theta)&- f_n(\theta_0) )  + N(1-t)  \sum_m v_m (f_m(\theta)-f_m(\theta_0))  \\
 & \geq \sum_n( f_n(\theta)- f_n(\theta_0) )  - 2 Nr(\epsilon')\|\theta - \theta_0\|^2
\left[  \mathbb{E}_p(R(X))+1 \right]\\
&= \sum_n( f_n(\theta)- f_n(\theta_0) )  - \gamma Nr(\epsilon')\|\theta - \theta_0\|^2, \quad \text{where}\quad \gamma \coloneqq 2\left[  \mathbb{E}_p(R(X))+1 \right].
\end{align*}
Then we can bound the denominator of $I_{\epsilon,1}$ from below by
\begin{align}
     &\int_\Theta e^{ t \sum_n( f_n(\theta)- f_n(\theta_0) )  + N(1-t)  \sum_m v_m (f_m(\theta)-f_m(\theta_0)) ]} \pi_0(\theta)d\theta  \notag\\
     \geq& \int_{B_{\epsilon'} } e^{ t \sum_n( f_n(\theta)- f_n(\theta_0) )  + N(1-t)  \sum_m v_m (f_m(\theta)-f_m(\theta_0)) ]} \pi_0(\theta)d\theta  \notag\\
     \geq& \int_{B_{\epsilon'} }  e^{\sum_n f_n(\theta)-f_n(\theta_0)- \gamma Nr(\epsilon')\|\theta - \theta_0\|^2} \pi_0(\theta)d\theta \notag\\
\geq&  \pi_{0,\inf}\int_{B_{\epsilon'} }  e^{\sum_n f_n(\theta)-f_n(\theta_0)- \gamma Nr(\epsilon')\|\theta - \theta_0\|^2} d\theta\notag\\
 \geq&
\pi_{0,\inf} e^{-t_a(\epsilon')}\int_{B_{\epsilon'}} \mathrm{1}_{ \sum_n f_n(\theta)-f_n(\theta_0) \geq N(\bar f(\theta) - \bar f(\theta_0) ) -t_a } e^{- L_1'(\epsilon')N \| \theta - \theta_0\|^2 } d\theta\label{eq:ie1lbd}\\
&\text{ where } \quad \pi_{0,\inf} \coloneqq \inf_{\theta\in B_{\epsilon'}}\pi_0(\theta), \quad L_1'(\epsilon') \coloneqq L_1+\gamma r(\epsilon'), \quad\text{and}\quad t_a(\epsilon') \coloneqq \left(\frac{ D L_2}{L_1'(\epsilon')a}\right)^{1/2}\text{ for } a>0,\notag
\end{align}
where the last line follows from assumption \textbf{A3} and the definition of $L_1'(\epsilon')$. 
The next goal is to bound \cref{eq:ie1lbd} from below by
\begin{align}\label{eq:D_N_lower_bound}
& \geq  \frac{ \pi_{0,\inf} e^{-t_a(\epsilon') } }{2 } \int_{B_{\epsilon'}}  e^{- L_1'(\epsilon')N \| \theta - \theta_0\|^2 } d\theta
\end{align} 
with high probability. To begin, by Markov's inequality, 
\begin{align}
\Pr &\left( \int_{B_{\epsilon'} } \mathrm{1}_{ \sum_n f_n(\theta)-f_n(\theta_0) \geq N(\bar f(\theta) - \bar f(\theta_0) ) -t_a(\epsilon') } e^{- L_1'(\epsilon')N \| \theta - \theta_0\|^2 } d\theta <  \frac{  \int_{B_{\epsilon'} }  e^{- L_1'(\epsilon')N \| \theta - \theta_0\|^2 } d\theta }{2 } \right) \notag \\
& = 
\Pr \left( \int_{B_{\epsilon'} } \mathrm{1}_{ \sum_n f_n(\theta)-f_n(\theta_0) < N(\bar f(\theta) - \bar f(\theta_0) ) -t_a(\epsilon')} e^{- L_1'(\epsilon')N \| \theta - \theta_0\|^2 } d\theta >  \frac{  \int_{B_{\epsilon'}}  e^{- L_1'(\epsilon')N \| \theta - \theta_0\|^2 } d\theta }{2 } \right) \notag \\
& \leq \frac{ 2 \int_{B_{\epsilon'}  } \Pr \left( \sum_n f_n(\theta)-f_n(\theta_0) < N(\bar f(\theta) - \bar f(\theta_0) ) -t_a(\epsilon')\right)  e^{- L_1'(\epsilon')N \| \theta - \theta_0\|^2 } d\theta }{\int_{B_{\epsilon'} }  e^{- L_1'(\epsilon')N \| \theta - \theta_0\|^2 } d\theta }. \label{eq:markabound}
\end{align}
The first equality here follows from the 
fact that 
\begin{align*}
    &\mathrm{1}_{ \sum_n f_n(\theta)-f_n(\theta_0) \geq N(\bar f(\theta) - \bar f(\theta_0) ) -t_a(\epsilon') } e^{- L_1'(\epsilon')N \| \theta - \theta_0\|^2 }\\
    &+ \mathrm{1}_{ \sum_n f_n(\theta)-f_n(\theta_0) < N(\bar f(\theta) - \bar f(\theta_0) ) -t_a(\epsilon')} e^{- L_1'(\epsilon')N \| \theta - \theta_0\|^2 } = e^{- L_1'(\epsilon')N \| \theta - \theta_0\|^2 }.
\end{align*}
Thus we can see that if the integral of one of the terms on the left hand side of the above is less than half the integral of the right hand side, then the integral of the other must be greater than half the integral of the right hand side.

We can apply Markov's inequality again to bound
the probability inside the integral,
\begin{align*}
    &\Pr \left( \sum_n f_n(\theta)-f_n(\theta_0) < N(\bar f(\theta) - \bar f(\theta_0) ) -t_a(\epsilon')\right) \\
    =& \Pr \left( \frac{1}{N}\sum_n f_n(\theta_0)-f_n(\theta) - \mathbb{E}_p\left(\frac{1}{N}\sum_n f_n(\theta_0)-f_n(\theta) \right)> \frac{t_a(\epsilon')}{N}\right) \\
    \leq& \frac{N^2}{t_a(\epsilon')^2}\mathrm{Var}_p\left(\frac{1}{N}\sum_n f_n(\theta_0)-f_n(\theta) \right)\\
    = &\frac{N}{t_a(\epsilon')^2}\mathrm{Var}_p\left( f_1(\theta_0)-f_1(\theta) \right)\\
    \leq& \frac{N}{t_a(\epsilon')^2}L_2\| \theta - \theta_0\|^2,
\end{align*}
using assumption \textbf{A3}. Substituting this into \cref{eq:markabound} yields
\begin{align}
&\frac{ 2 \int_{B_{\epsilon'}  } \Pr \left( \sum_n f_n(\theta)-f_n(\theta_0) < N(\bar f(\theta) - \bar f(\theta_0) ) -t_a(\epsilon')\right)  e^{- L_1'(\epsilon')N \| \theta - \theta_0\|^2 } d\theta }{\int_{B_{\epsilon'} }  e^{- L_1'(\epsilon')N \| \theta - \theta_0\|^2 } d\theta } \notag\\
& \leq \frac{ 2L_2 \int_{B_{\epsilon'}  } N\|\theta - \theta_0\|^2   e^{- L_1'(\epsilon')N \| \theta - \theta_0\|^2 } d\theta }{ t_a(\epsilon')^2\int_{B_{\epsilon'} }  e^{- L_1'(\epsilon')N \| \theta - \theta_0\|^2 } d\theta } \notag\\
& = \frac{  L_2 }{  L_1'(\epsilon') t_a(\epsilon')^2} \frac{ \int_{\|u\| \leq \epsilon'\sqrt{2NL_1'(\epsilon')}  }\|u\|^2   e^{- \| u\|^2/2 } du }{ \int_{\|u\| \leq \epsilon'\sqrt{2NL_1'(\epsilon')}  }   e^{- \| u\|^2/2 } du } \label{eq:markprob2}
\end{align} 
where we make the substitution $u = \sqrt{2NL_1'(\epsilon')} (\theta-\theta_0)$,
and cancel the Jacobian terms in numerator and 
denominator.
If $\epsilon' \geq \sqrt{\frac{D}{2NL'_1(\epsilon')}}$, the integral in the denominator can be 
bounded below by
\begin{align*}
      \int_{\|u\| \leq \epsilon'\sqrt{2NL_1'(\epsilon')}  }   e^{- \| u\|^2/2 } du   &\geq  \left(2\pi\right)^{D/2} \int_{\|u\|^2 \leq D } \left(2\pi\right)^{-D/2}  e^{- \| u\|^2/2 } du \\
      &= \left(2\pi\right)^{D/2}\Pr\left(Y\leq D\right), \qquad Y \sim \chi^2_D \\
      &\geq \frac{1}{2}\left(2\pi\right)^{D/2},
\end{align*}
since the median of a $\chi^2_k$-distribution is $\leq$ $k$. Since $\Theta \subseteq \mathbb{R}^D$, we can upper bound the integral in the numerator by 
\begin{align*}
    \int_{\|u\| \leq \epsilon'\sqrt{2NL_1'(\epsilon')}  }\|u\|^2   e^{- \| u\|^2/2 } du \leq \left(2\pi\right)^{D/2} \int\|u\|^2  \left(2\pi\right)^{-D/2} e^{- \| u\|^2/2 } du = D\left(2\pi\right)^{D/2},
\end{align*}
and thus \cref{eq:markprob2} can be upper bounded by 
\begin{align*}
    \frac{  L_2 }{  L_1'(\epsilon') t_a(\epsilon')^2} \frac{ \int_{\|u\| \leq \epsilon'\sqrt{2NL_1'(\epsilon')}  }\|u\|^2   e^{- \| u\|^2/2 } du }{ \int_{\|u\| \leq \epsilon'\sqrt{2NL_1'(\epsilon')}  }   e^{- \| u\|^2/2 } du } \leq \frac{2L_2D}{L_1'(\epsilon')t_a(\epsilon')^2} = 2a,
\end{align*}    
using the definition of $t_a(\epsilon')$. 
Thus with probability at least $1-2a$,
we can bound \cref{eq:ie1lbd} from below by
\begin{align}\label{denom_lower} 
\text{\cref{eq:ie1lbd}} & \geq  \frac{ \pi_{0,\inf} e^{-t_a(\epsilon') } }{2 } \int_{B_{\epsilon'}}  e^{- L_1'(\epsilon')N \| \theta - \theta_0\|^2 } d\theta \nonumber \\
&\geq \frac{ \pi_{0,\inf} e^{-t_a(\epsilon') } }{2 } \left(\frac{1}{2L_1'(\epsilon')N}\right)^{D/2} \int_{\|u\| \leq \epsilon' \sqrt{2L_1'(\epsilon')N}}  e^{- \| u\|^2/2 } d\theta \nonumber \\
&\geq \frac{ \pi_{0,\inf} e^{-t_a(\epsilon') } }{4 } \left(\frac{\pi}{L_1'(\epsilon')N}\right)^{D/2},
\end{align} 
using the same substitution and $\chi^2$ bound as before. This gives us our final lower bound on the denominator of $I_{\epsilon,1}$.
To summarize the result at this point:
assuming that
\begin{align}
\sum_m v_m \nabla_\theta f_m(\theta_0) &= \frac{ S_N(\theta_0)  }{ \sqrt{N} }\label{eq:conditions1}\\
\sum_m v_m (f_m^{(1)}(\theta)-f_m^{(1)}(\theta_0)) &= \frac{ \sum_n( f_n^{(1)}(\theta)- f_n^{(1)}(\theta_0) )}{N}\label{eq:conditions2}\\ 
\left|\sum_m v_m (R(x_m)-\mathbb{E}_p(R(X)) \right| &\leq 1,\label{eq:conditions3}
\end{align}
we have that for any $a>0$, $\sqrt{\frac{D}{2NL'_1(\epsilon')}} \leq \epsilon' < \epsilon_0$, and $0 \leq \epsilon < \epsilon_0$,
\begin{align}\label{I1}
I_{\epsilon,1} &\leq \frac{e^{-NL_0\epsilon^2/8} \int_\Theta \mathbb{E}_p(\Delta_1 - \bar \Delta)^2(\theta)\pi_0(\theta)d\theta}{ \int_\Theta e^{ t \sum_n( f_n(\theta)- f_n(\theta_0) )  + N(1-t)  \sum_m v_m (f_m(\theta)-f_m(\theta_0)) } \pi_0(\theta)d\theta  } \nonumber \\
&\leq N^{D/2}e^{t_a(\epsilon') -NL_0\epsilon^2/8}\left(\frac{4}{\pi_{0,\inf}} \int_\Theta \mathbb{E}_p(\Delta_1 - \bar \Delta)^2(\theta)\pi_0(\theta)d\theta\left(\frac{L'_1(\epsilon')}{\pi}\right)^{D/2}\right),
\end{align}
with probability at least
\begin{align*}
1 - 4Me^{-NL_0\epsilon^2/4} 
- 3\frac{\mathbb{E}_p\left(\| \nabla_\theta f_1 (\theta_0)\|^2\right)}{\epsilon^2NL_0^2}
- \frac{\mathrm{Var}_p(R(X))}{N} - 2a.
\end{align*}

We now bound $I_{\epsilon,2}$, corresponding to the integral over $B_\epsilon$. Throughout we assume that
the same set of events occur as used in the analysis of $I_{\epsilon,1}$.
Note that the denominator of $I_{\epsilon,2}$ is the same as in $I_{\epsilon,1}$, so we apply the same bound;
here we focus on the numerator.
Since $r(\epsilon) \gamma\leq L_0/2 $ by \textbf{A2}, 
we can obtain the following bound for all $\theta \in B_\epsilon$:
\begin{align*} 
 t \sum_n( f_n(\theta)&- f_n(\theta_0) )  + N(1-t)  \sum_m v_m (f_m(\theta)-f_m(\theta_0))  \\
 & \leq \sum_n( f_n(\theta)- f_n(\theta_0) )  +  N(1-t)r(\epsilon)\|\theta - \theta_0\|^2
\gamma\\
&\leq \sqrt{N}S_N(\theta_0)^T(\theta - \theta_0) -NL_0\|\theta - \theta_0\|^2 +  N(1-t)r(\epsilon)\|\theta - \theta_0\|^2 \gamma\\
&\leq \sqrt{N}S_N(\theta_0)^T(\theta - \theta_0) -\frac{NL_0}{2}\|\theta - \theta_0\|^2 \\
&= -\frac{NL_0}{2}\left(\|\theta - \theta_0\|^2 - \frac{2S_N(\theta_0)^T(\theta - \theta_0)}{\sqrt{N}L_0} \right) \\
&\leq \frac{ \|S_N(\theta_0)\|^2 }{ 2L_0 } -\frac{ NL_0 }{2 } \left\|\theta - \theta_0 -\frac{ S_N(\theta_0)}{ \sqrt{N}L_0 }\right\|^2.
\end{align*}
As before we write, for $\theta \in  B_\epsilon$,
\begin{align*}
\frac{\sum_n f_n(\theta) -f_n(\theta_0) }{N}  & - \sum_m v_m (f_m(\theta) -f_m(\theta_0)) = 
\frac{\sum_n f_n^{(2)}(\theta) -f_n^{(2)}(\theta_0) }{N}   - \sum_m v_m (f_m^{(2)}(\theta) -f_m^{(2)}(\theta_0)),
 \end{align*}
so the right hand side is bounded above by 
\begin{align*}
r(\epsilon) \|\theta-\theta_0\|^2 &\left[ 2 \mathbb{E}_p(R(X)) +  \frac{ |\sum_n (R(x_n)-\mathbb{E}_p(R(X))| }{N}  +\left|\sum_m v_m((R(x_m)-\mathbb{E}_p(R(X)))\right|\right] \\
  &\leq r(\epsilon) \|\theta-\theta_0\|^2 \gamma,
\end{align*}
and therefore
\begin{align*}
I_{\epsilon, 2} &\leq \frac{4r(\epsilon)^2\gamma^2e^{t_a(\epsilon')}(L'_1(\epsilon')N)^{D/2} \int_{B_\epsilon}  \|\theta-\theta_0\|^4 e^{\frac{ \|S_N(\theta_0)\|^2 }{ 2L_0 } -\frac{ NL_0 }{2 } \left\|\theta - \theta_0 -\frac{ S_N(\theta_0)}{ \sqrt{N}L_0 }\right\|^2}\pi_0(\theta)d\theta   }{\pi^{D/2}\pi_{0,\inf}}.
\end{align*}
We further bound the numerator using a similar technique as in the analysis of $I_{\epsilon,1}$:
\begin{align*}
    \|\theta-\theta_0\|^2 &= \left(\|\theta-\theta_0\|  - \frac{ \|S_N(\theta_0)\|}{ \sqrt{N}L_0 } + \frac{ \|S_N(\theta_0)\|}{ \sqrt{N}L_0 } \right)^2 \\
    &\leq 2\left(\|\theta-\theta_0\|  - \frac{ \|S_N(\theta_0)\|}{ \sqrt{N}L_0 }\right)^2  + 2\left(\frac{ \|S_N(\theta_0)\|}{ \sqrt{N}L_0 } \right)^2 \\
    &\leq 2\left\|(\theta-\theta_0) - \frac{ S_N(\theta_0)}{ \sqrt{N}L_0 }\right\|^2 + 2\left(\frac{ \|S_N(\theta_0)\|}{ \sqrt{N}L_0 } \right)^2.
\end{align*}
Note that by Markov's inequality,
\[
\forall \zeta > 0, \quad \|S_N(\theta_0)\|^2 \leq D/\zeta,
\]
with probability 
$\geq 1- \zeta \mathbb{E}_p( \|\nabla_\theta f_1(\theta_0)\|^2)/D$.
If we further restrict to this event, we have that 
\begin{align} \label{I2}
I_{\epsilon, 2}&\leq \frac{ 4e^{t_a(\epsilon')} r(\epsilon)^2\gamma^2 \pi_{0,\sup} (L_1'(\epsilon') N)^{D/2}}{\pi^{D/2}\pi_{0,\inf}} \int_{B_\epsilon} \|\theta-\theta_0\|^4 e^{\frac{ \|S_N(\theta_0)\|^2 }{ 2L_0 } -\frac{ NL_0 }{2 } \left\|\theta - \theta_0 -\frac{ S_N(\theta_0)}{ \sqrt{N}L_0 }\right\|^2}d\theta \nonumber \\
&\leq \frac{ 16e^{t_a(\epsilon')} r(\epsilon)^2\gamma^2 \pi_{0,\sup} (L_1'(\epsilon') N)^{D/2}e^{\frac{D}{2L_0\zeta}}}{\pi^{D/2}\pi_{0,\inf}} \nonumber \\
&~~~~~~~~~~~~~~~~\times\int_{B_\epsilon} \left(\left\|(\theta-\theta_0) - \frac{ S_N(\theta_0)}{ \sqrt{N}L_0 }\right\|^2 + \left(\frac{ \|S_N(\theta_0)\|}{ \sqrt{N}L_0 } \right)^2\right)^2 e^{-\frac{ NL_0 }{2 } \left\|\theta - \theta_0 -\frac{ S_N(\theta_0)}{ \sqrt{N}L_0 }\right\|^2}d\theta \nonumber \\
&\leq \frac{ 16e^{t_a(\epsilon')} r(\epsilon)^2\gamma^2 \pi_{0,\sup} (L_1'(\epsilon') N)^{D/2}e^{\frac{D}{2L_0\zeta}}(NL_0)^{-(D/2+2)}}{\pi^{D/2}\pi_{0,\inf}} \nonumber \\
&~~~~~~~~~~~~~~~~\times\int\left(\|u\|^4 + \frac{2\|u\|^2\|S_N(\theta_0)\|^2}{L_0} + \frac{\|S_N(\theta_0)\|^4}{L_0^2}\right) e^{-\frac{1}{2}\|u\|^2}du \nonumber \\
&\leq \frac{ 16e^{t_a(\epsilon')} r(\epsilon)^2\gamma^2 \pi_{0,\sup} (L_1'(\epsilon') N)^{D/2}e^{\frac{D}{2L_0\zeta}}(NL_0)^{-(D/2+2)}(2\pi)^{D/2}}{\pi^{D/2}\pi_{0,\inf}}\nonumber \\
&~~~~~~~~~~~~~~~~\times\int\left(\|u\|^4 + \frac{2\|u\|^2D}{L_0\zeta} + \frac{D^2}{L_0^2\zeta^2}\right) (2\pi)^{-D/2}e^{-\frac{1}{2}\|u\|^2}du \nonumber \\
& \leq \frac{ 16e^{t_a(\epsilon')} D^2r(\epsilon)^2 \gamma^2 \pi_{0,\sup} (2L_1'(\epsilon')/L_0)^{D/2} e^{\frac{D}{2L_0\zeta}}}{N^2L_0^2\pi_{0,\inf}} \left[3 + \frac{2}{L_0\zeta} + \frac{1}{L_0^2\zeta^2}\right] \nonumber \\
& \leq \frac{ 16e^{t_a(\epsilon')} D^2r(\epsilon)^2 \gamma^2 \pi_{0,\sup} (2L_1'(\epsilon')/L_0)^{D/2} e^{\frac{D}{2L_0\zeta}}}{\zeta^2N^2L_0^2\pi_{0,\inf}} \left[3 + \frac{2}{L_0} + \frac{1}{L_0^2}\right], 
\end{align}
using the substitution
$u=\sqrt{NL_0}\left((\theta-\theta_0) - S_N(\theta_0)/(\sqrt{N}L_0)\right)$.
Noting that $t_a(\epsilon') = \Theta(\sqrt{1/a})$, we will set $\sqrt{a} = \zeta$.
What remains is to lower bound the probability that the three conditions \cref{eq:conditions1,eq:conditions2,eq:conditions3} hold.
Denoting by $\psi(x) \in \mathbb R^{d_1}$ the 
coordinates of $f^{(1)}(x, \cdot)-f^{(1)}(x,\theta_0)$ in an
orthonormal basis of $S_1$, we define 
$Z(x)\in \mathbb R^d$, $d \leq d_1 + D$, as
\begin{align}\label{eq:z_def}
    Z(x) :=\left[\begin{array}{c}
\psi(x) \\
\text{indep}\left(\nabla_\theta f(x; \theta_0)\right)
\end{array}\right], 
\end{align}
where $\text{indep}\left(\cdot\right)$ selects the 
linearly independent components of the 
$\nabla_\theta f(x; \theta_0)$ (considered as functions of $x$).
Without loss of generality we can choose $R(x)$ to be linearly independent of 
$Z(x)$ as a function of $x$. 
In this case,  the three conditions hold simultaneously if 
\begin{align*}
\sum_m v_m\psi_j(x_m) = \frac{ \sum_n \psi_j(x_n)}{N} \quad \forall j, \quad 
\sum_m v_m (R(x_m) - \mathbb{E}_pR(X)) = 0, \quad \sum_m v_m \nabla f_m(\theta_0) = \frac{ \sum_n \nabla f_n(\theta_0) }{N}.
\end{align*}
As in the proof of \cref{thm:optimal_kl:finitd}, we can derive 
that for $\delta \in (0,1]$, if $M\gtrsim J(\delta)^{-1} D ( \log N + 1 ) $,
where
\begin{align}
J(\delta) 
 = \!\!\!\inf_{\substack{a \in \mathbb R^{d+1};\\ \|a\|=1}} \!\!\! \Pr \left( \left\langle a, [Z(X),R(X)]^T\right\rangle\!>\! \frac{2  \mathbb{E}_p(\|Z(X)\|^2)}{\sqrt{N\delta}} \right),
\end{align} 
then with probability greater than
$1-  \delta - e^{-J(\delta) M/4}$, 
\cref{eq:conditions1,eq:conditions2,eq:conditions3} hold.
We will set $\delta = \zeta^2$.
Our particular choice of $Z$ using independent components
ensures that $J(\delta)$ is of full rank, and therefore
is bounded from below by $0$.

We can now obtain a final high probability bound on the KL divergence by combining the bounds on $I_{\epsilon,1}$ 
and $I_{\epsilon,2}$. In particular, we have that
if $M \gtrsim J(\zeta^2)^{-1}D(\log N+1)$,
we have that for any 
$\zeta>0$, $\sqrt{\frac{D}{2NL'_1(\epsilon')}} \leq \epsilon' < \epsilon_0$, and $0 \leq \epsilon < \epsilon_0$,
\begin{align*}
KL(\pi_{w(v)}||\pi) &\leq 4N^2\int_0^1 (1-t) \left(I_{\epsilon,1} + I_{\epsilon,2} \right) \mathrm{d}t \\
\leq  
&
4N^{D/2+2}e^{\frac{1}{\zeta}\sqrt{\frac{DL_2}{L'_1(\epsilon')}} -NL_0\epsilon^2/8}\left(\frac{2}{\pi_{0,\inf}} \int_\Theta \mathbb{E}_p(\Delta_1 - \bar \Delta)^2(\theta)\pi_0(\theta)d\theta\left(\frac{L'_1(\epsilon')}{\pi}\right)^{D/2}\right)\nonumber\\
&  +
 \frac{4r(\epsilon)^2e^{\frac{1}{\zeta}\left(\sqrt{\frac{DL_2}{L'_1(\epsilon')}}+\frac{D}{2L_0}\right)}}{\zeta^2}
\left(\frac{ 8 D^2 \gamma^2 \pi_{0,\sup} (2L_1'(\epsilon')/L_0)^{D/2} }{L_0^2\pi_{0,\inf}} \left[3 + \frac{2}{L_0} + \frac{1}{L_0^2}\right]\right)
\end{align*}
with probability at least
\begin{align*}
1 - e^{-J(\zeta^2)M/4} - 4Me^{-NL_0\epsilon^2/4} 
- 3\frac{\mathbb{E}_p\left(\| \nabla_\theta f_1 (\theta_0)\|^2\right)}{\epsilon^2NL_0^2}
- \frac{\mathrm{Var}_p(R(X))}{N} - 3\zeta^2 - \zeta \frac{\mathbb{E}_p( \|\nabla_\theta f_1(\theta_0)\|^2)}{D}.
\end{align*}
If $\zeta^{-1} = o(-\log r(\epsilon))$ as $\epsilon \to 0$, then
the second term in the KL bound converges to 0.
Consequently if $\log N + |\log r(\epsilon)| = o(N\epsilon^2)$ as $N \to \infty$,
the KL divergence converges to 0 as $N\to\infty$.
Similarly, $J(\zeta^2)$ converges to a nonzero constant as long as $\zeta\sqrt{N} \to \infty$.
Both conditions are possible to satisfy, for any $r(\epsilon)$, by making $\epsilon \to 0$ slowly enough as a function of $N$.

\end{proof}
\subsection{Proof of \cref{thm:convergence}}
\begin{proof}
For step size $\gamma \in [0,1]$,  the result $\hat w_{k+1}$ of 
the approximate Newton step prior to projection onto $\mathcal{W}$ is
\begin{align*}
\hat w_{k+1} &= w_k + \gamma(G(w_k)+\tau I)^{-1}H(w_k)(1-w_k)\\
&= w_k + \gamma(G(w_k)+\tau I)^{-1}H(w_k)(w^\star_k-w_k) + \gamma(G(w_k)+\tau I)^{-1}H(w_k)(1-w^\star_k) \\
&= w_k + \gamma(G(w_k)+\tau I)^{-1}G(w_k)(w^\star_k-w_k) + \gamma(G(w_k)+\tau I)^{-1}H(w_k)(1-w^\star_k),
\end{align*}
where the last line follows from the fact that $w_k, w^\star_k \in \mathcal{W}$.
Consider the distance $\left\|w_{k+1} - w^\star_{k+1}\right\|$.
Since $w^\star_{k+1}$ is the projection of $w_{k+1}$ onto a convex set $\mathcal{W}^\star$,
we have that
\begin{align*}
\left\|w_{k+1} - w^\star_{k+1}\right\|  \leq \left\|w_{k+1} - w^\star_{k}\right\|.
\end{align*} 
Furthermore, since $w_{k+1}$ is the projection of $\hat w_{k+1}$ onto $\mathcal{W}$,
and $w^\star_k \in \mathcal{W}$,
\begin{align*}
\left\|w_{k+1} - w^\star_{k}\right\| \leq \left\|\hat w_{k+1} - w^\star_k\right\|.
\end{align*}
Then, we have that:
\begin{align*}
\left\|\hat w_{k+1} - w^\star_k\right\|&= 
\left\|\left(I - \gamma(G(w_k)+\tau I)^{-1}G(w_k)\right)(w_k-w^\star_k) + \gamma(G(w_k)+\tau I)^{-1}H(w_k)(1-w^\star_k)\right\|\\
&\leq \left\|\left(I - \gamma(G(w_k)+\tau I)^{-1}G(w_k)\right)(w_k-w^\star_k)\right\| + \gamma(\epsilon\|w_k-w^\star_k\|+\delta)\\
&\leq \eta\left\|w_k-w^\star_k\right\| + \gamma\delta.
\end{align*}
The first bound follows from our assumption in the theorem statement, and the triangle inequality. The second bound follows via the following logic. Since $\pi_w \propto \exp(w^Tf_{[M]}(\theta))\pi_0(\theta)$,
it has the same support for all $w\in\mathcal{W}$. Since
$G(w)$ is the covariance matrix of $f_{[M]}(\theta)$, the null space of $G(w)$ is spanned by the set
of vectors $u\in\mathbb{R}^M$ such that $u^Tf_{[M]}(\theta) = c$ holds $\pi_w$-almost everywhere, for some constant $c\in\mathbb{R}$. Therefore the null space of $G(w)$ is the same for all $w\in\mathcal{W}$.
Since $w^Tf_{[M]} = (w+u)^Tf_{[M]}$ (up to a constant in $\theta$) for any $w\in\mathcal{W}$ and $u$ in the null space of $G(w)$, we can 
augment any closed convex subset $\mathcal{W}^\star \subseteq \arg\min_{w\in\mathcal{W}}\mathrm{KL}(\pi_w||\pi)$
with the null space of $G(w)$ to form 
$\mathcal{W}^\star_2 = \left\{ w+u : w\in\mathcal{W}^\star, \, u\in \mathrm{null}\,G(w), w+u\in\mathcal{W}\right\}$,
which is still a closed convex subset of $\mathcal{W}^\star \subseteq \arg\min_{w\in\mathcal{W}}\mathrm{KL}(\pi_w||\pi)$. 
Therefore any maximal convex subset $\mathcal{W}^\star$ must be of the previous form augmented with the null space of $G(w)$.
Because $w_k^\star$ is a projection, $w_k^\star - w_k$ must be orthogonal to this null space. 
Denoting the full eigendecomposition $G(w) = V\Lambda(w)V^T$, $V\in\mathbb{R}^{M\times M}$, $\Lambda(w) \in \mathbb{R}^{M\times M}$,
including zero eigenvalues, and similarly the reduced eigendecomposition $G(w) = V_+\Lambda_+(w)V_+^T$ with the null space removed,
\begin{align*}
\left\|\left(I - \gamma(G(w_k)+\tau I)^{-1}G(w_k)\right)(w_k-w^\star_k)\right\|
&= \left\|V\left(I - \gamma (\Lambda (w_k)+\tau )^{-1}\Lambda(w_k)\right)V^T(w_k-w^\star_k)\right\|\\
&= \left\|V_+\left(I - \gamma (\Lambda_+ (w_k)+\tau )^{-1}\Lambda_+(w_k)\right)V_+^T(w_k-w^\star_k)\right\|\\
&\leq \left(1-\gamma \xi\right)\left\|V_+^T(w_k-w^\star_k)\right\|\\
&= \left(1-\gamma \xi\right)\left\|V^T(w_k-w^\star_k)\right\|\\
&= \left(1-\gamma \xi\right)\left\|(w_k-w^\star_k)\right\|.
\end{align*}
Finally solving the earlier recursion,
\begin{align*}
\left\|w_{k} - w^\star_k\right\|
&\leq \eta^{k}\left\|w_0 - w^\star_0\right\|+ \gamma\delta \sum_{j=0}^{k-1}\eta^j\\
&=\eta^{k}\left\|w_0 - w^\star_0\right\| + \gamma\delta \left(\frac{1-\eta^k}{1-\eta}\right).
\end{align*}
\end{proof}

\section{Experiments}\label{app:experiments}
In this section, we present some additional results, 
along with the full details of the 
Bayesian radial basis function regression experiment.
We also discuss the use of a Laplace approximation 
for the low-cost approximation $\hat{\pi}$ needed for the 
sparse regression methods: greedy iterative geodesic
ascent (GIGA) and iterative hard thresholding (IHT).
Finally, we include a note on the overall performance
of those methods.

\subsection{Synthetic Gaussian location model}\label{app:sub:synth_gauss}
\begin{figure}[t!]
\begin{center}
\begin{subfigure}{0.24\columnwidth}
\includegraphics[width=\columnwidth]{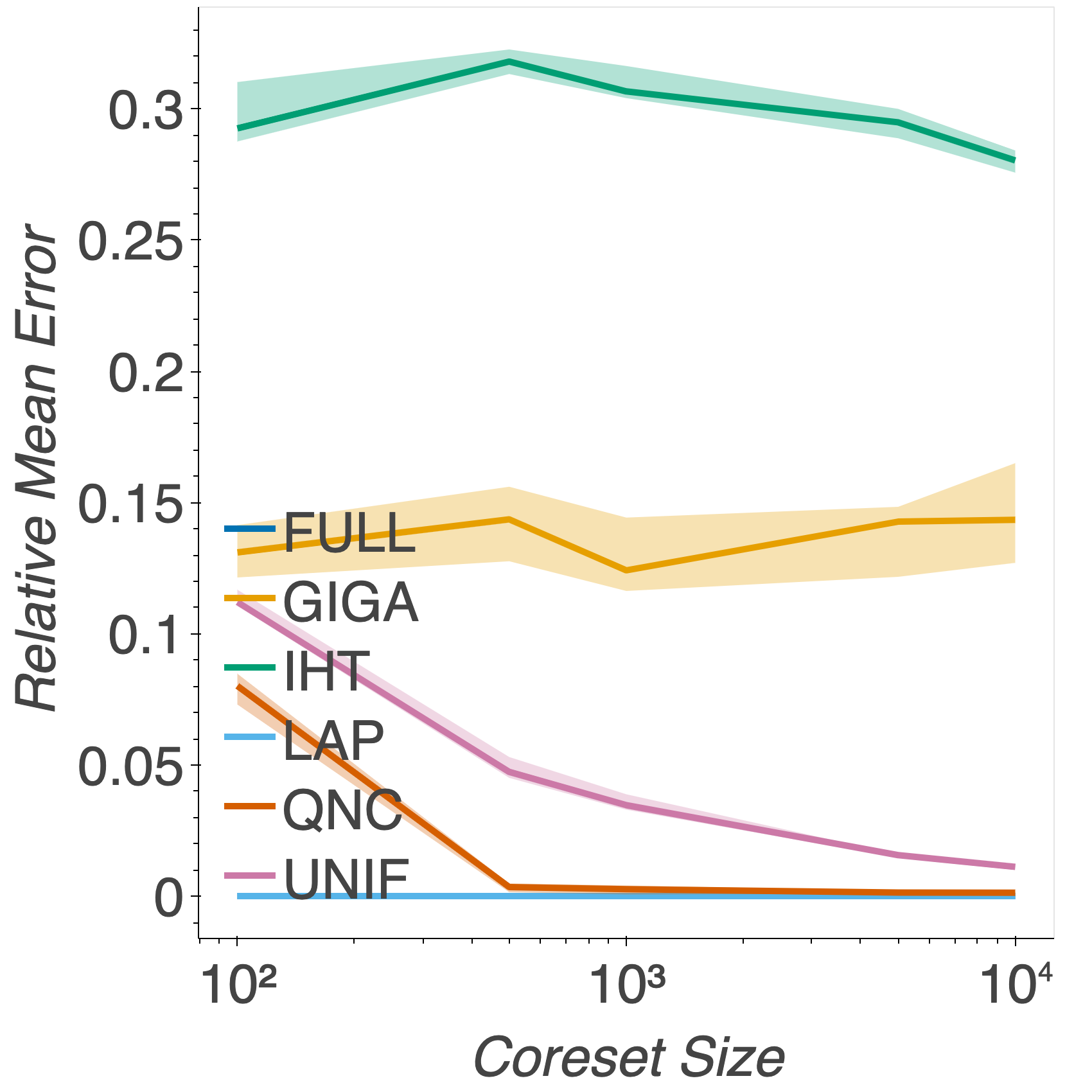}
\end{subfigure}
\begin{subfigure}{0.24\columnwidth}
\includegraphics[width=\columnwidth]{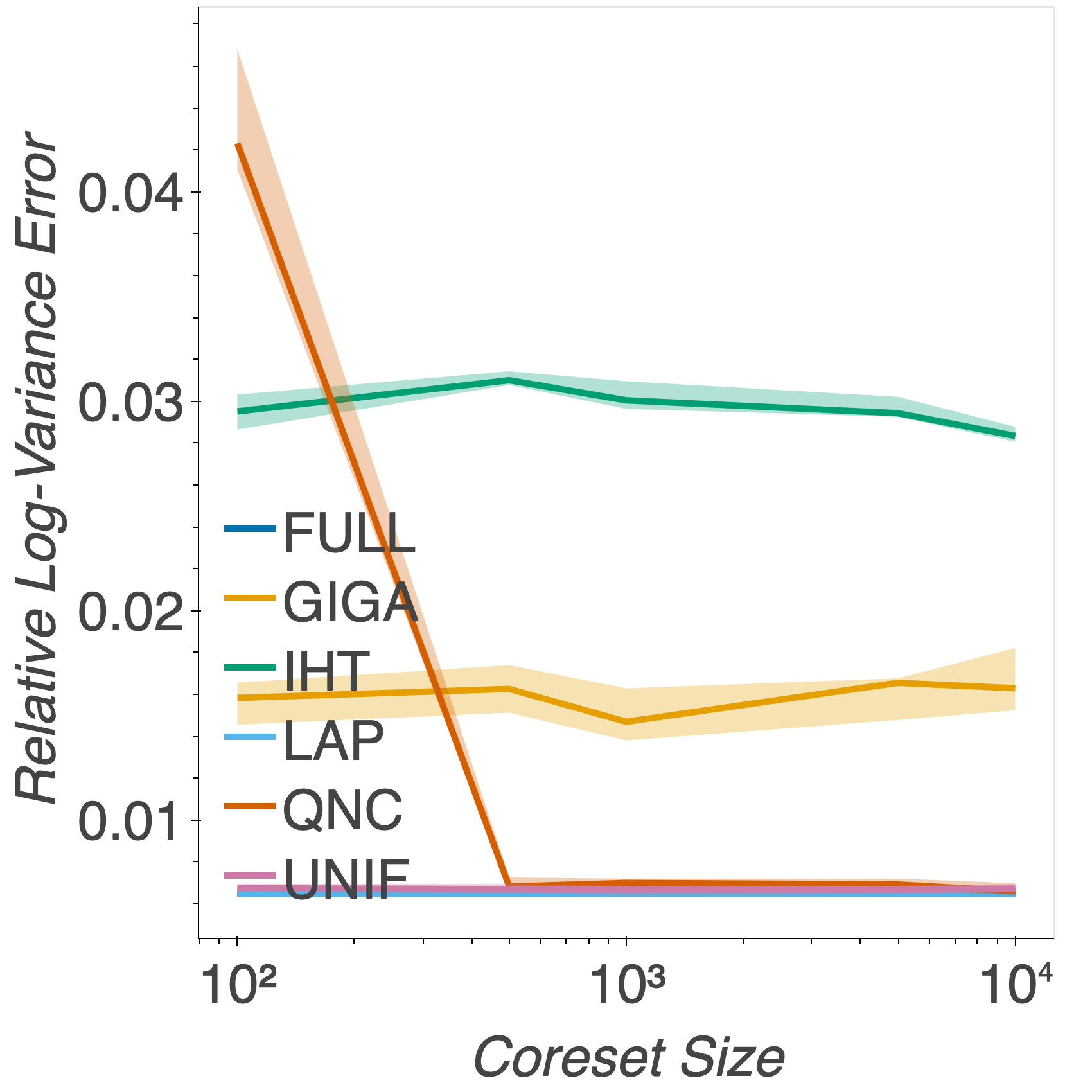}
\end{subfigure}
\begin{subfigure}{0.24\columnwidth}
\includegraphics[width=\columnwidth]{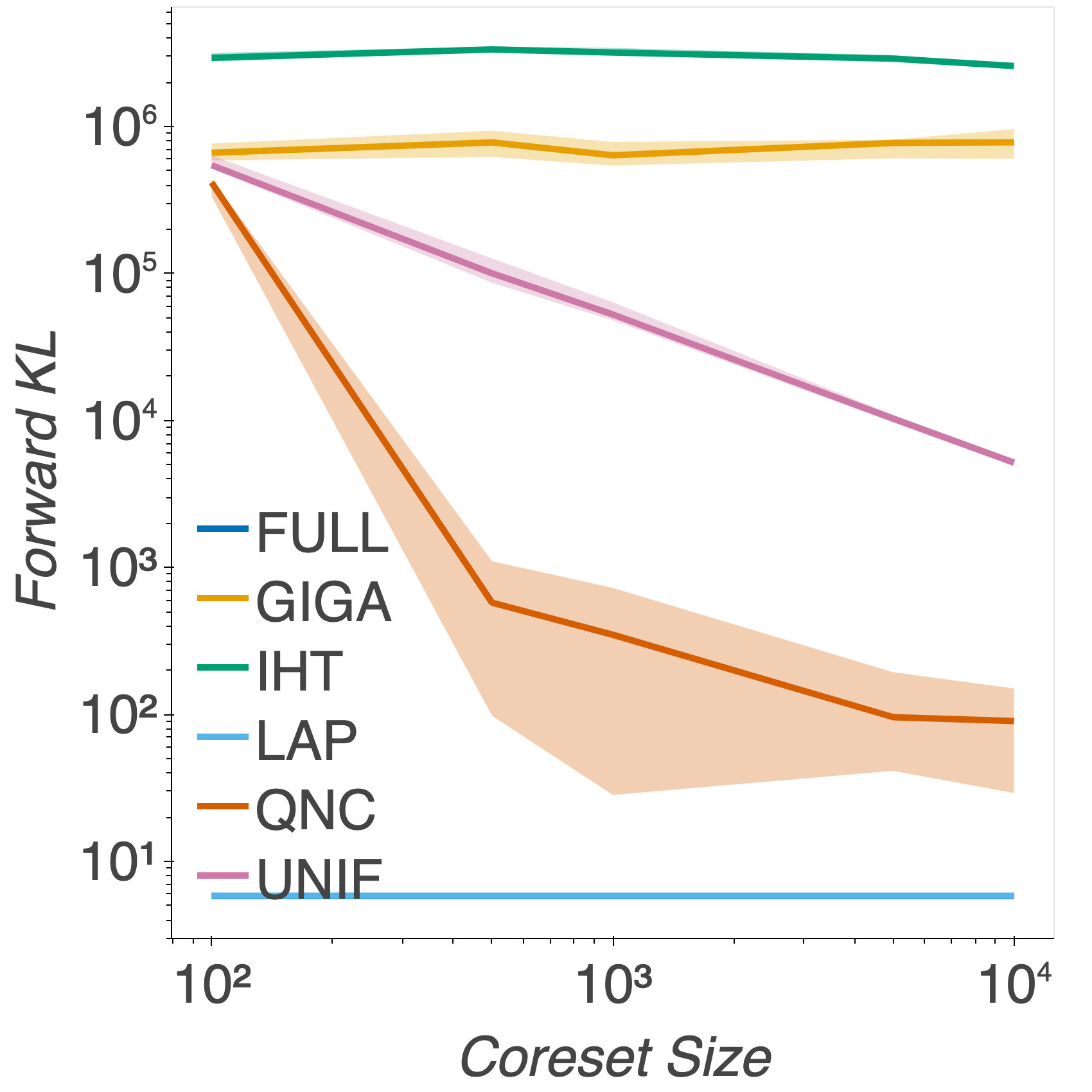}
\end{subfigure}
\begin{subfigure}{0.24\columnwidth}
\includegraphics[width=\columnwidth]{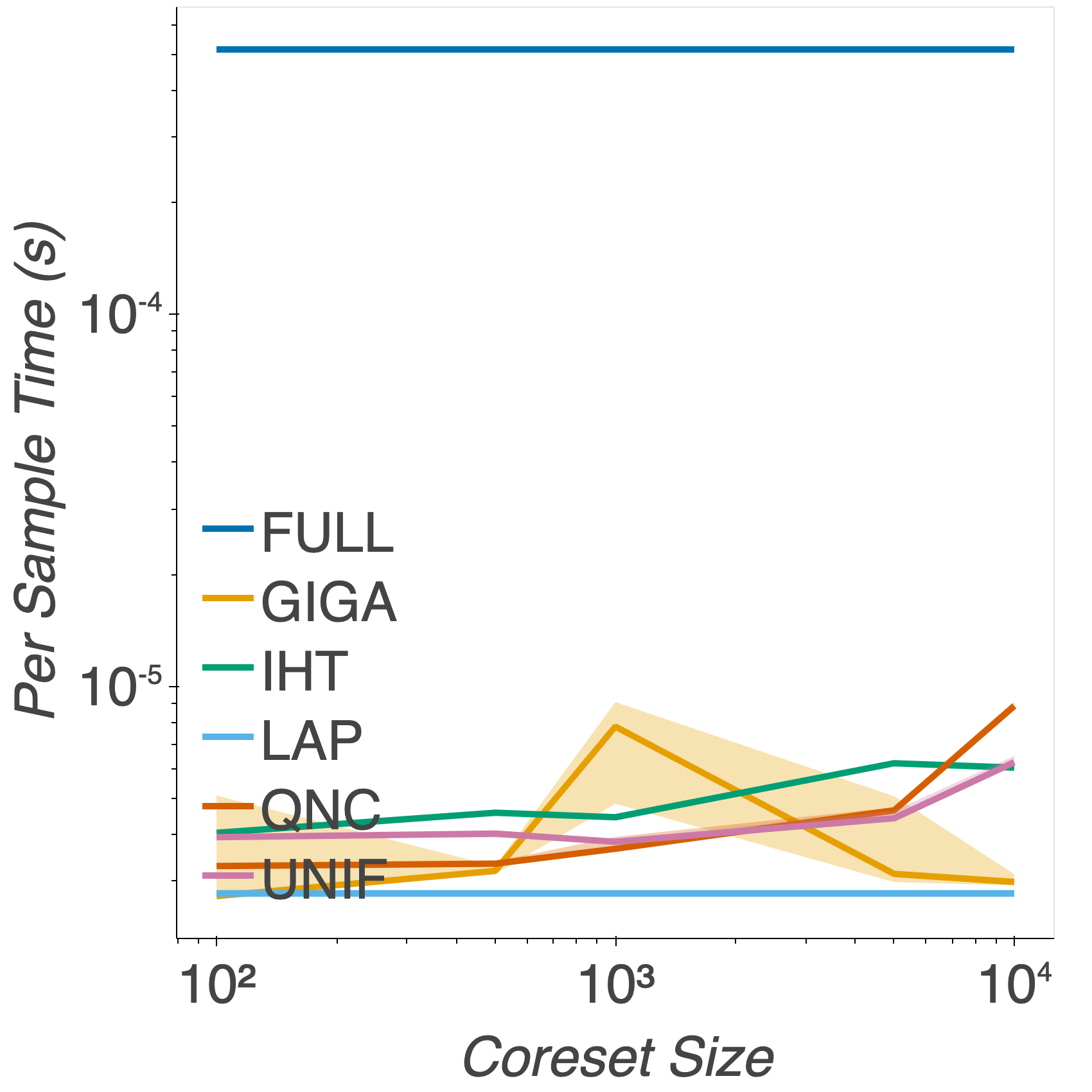}
\end{subfigure}
\caption{Relative mean and log-variance error, forward 
KL divergence and per sample time to sample from the 
respective posteriors for the synthetic Gaussian 
experiment. Our algorithm (QNC) consistently provides 
an improvement in coreset quality over the other 
subsampling methods. All methods provide a significant
reduction in time to sample from the respective posteriors.}
\label{app:fig:synth_gauss_kl_build_time}
\end{center}
\vskip -0.2in
\end{figure}
From \cref{app:fig:synth_gauss_kl_build_time} we 
see that our additional results match closely those 
in \cref{sec:experiments}. Our method (QNC) consistently
outperforms the other subsampling methods, while the 
Laplace approximation gives the best results, by design.

\subsection{Bayesian sparse linear regression}\label{app:sub:sparsereg}
\begin{figure}[t!]
\begin{center}
\begin{subfigure}{0.24\columnwidth}
\includegraphics[width=\columnwidth]{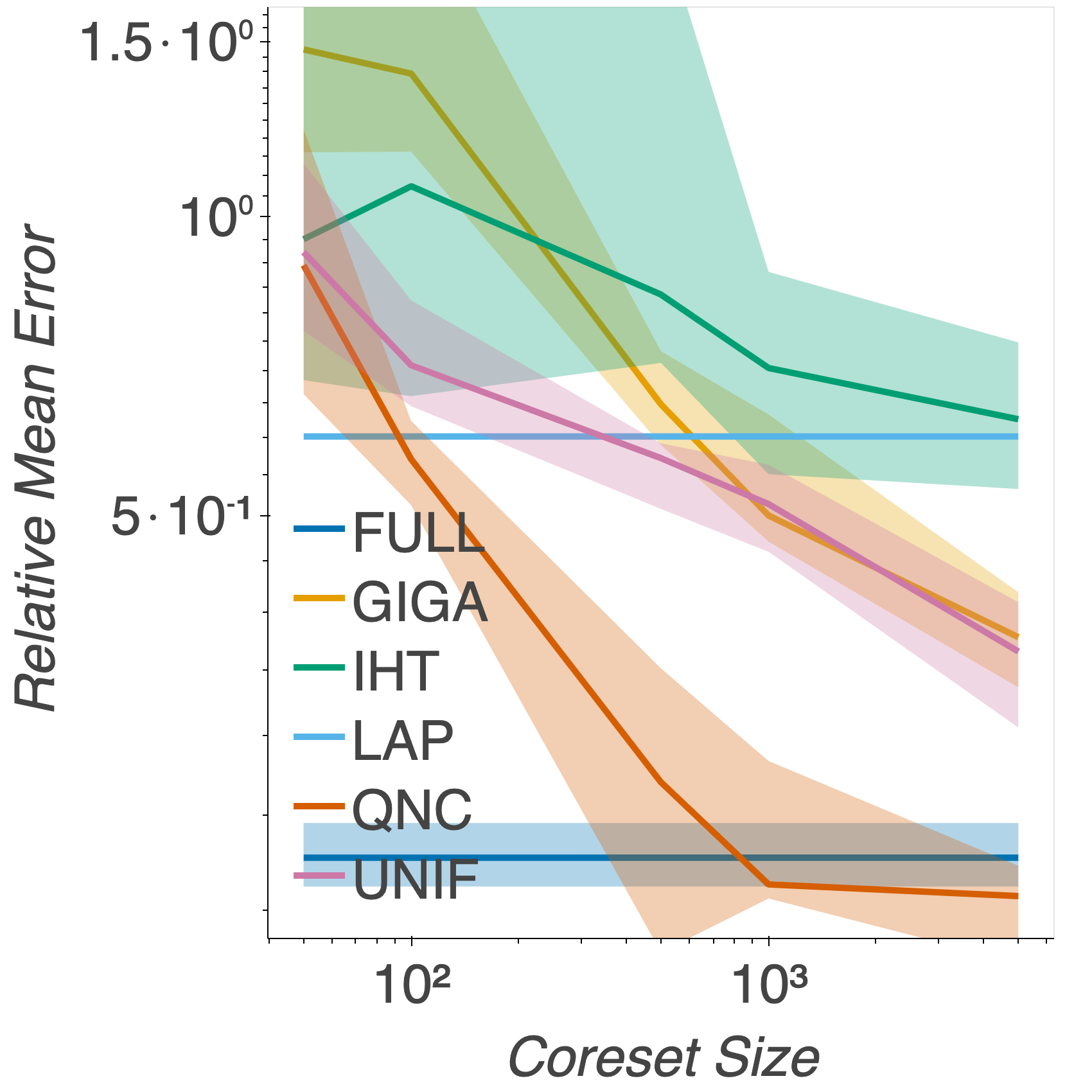}
\end{subfigure}
\begin{subfigure}{0.24\columnwidth}
\includegraphics[width=\columnwidth]{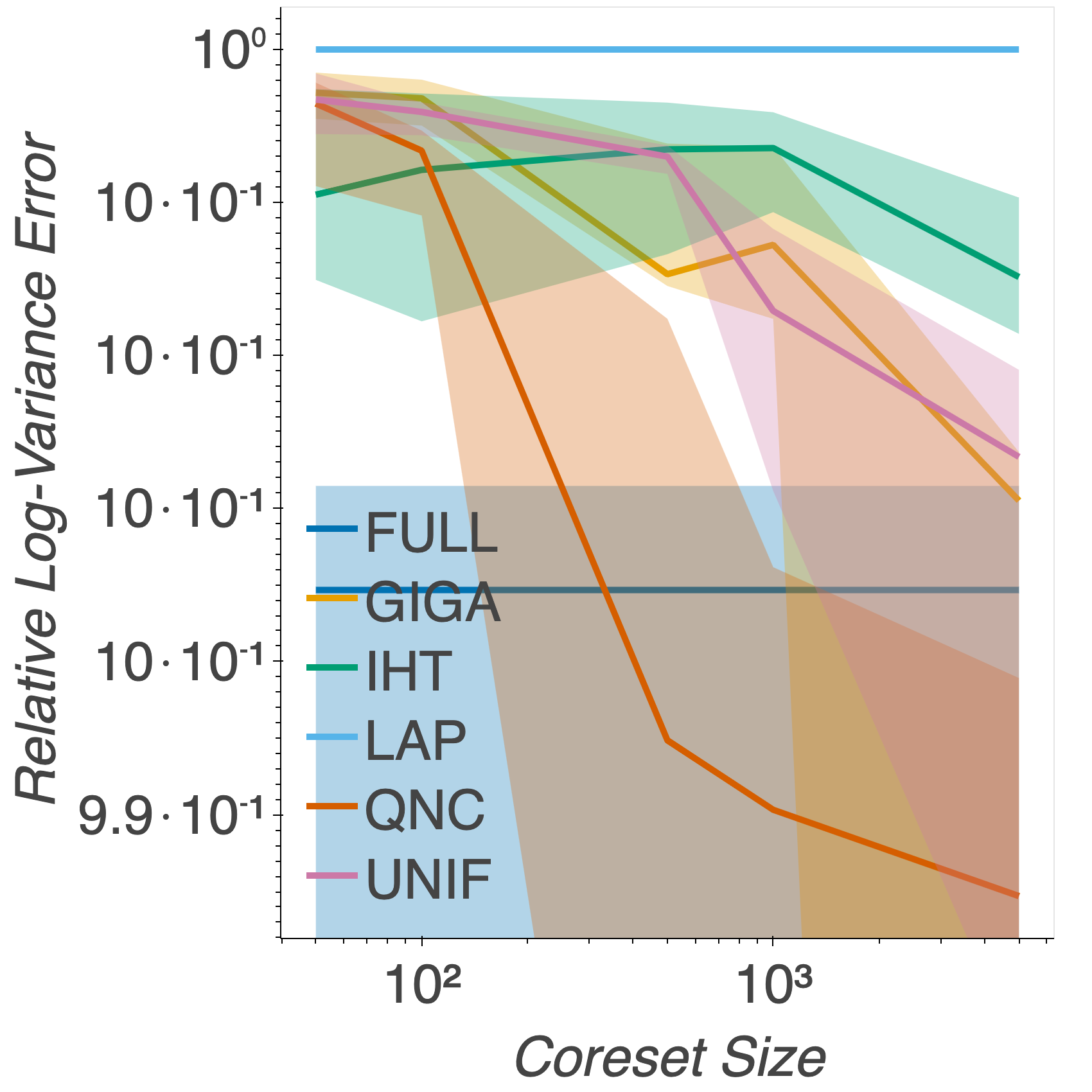}
\end{subfigure}
\begin{subfigure}{0.24\columnwidth}
\includegraphics[width=\columnwidth]{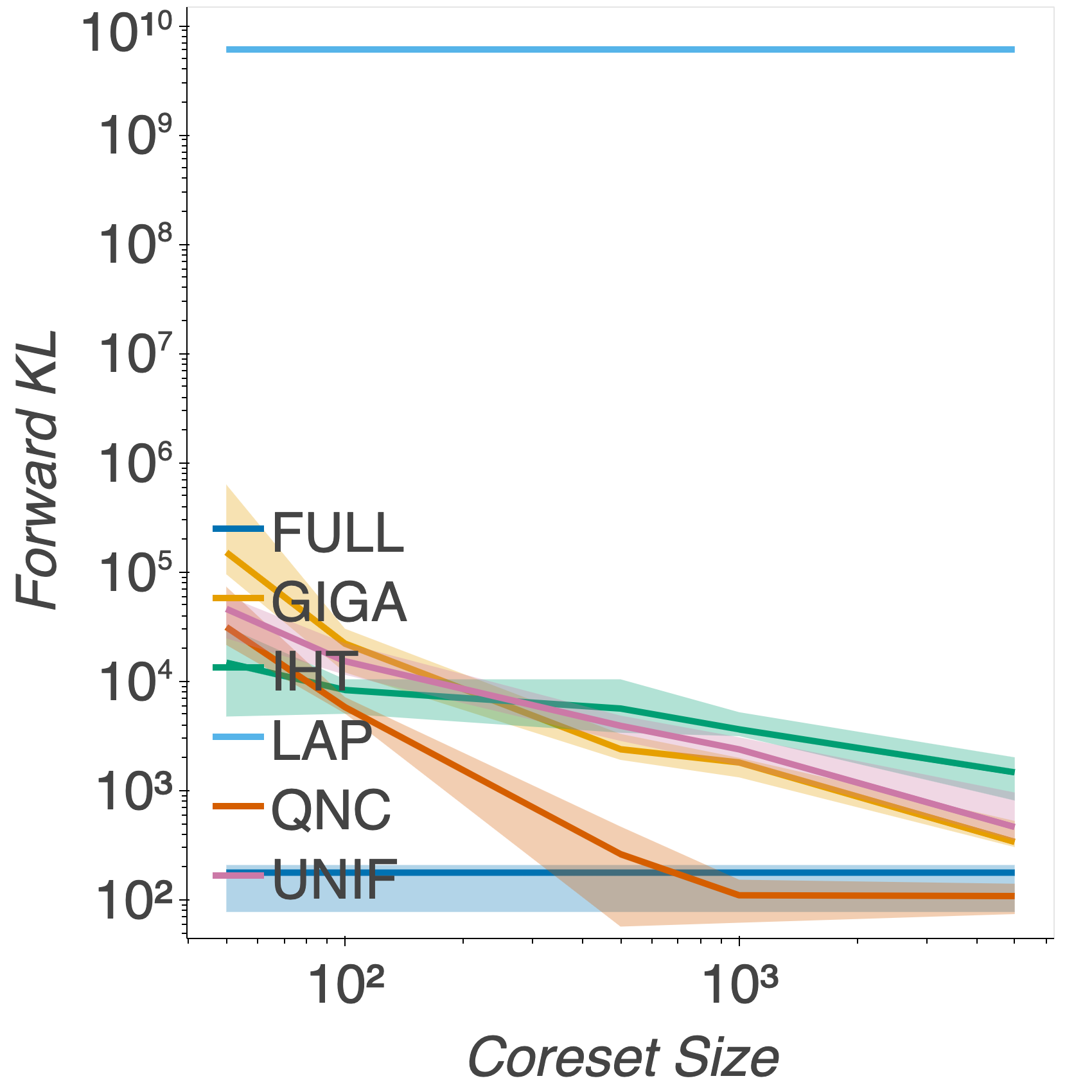}
\end{subfigure}
\begin{subfigure}{0.24\columnwidth}
\includegraphics[width=\columnwidth]{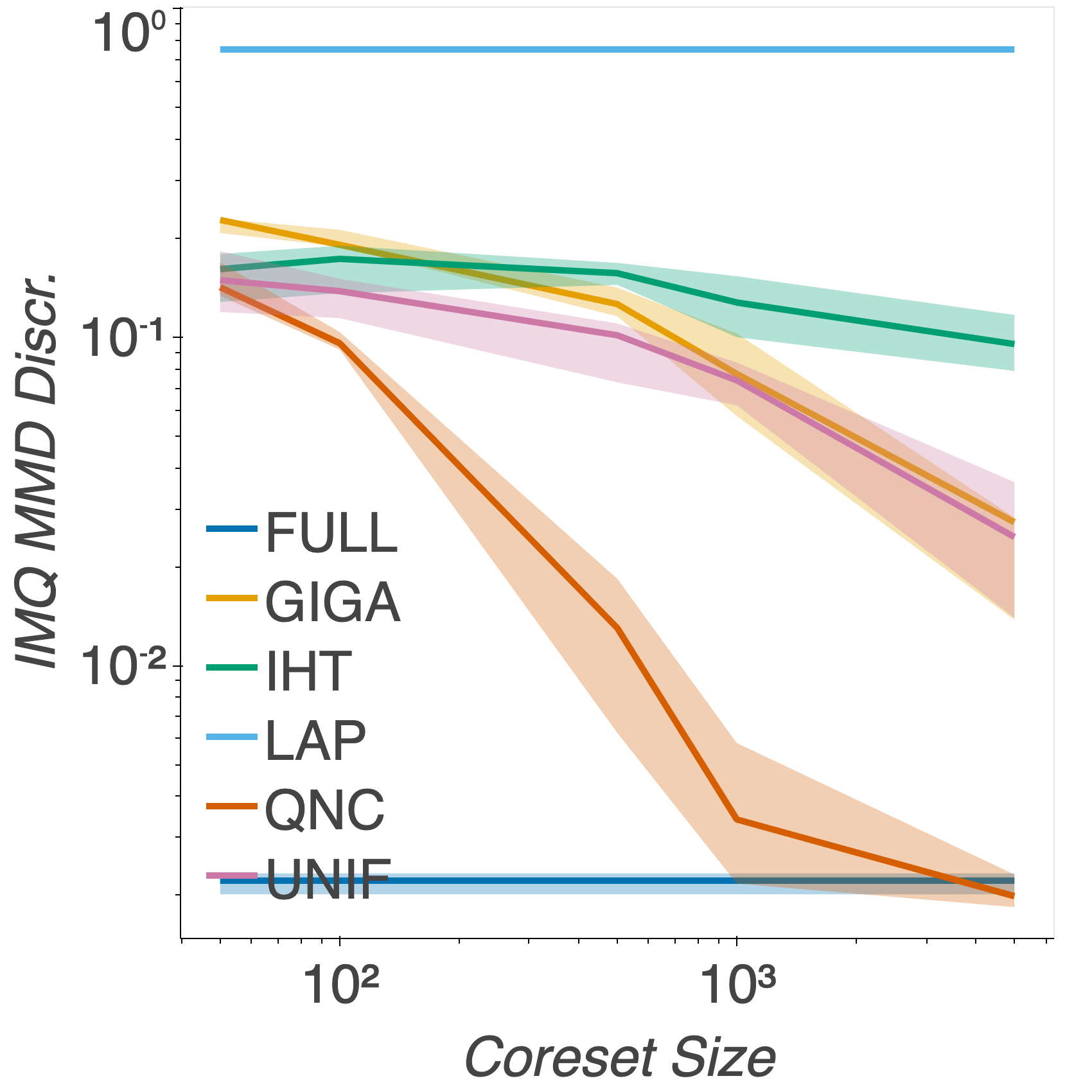}
\end{subfigure}
\begin{subfigure}{0.24\columnwidth}
\includegraphics[width=\columnwidth]{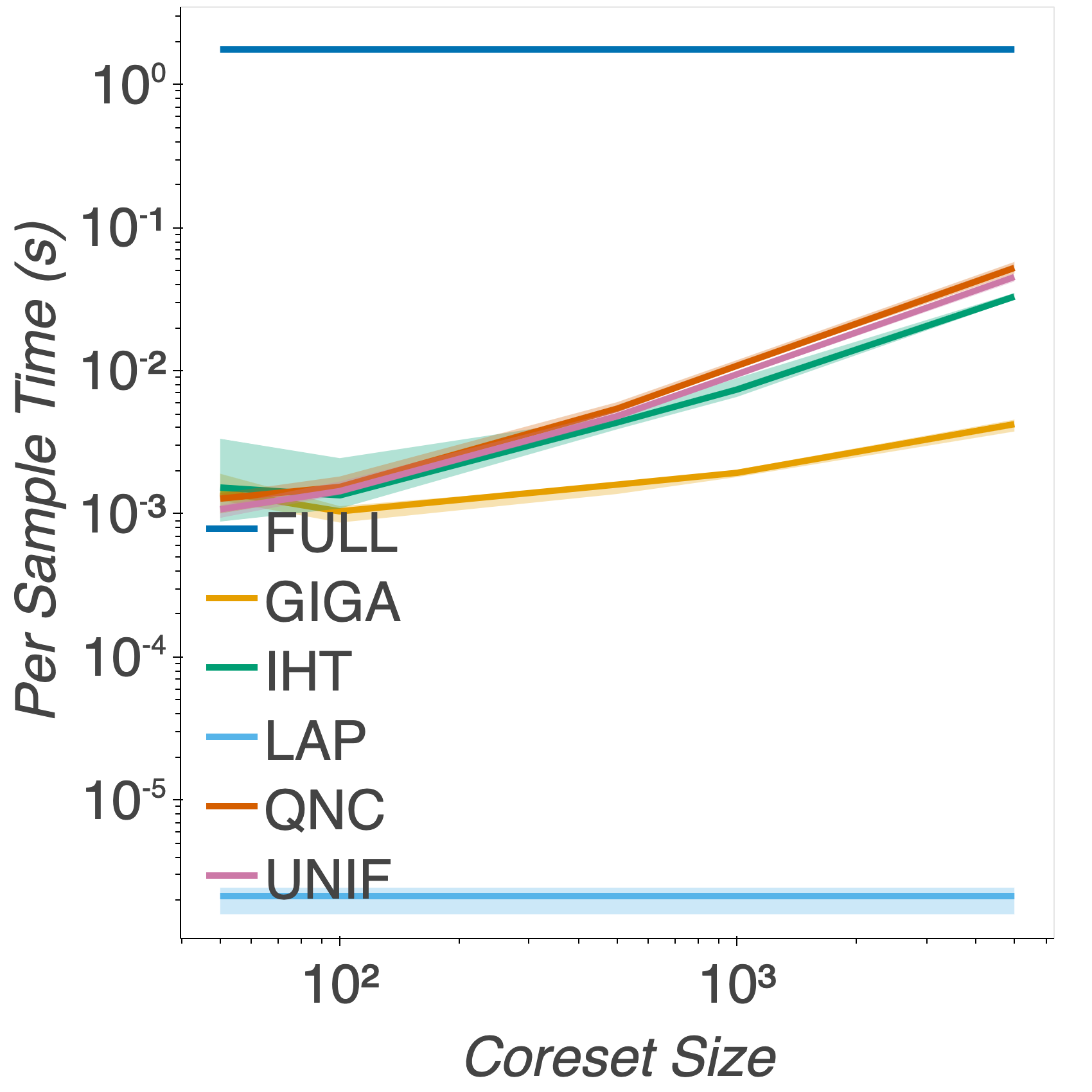}
\end{subfigure}
\caption{Relative mean and log-variance error, forward 
KL divergence, IMQ maximum mean discrepancy (MMD)
and per sample time to sample from the 
respective posteriors for the sparse regression 
experiment. Our algorithm (QNC) consistently provides 
an improvement in coreset quality over the other 
methods. All methods provide a significant
reduction in time to sample from the respective posteriors.
The poor performance of the Laplace approximation 
in this heavy-tailed example can be seen particularly
clearly in the forward KL and IMQ MMD plots.}
\label{app:fig:delays_sparsereg_stein_kl_build_time}
\end{center}
\vskip -0.2in
\end{figure}
From \cref{app:fig:delays_sparsereg_stein_kl_build_time} we 
see that our additional results match closely those 
in \cref{sec:experiments}. Our method (QNC) consistently
outperforms the other methods. In order to assess the computation gains our coreset approach achieves, a useful metric is to calculate the number of posterior samples $N_{sample}$ for which the time taken to obtain $N_{sample}$ samples from the full posterior is the same as the time taken to construct a coreset using QNC, and then take $N_{sample}$ samples from the coreset posterior. Here, we find that $N_{sample} \approx 60$ for a coreset of size $1000$. This is far fewer than you would ideally like to have in practice.

However, we note that these results are conservative, and in fact we expect that the gains from the coreset approach are more significant. This is because we perform sampling in each case using STAN\citep{carpenter2017stan}, which uses C++. However, we construct the coresets in Python, and most of the coreset build time comes from the slowness of a non-compiled language. We expect that our estimated value of $N_{sample}$ would be far smaller if we not only performed sampling in C++, but also constructed the coresets in C++.

\begin{figure}[t!]
\begin{center}
\includegraphics[width=0.5\columnwidth]{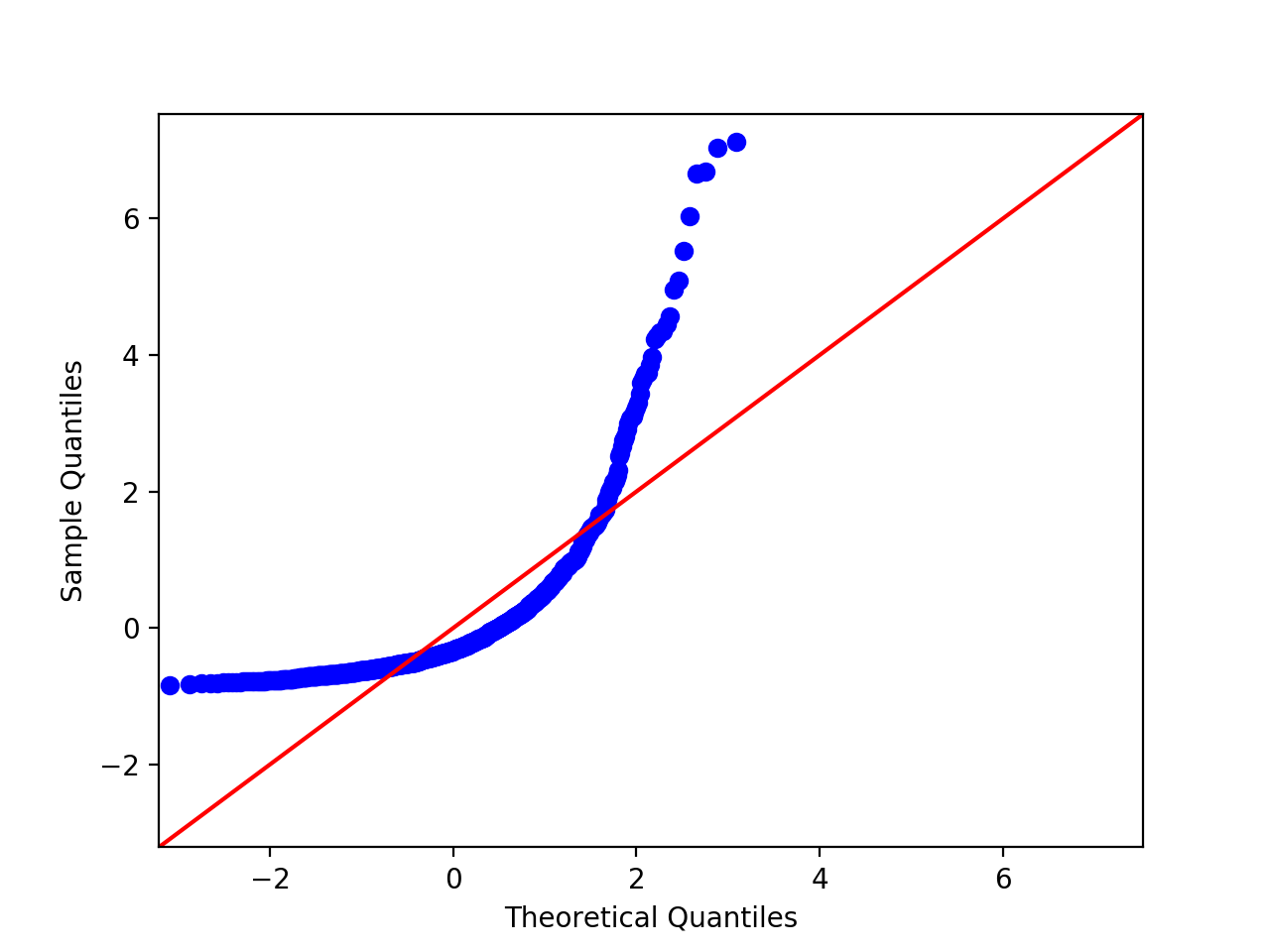}
\caption{Quantile-quantile plot, comparing the 
quantiles of the (centred and scaled) 
marginal posterior distribution 
of $\lambda_{13}$ to the quantiles of a standard normal
distribution.}
\label{app:fig:sparsereg_qq}
\end{center}
\vskip -0.2in
\end{figure}

In this experiment, the prior
is heavy tailed, whilst the likelihood has sub-exponential tails.
However, when the data is in fact concentrated on a particular
subspace of the overall space, then the posterior can have 
heavy tails, even when the number of data points is large.
We can clearly see this in \cref{app:fig:sparsereg_qq},
where we plot the quantiles of the (centred and scaled) 
marginal posterior distribution 
of the parameter $\lambda_{13}$ to the quantiles of a standard normal
distribution. This parameter is strictly positive, but 
we see that the right hand tail of the distribution 
is significantly heavier than a normal distribution.

In the forward KL and IMQ maximum mean discrepancy (MMD) 
plots, we can see clearly that the Laplace approximation
is performing poorly, as it is underestimating the tails
of the posterior distribution. Calculating the IMQ stein
discrepancy in this experiment is computationally 
intractable with the size of dataset we consider.

\subsection{Heavy-tailed Bayesian logistic regression}

\begin{figure}[t!]
\begin{center}
\begin{subfigure}{0.24\columnwidth}
\includegraphics[width=\columnwidth]{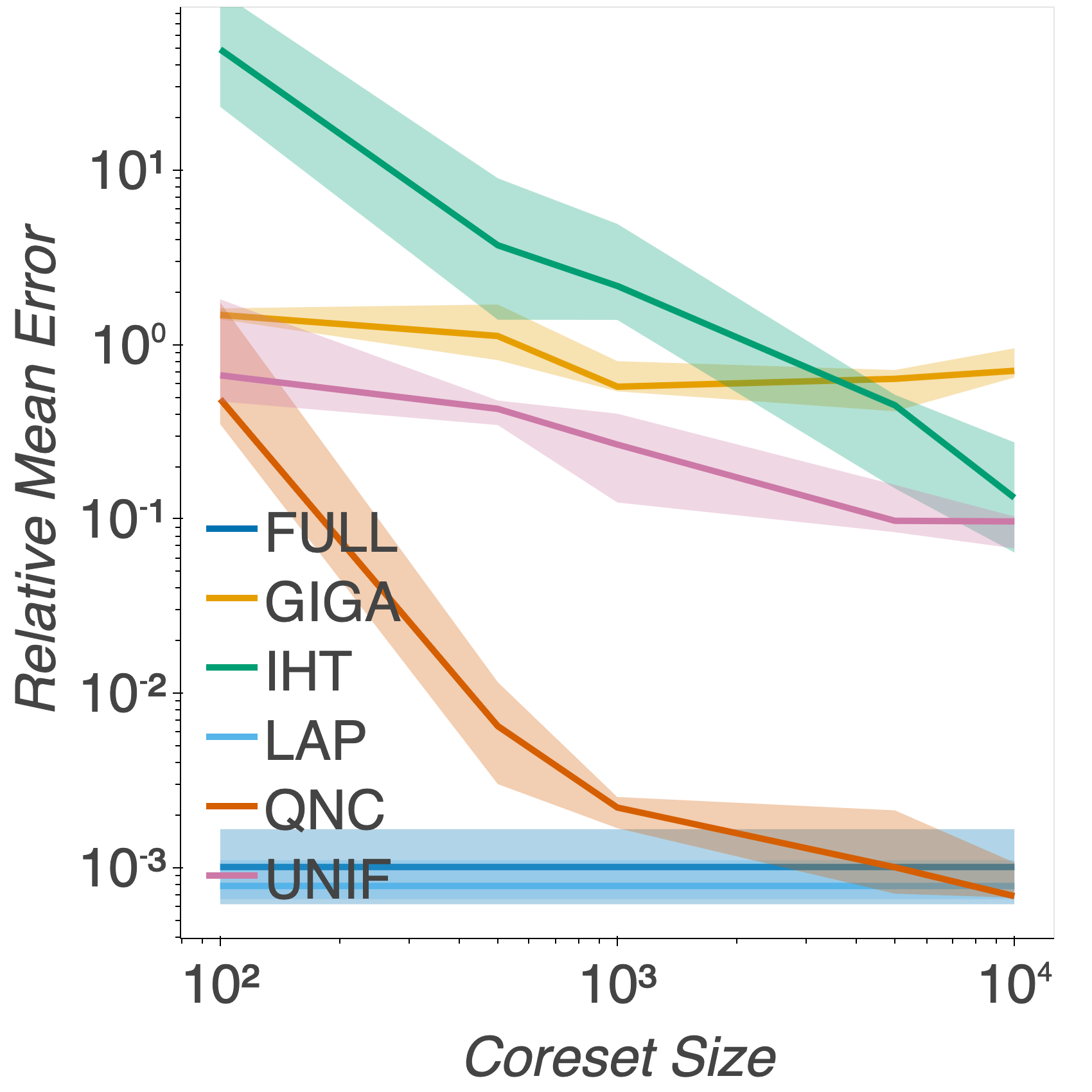}
\end{subfigure}
\begin{subfigure}{0.24\columnwidth}
\includegraphics[width=\columnwidth]{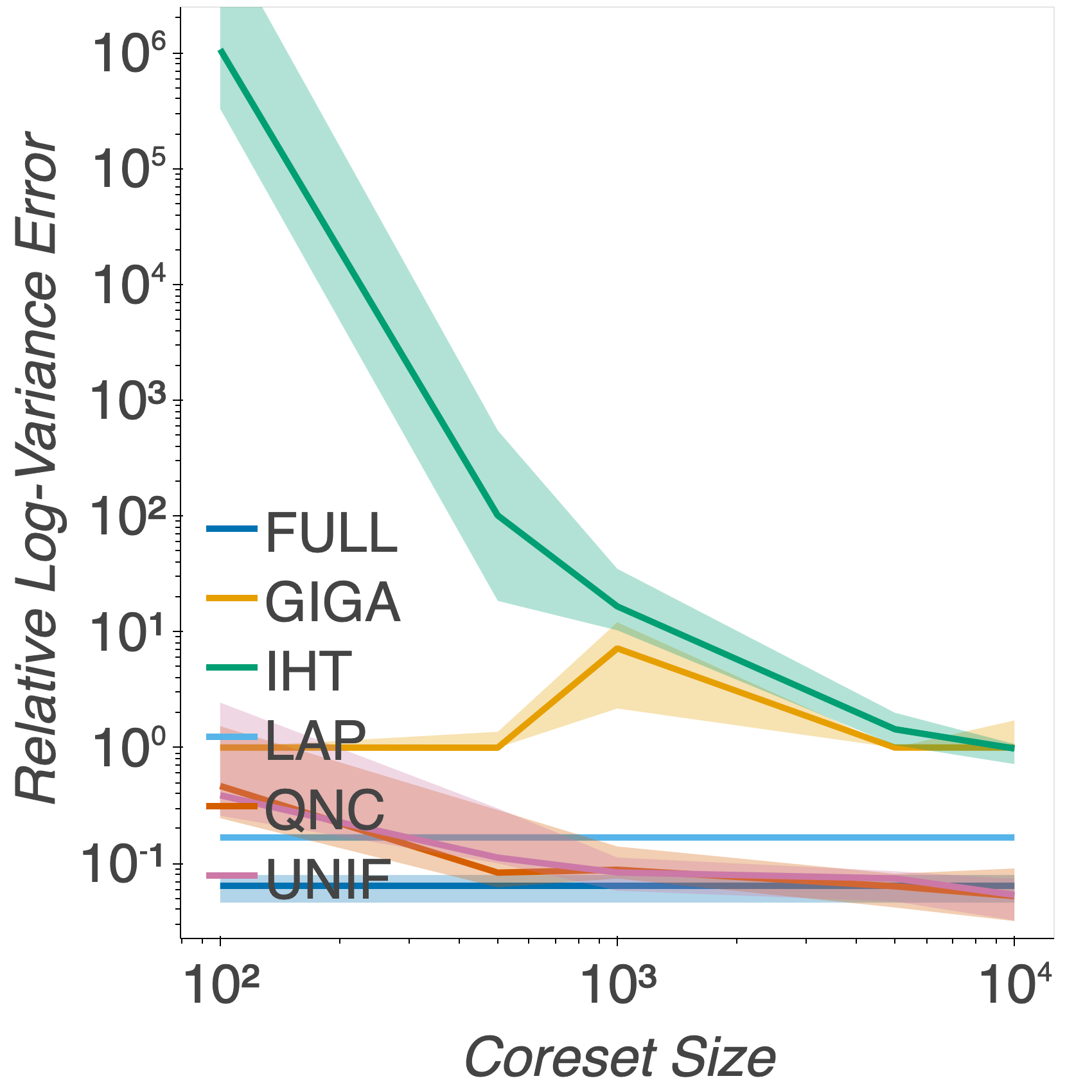}
\end{subfigure}
\begin{subfigure}{0.24\columnwidth}
\includegraphics[width=\columnwidth]{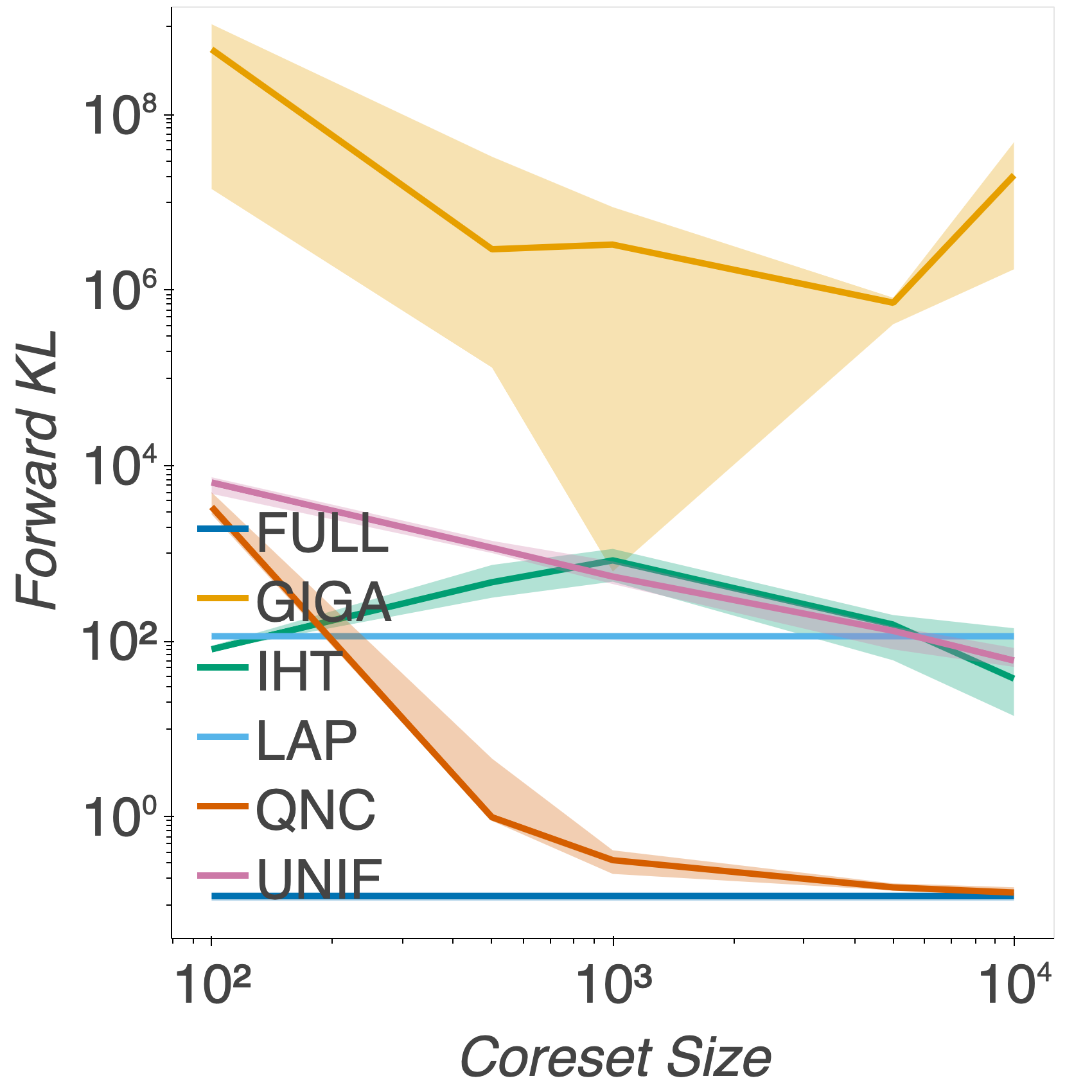}
\end{subfigure}
\begin{subfigure}{0.24\columnwidth}
\includegraphics[width=\columnwidth]{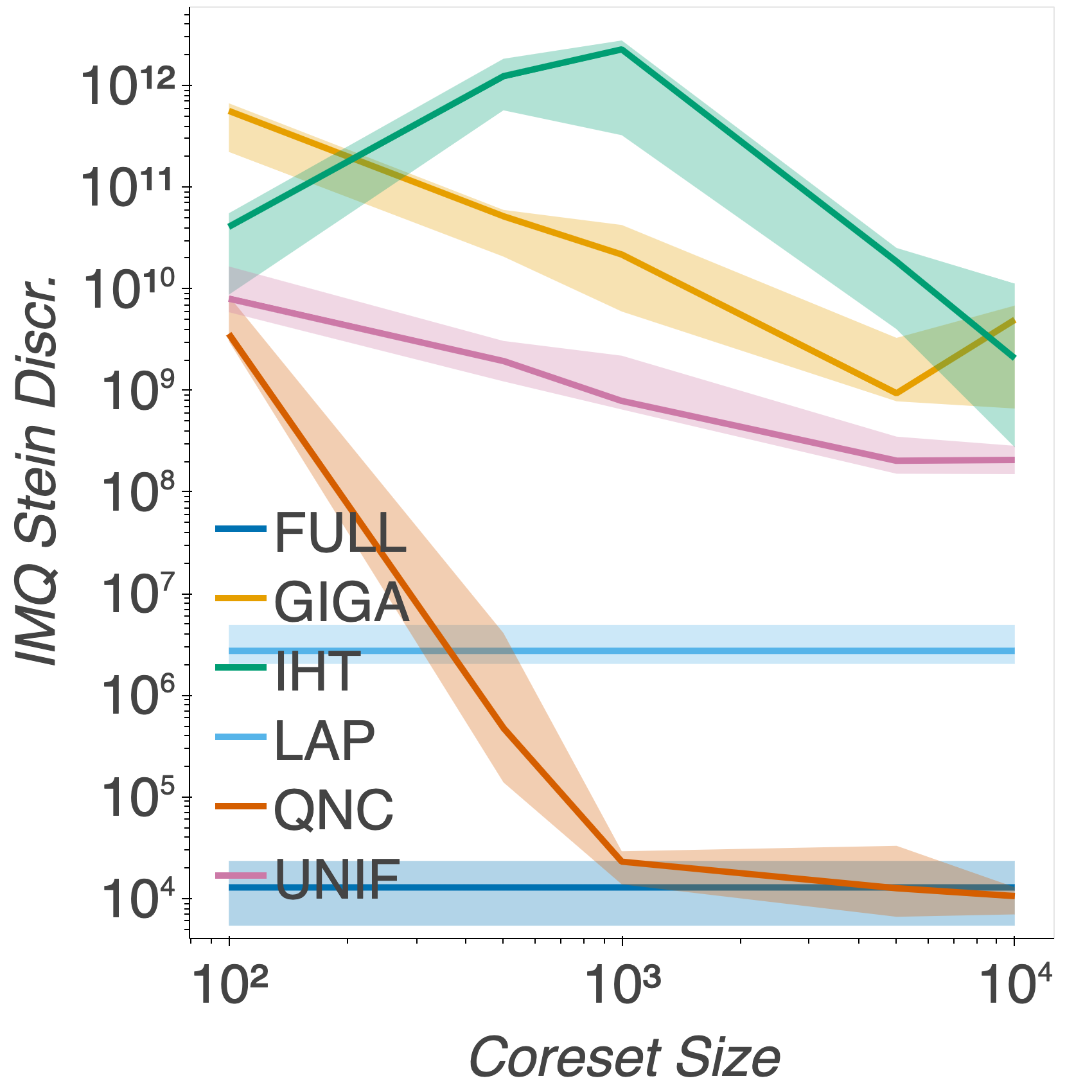}
\end{subfigure}
\begin{subfigure}{0.24\columnwidth}
\includegraphics[width=\columnwidth]{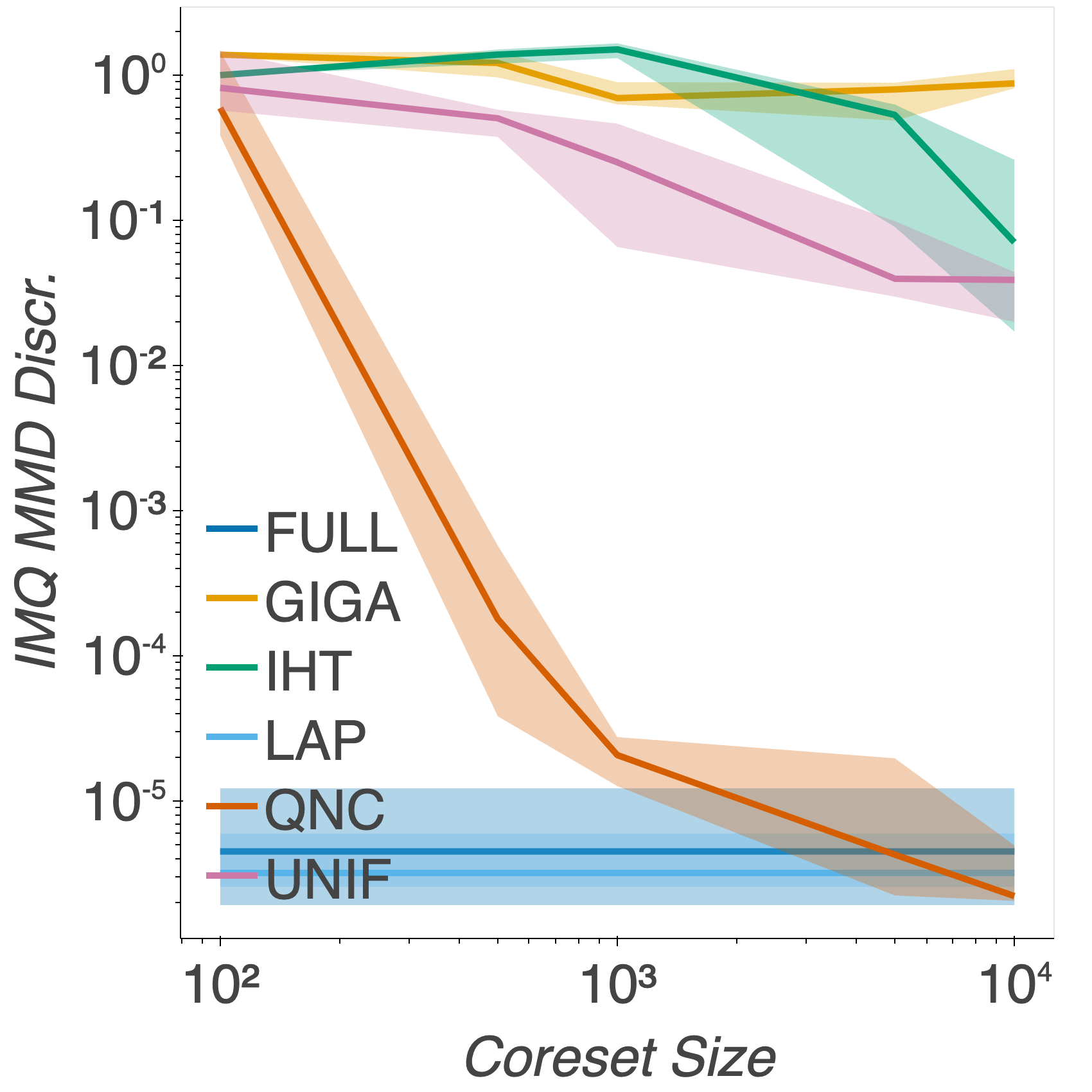}
\end{subfigure}
\begin{subfigure}{0.24\columnwidth}
\includegraphics[width=\columnwidth]{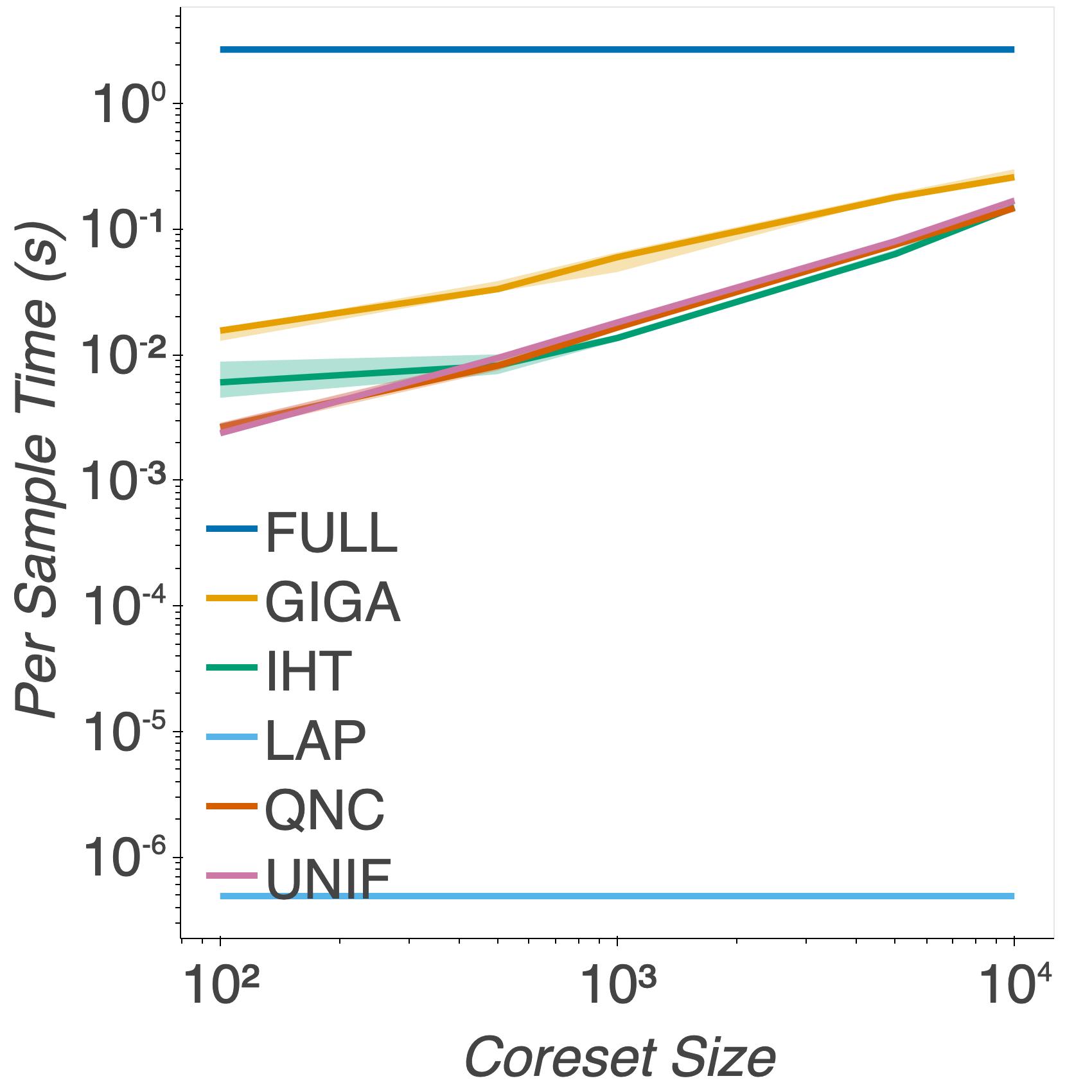}
\end{subfigure}
\caption{Relative mean and log-variance error, forward 
KL divergence, IMQ stein discrepancy,
IMQ maximum mean discrepancy (MMD)
and per sample time to sample from the 
respective posteriors for the logistic regression 
experiment. Our algorithm (QNC) generally provides 
an improvement in coreset quality over the other 
methods. All methods provide a significant
reduction in time to sample from the respective posteriors.}
\label{app:fig:delays_log_stein_kl_build_time}
\end{center}
\vskip -0.2in
\end{figure}
From \cref{app:fig:delays_log_stein_kl_build_time} we 
see that our additional results match closely those 
in \cref{sec:experiments}. Our method (QNC) consistently
outperforms the other methods. As before, we can calculate the number of posterior samples $N_{sample}$ for which the time taken to obtain $N_{sample}$ samples from the full posterior is the same as the time taken to construct a coreset using QNC, and then take $N_{sample}$ samples from the coreset posterior. Here, we find that $N_{sample} \approx 120$ for a coreset of size $1000$. This is again far fewer than you would ideally like to have in practice.

\subsection{Bayesian radial basis function regression}
Our final comparison is on a Bayesian basis function regression example,
and we provide the full details of that experiment here.
We perform inference for the coefficients $\alpha \in \mathbb{R}^{D}$ 
in a linear combination of radial basis functions 
$b_{k}(x)=\exp \left(-1 / 2 \sigma_{k}^{2}\left(x-\mu_{k}\right)^{2}\right), k=1, \ldots, D$,
\begin{align*}
y_{n}&=b_{n}^{T} \alpha+\epsilon_{n},\quad \epsilon_{n} \stackrel{\text { i.i.d. }}{\sim} \mathcal{N}\left(0, \sigma^{2}\right), \quad b_{n}&=\left[\begin{array}{llll}
b_{1}\left(x_{n}\right) & \cdots & b_{D}\left(x_{n}\right)
\end{array}\right]^{T}, \quad \alpha \sim \mathcal{N}\left(\mu_{0}, \sigma_{0}^{2} I\right) .
\end{align*}
The data consists of $N=100,000$ records of house sale 
$\log$-price $y_{n} \in \mathbb{R}$ as a function of latitude / 
longitude coordinates $x_{n} \in \mathbb{R}^{2}$ in the UK.
\footnote{This dataset was constructed by merging 
housing prices from the UK land registry data 
\url{https://www.gov.uk/government/statistical-data-sets/price-paid-data-downloads} 
with latitude \& longitude coordinates from the Geonames postal code data 
\url{http://download.geonames.org/export/zip/}. The housing price dataset 
contains HM Land Registry data \copyright\ Crown copyright and database right 2021. This data is licensed under the Open Government Licence v3.0. The postal code data is licensed under a Creative Commons Attribution 4.0 License.} In order to perform 
inference we generate 50 basis functions for each of 6 scales 
$\sigma_{k} \in\{0.2,0.4,0.8,1.2,1.6,2.0\}$ by generating means 
$\mu_{k}$ uniformly from the data, and including a 
basis with scale 100 (i.e. nearly constant) and mean 
corresponding to the mean latitude and longitude 
of the data. Thus, we have 301 total basis functions, and so $D=301$. The prior and 
noise parameters $\mu_{0}, \sigma_{0}^{2}, \sigma^{2}$ are equal to the 
empirical mean, second moment, and variance of the price paid 
$\left(y_{n}\right)_{n=1}^{N}$ across the whole dataset, respectively.
Closed form expressions are available for the 
subsampled posterior distributions \citep[Appendix B]{Campbell19}, and we 
can sample from them without MCMC.

\begin{figure}[t!]
\begin{center}
\begin{subfigure}{0.24\columnwidth}
\includegraphics[width=\columnwidth]{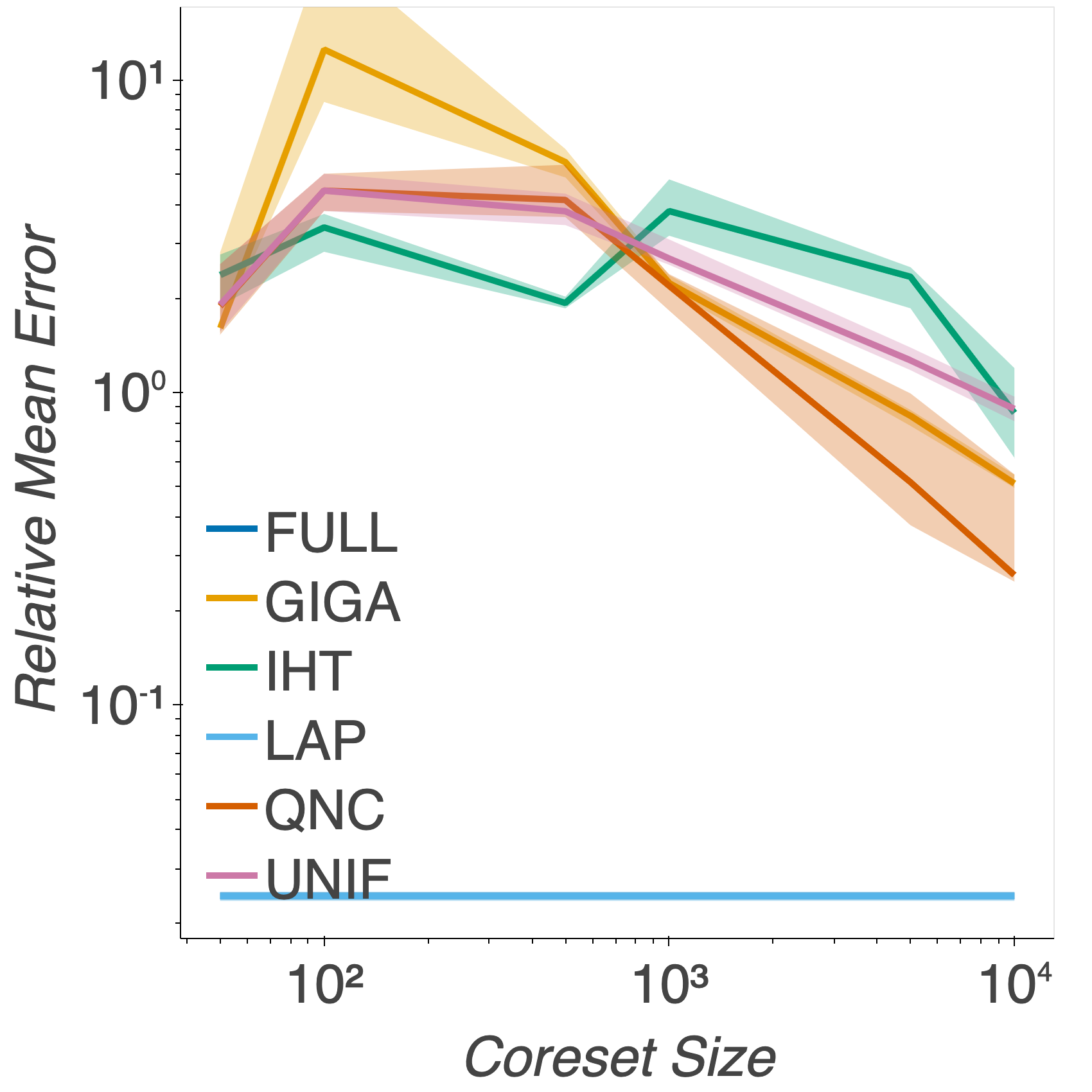}
\end{subfigure}
\begin{subfigure}{0.24\columnwidth}
\includegraphics[width=\columnwidth]{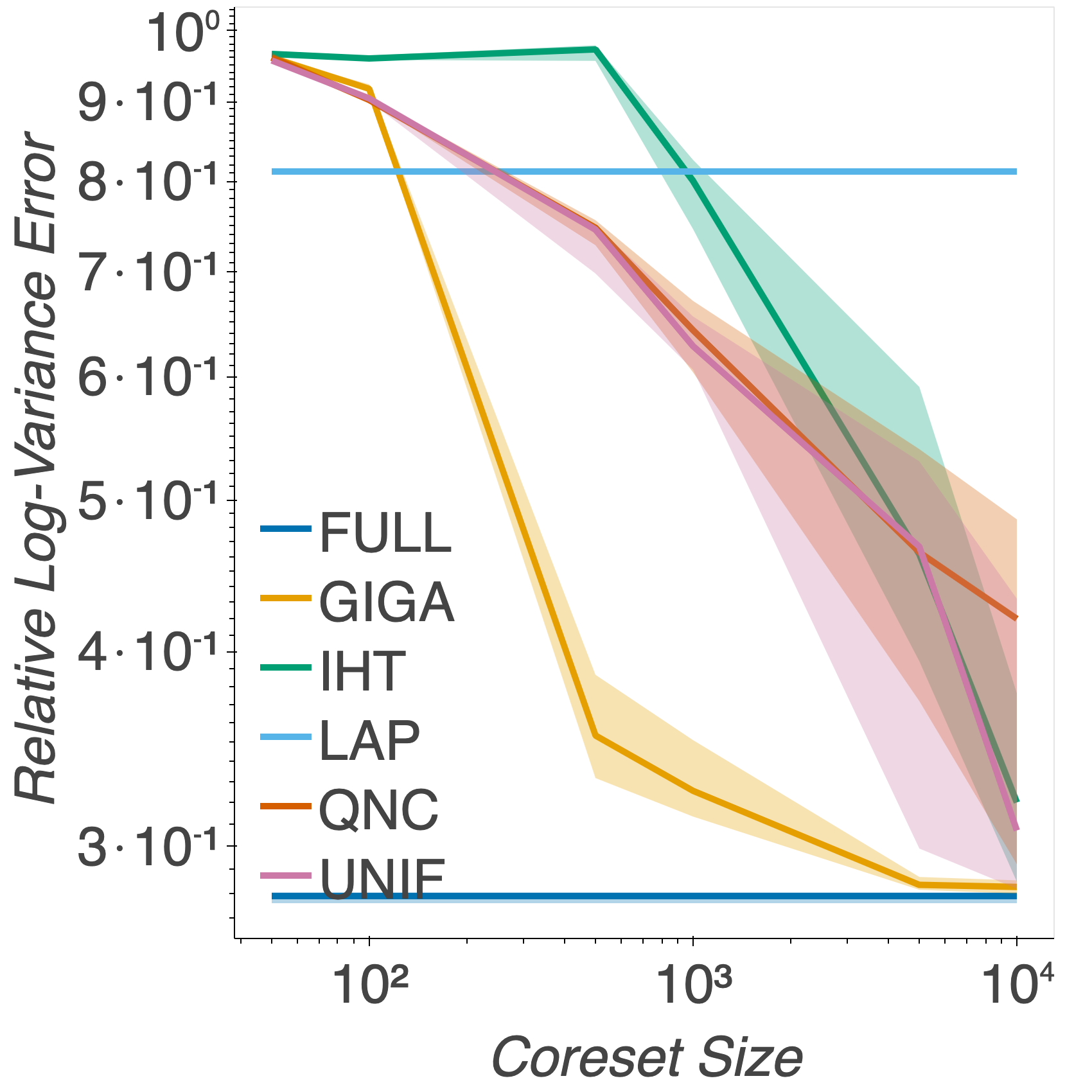}
\end{subfigure}
\begin{subfigure}{0.24\columnwidth}
\includegraphics[width=\columnwidth]{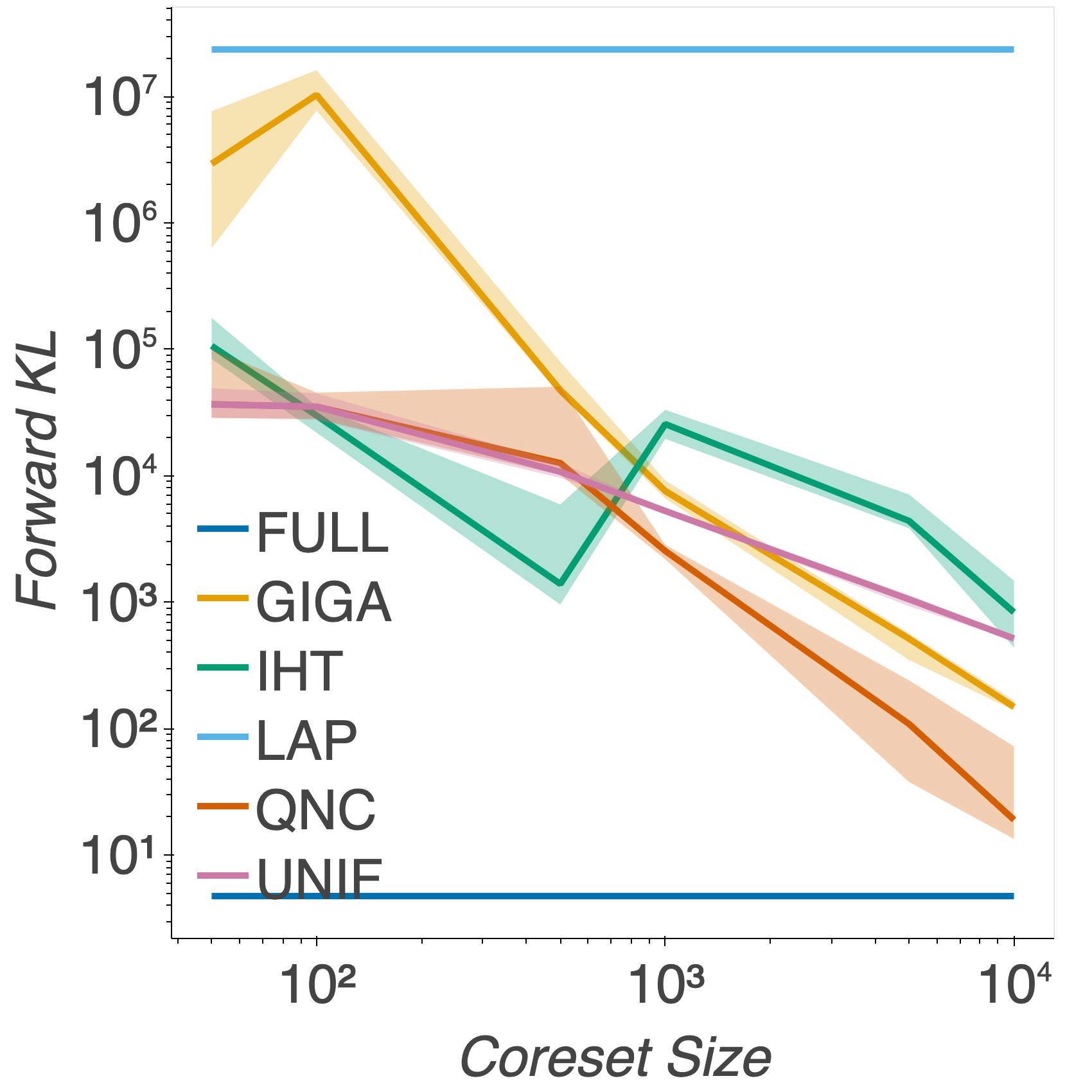}
\end{subfigure}
\begin{subfigure}{0.24\columnwidth}
\includegraphics[width=\columnwidth]{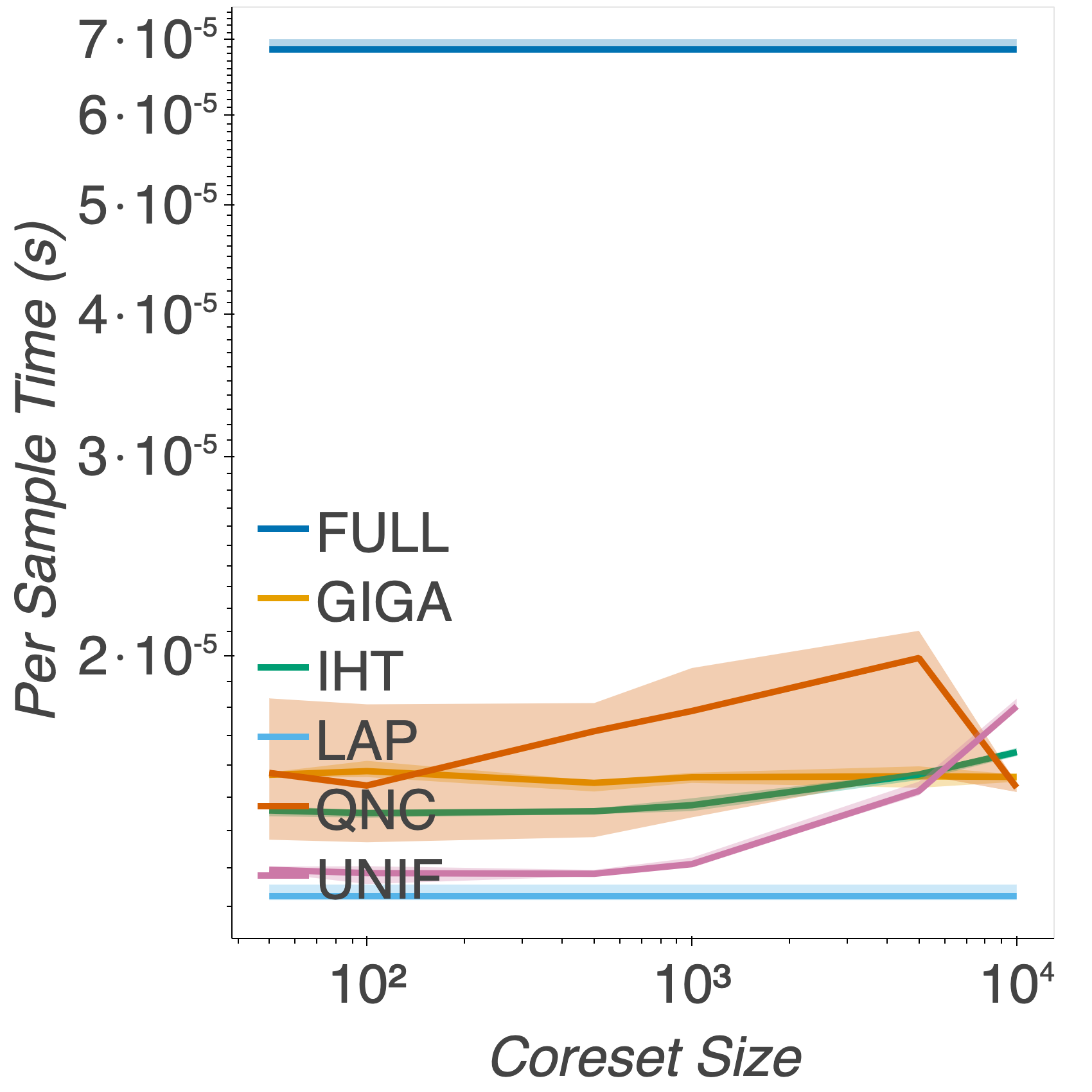}
\end{subfigure}
\caption{Relative mean and log-variance error, forward 
KL divergence and per sample time to sample from the 
respective posteriors for the basis regression 
experiment. Our algorithm (QNC) generally provides 
an improvement in coreset quality over the other 
methods. All methods provide a significant
reduction in time to sample from the respective posteriors.
The poor performance of the Laplace approximation 
in this high-dimensional example can be seen particularly
clearly in the forward KL and log-variance plots.}
\label{app:fig:housing_log_stein_kl_build_time}
\end{center}
\vskip -0.2in
\end{figure}
From \cref{app:fig:housing_log_stein_kl_build_time} we 
see that our additional results match closely those 
in \cref{sec:experiments}. Our method (QNC) generally
outperforms the other methods, but has more trouble
capturing the variance of the posterior, compared
to the other subsampling approaches.
Though the Laplace approximation
captures the mean of the posterior well, it does not 
approximated the variance well, leading to its overall
poor performance.

\subsection{Comparison with Sparse Variational Inference}\label{app:sub:synth_gauss_svi}
\begin{figure}[h]
\begin{center}
\begin{subfigure}{0.45\columnwidth}
\includegraphics[width=\columnwidth]{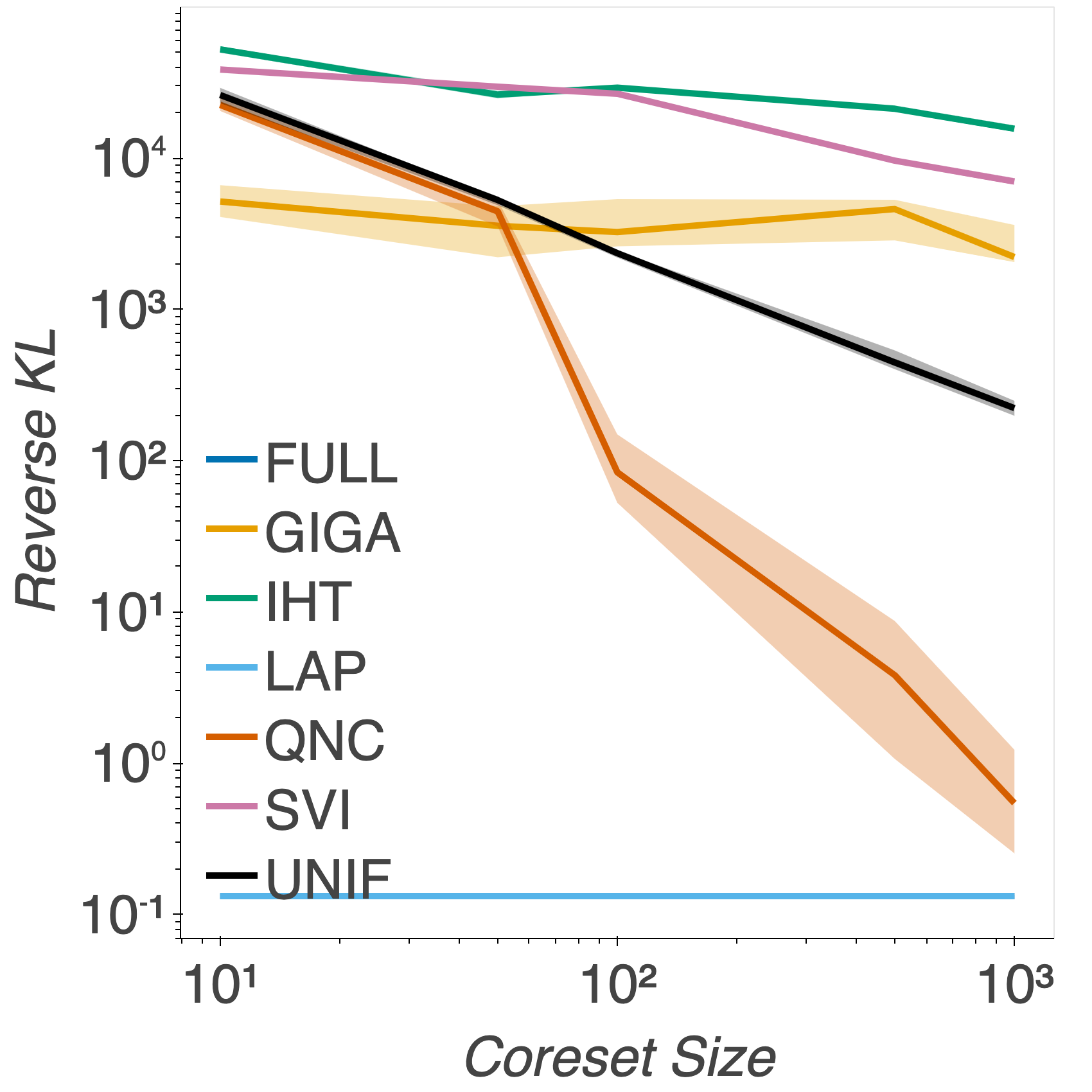}
\end{subfigure}
\begin{subfigure}{0.45\columnwidth}
\includegraphics[width=\columnwidth]{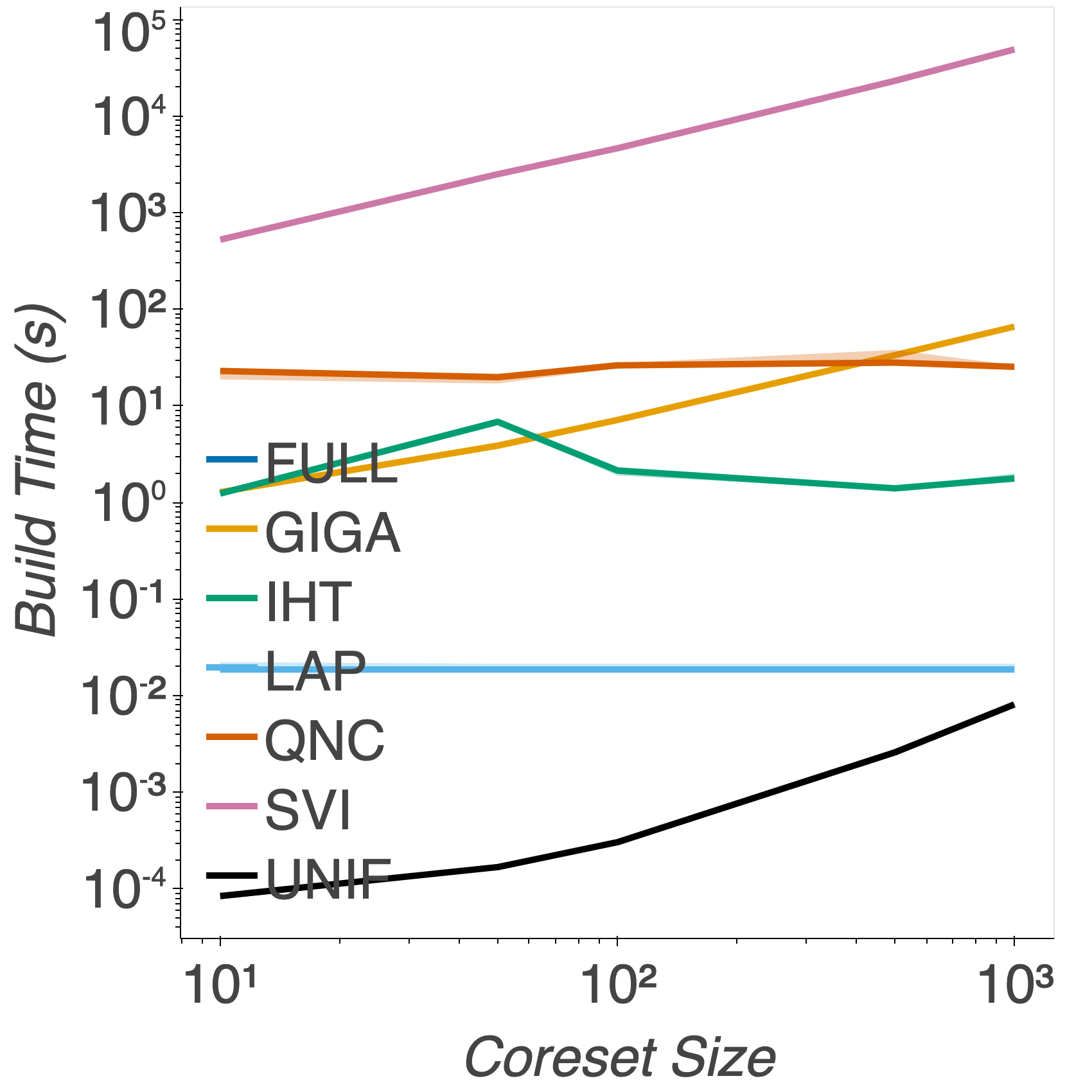}
\end{subfigure}
\caption{ Reverse KL divergence (left) and build time 
in seconds (right) for the small synthetic Gaussian experiment.}
\label{app:synth_gauss_svi}
\end{center}
\vskip -0.2in
\end{figure}

In this section, we provide a comparison of our method against SVI on a smaller dataset than that used in our original experiments. This experiment is the same as that in \cref{sec:sub:synth_gauss}, except that we have reduced the dimension from $100$ to $50$, and the dataset size from $1,000,000$ to $10,000$. For SVI we use $100$ optimization iterations, and only run 1 trial. From \cref{app:synth_gauss_svi}, we see that, even in this smaller data setting, the performance of SVI is not comparable. In reality, the number of optimization iterations needed is much higher than $100$, which is reflected in the poor performance. However, we see that even for this number of optimization iterations the build time can be several orders of magnitude slower than any other method. Thus, we conclude that using SVI is not feasible on the size of datasets that we consider.

\subsection{Sparse regression methods with Laplace approximation}
In \cref{sec:experiments}, we use a uniformly sampled coreset 
approximation of size $M$ as the low-cost approximation $\hat{\pi}$
for GIGA and IHT. Here, we include additional results
with a Laplace approximation used for $\hat{\pi}$. These are labeled
as GIGA-LAP and IHT-LAP respectively.
\begin{figure}[t!]
\begin{center}
\begin{subfigure}{0.48\columnwidth}
\includegraphics[width=0.48\columnwidth]{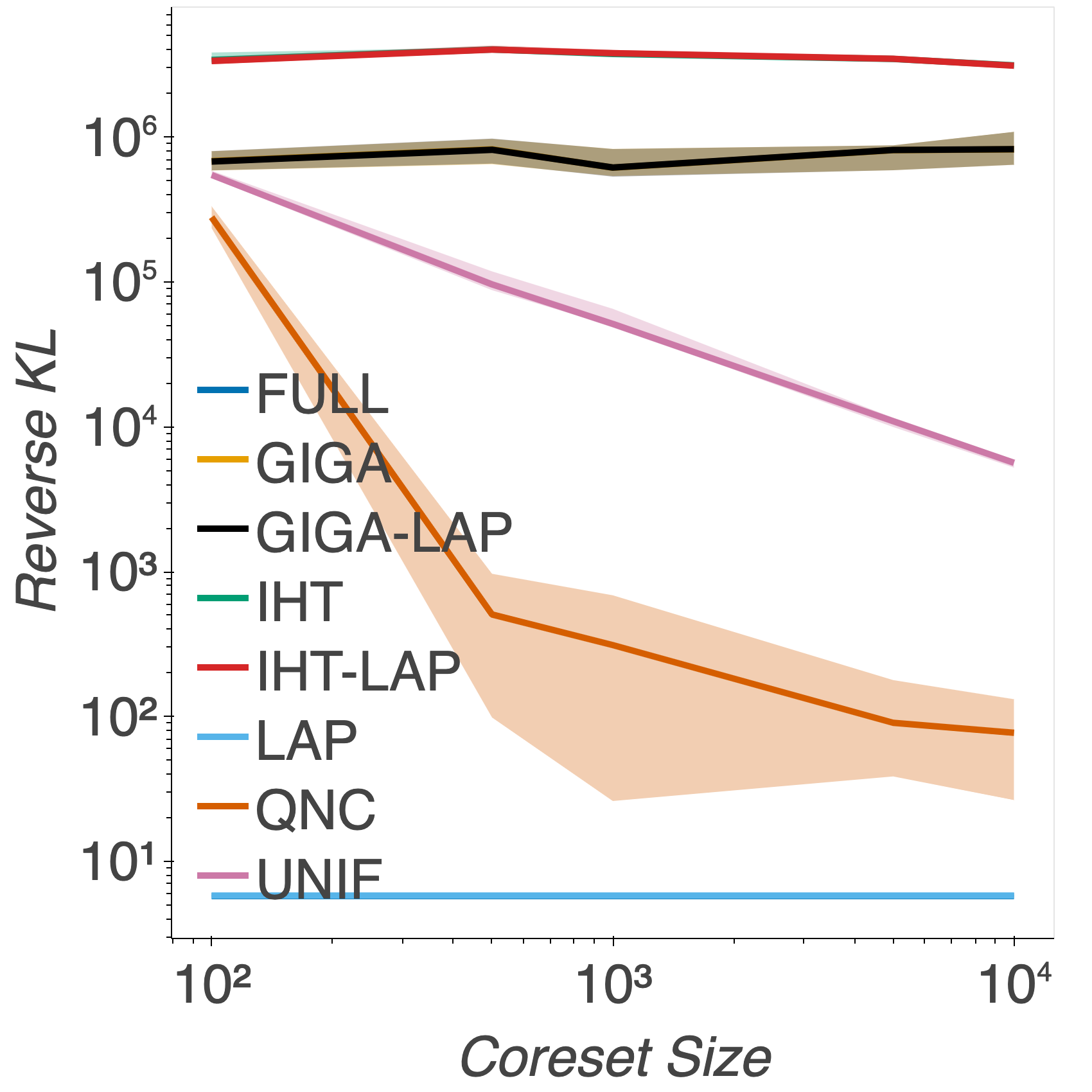}
\includegraphics[width=0.48\columnwidth]{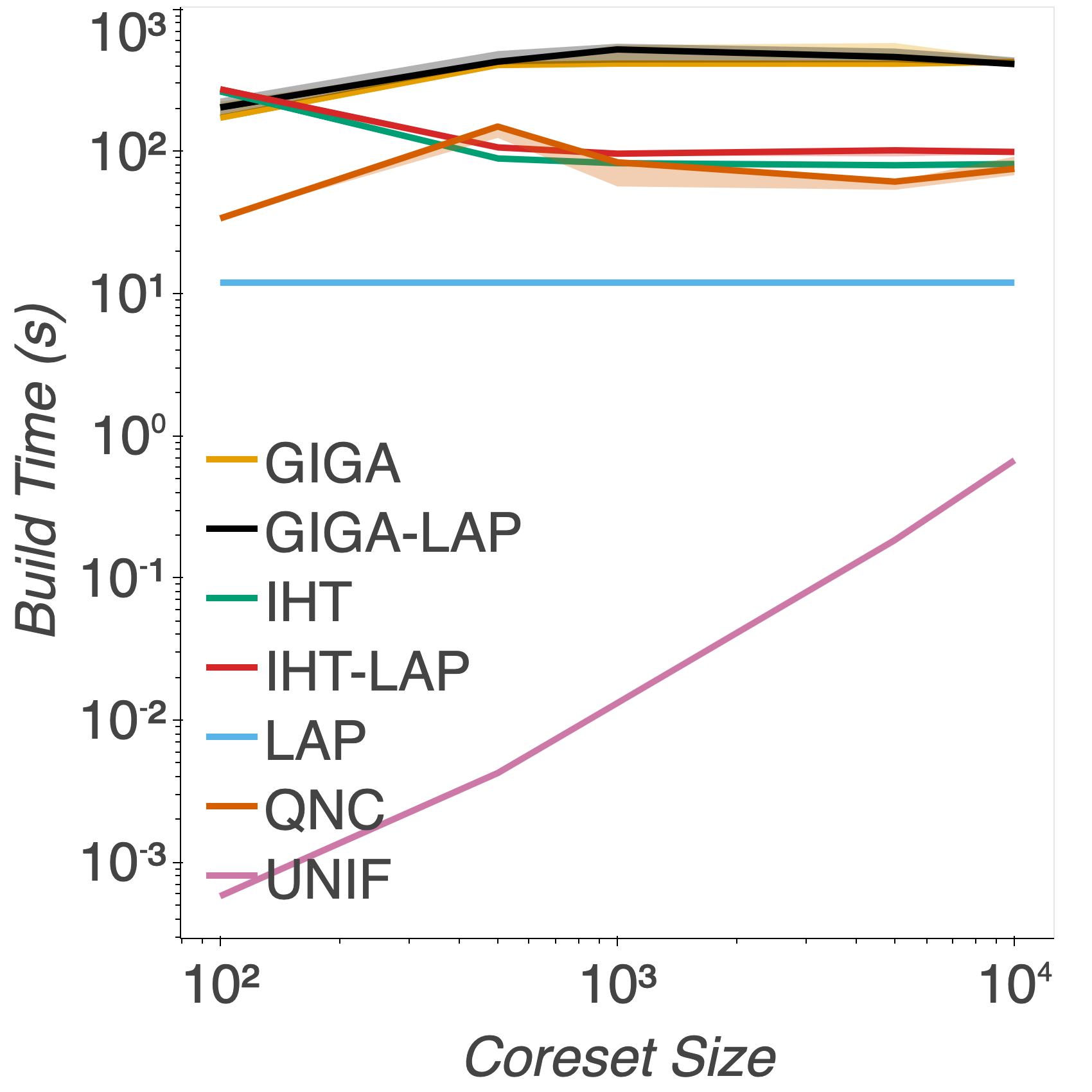}
\caption{Gaussian}
\label{fig:synth_gauss_kl_build_time_app}
\end{subfigure}
\begin{subfigure}{0.48\columnwidth}
\includegraphics[width=0.48\columnwidth]{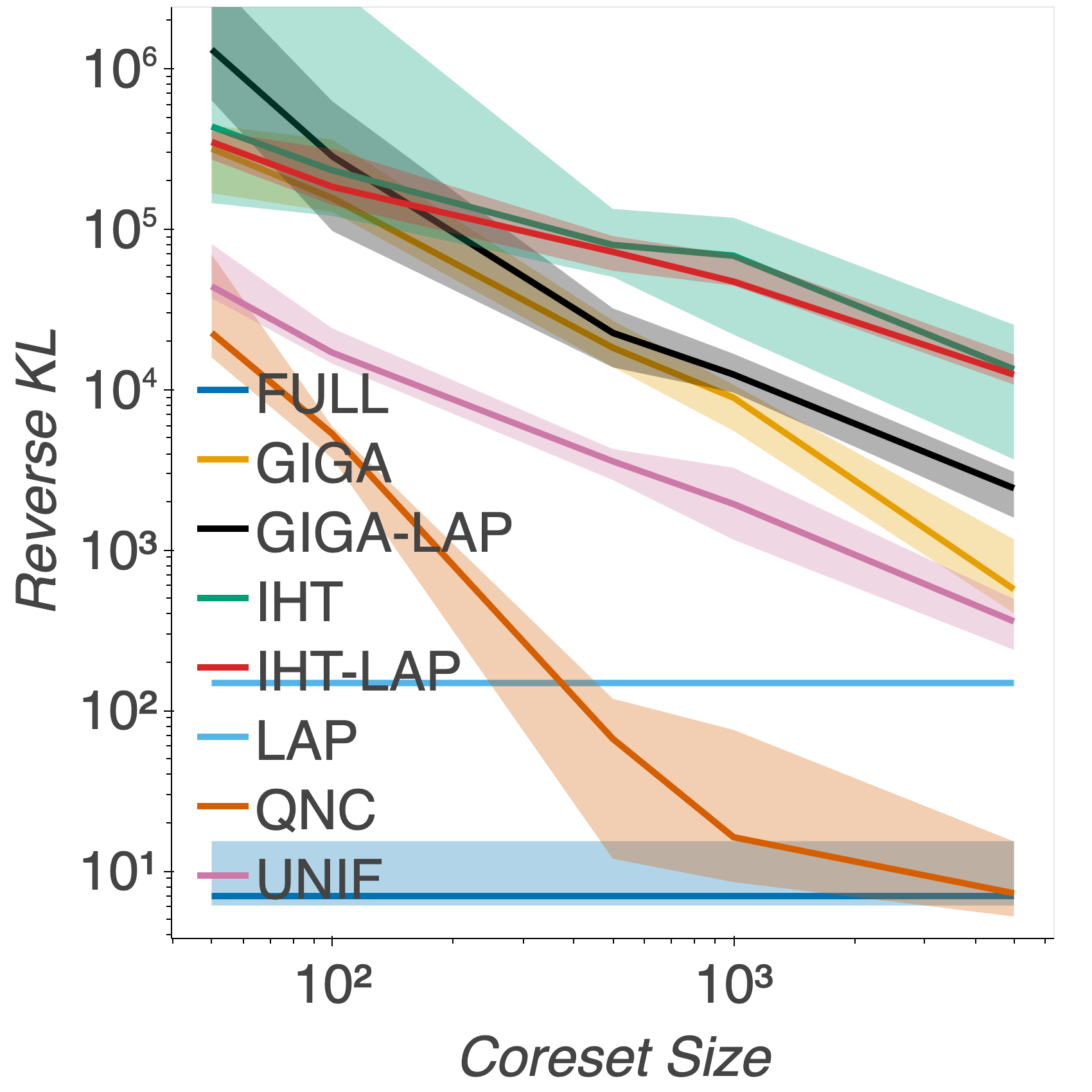}
\includegraphics[width=0.48\columnwidth]{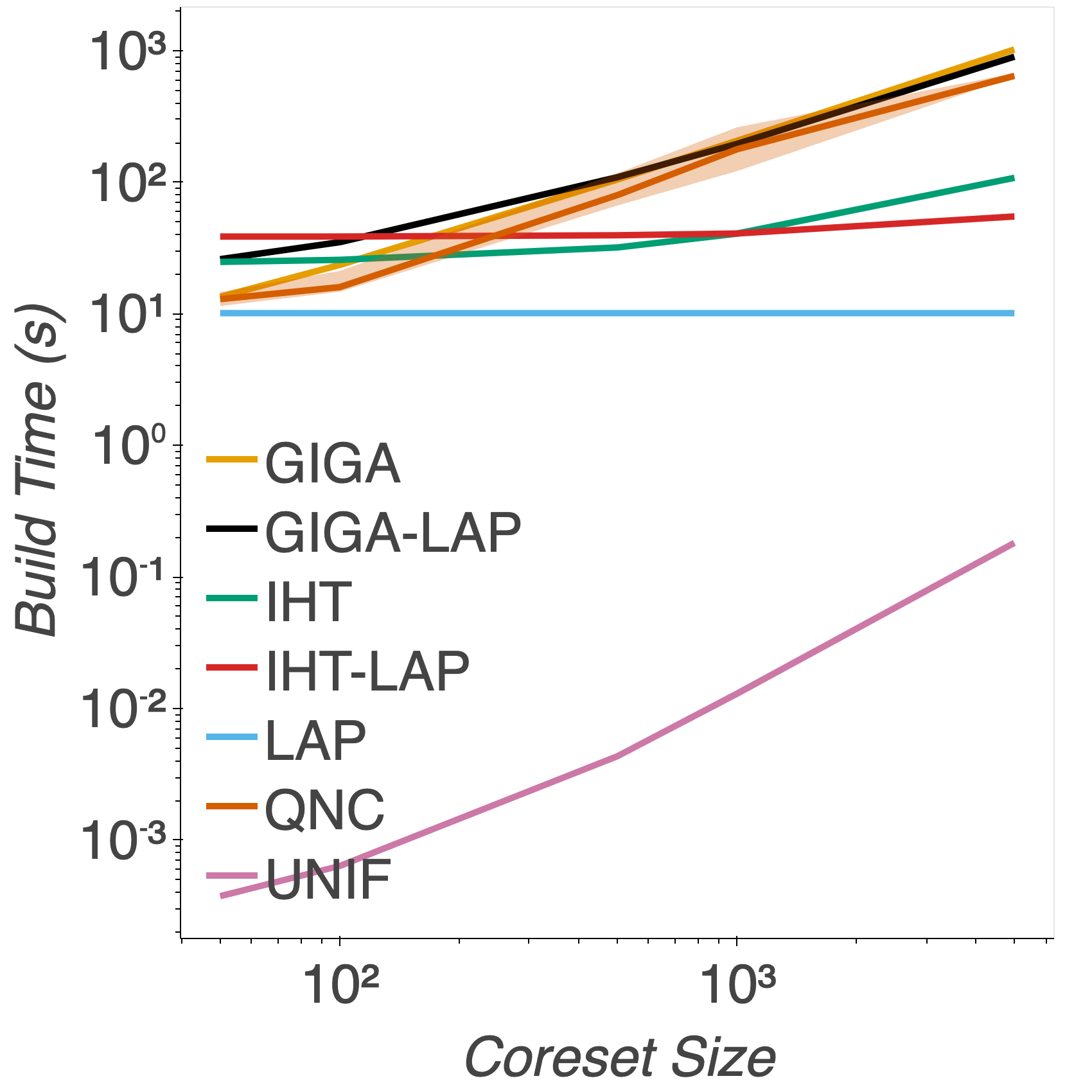}
\caption{Sparse linear regression}
\label{fig:delays_sparsereg_stein_kl_build_time_app}
\end{subfigure}

\begin{subfigure}{0.48\columnwidth}
\includegraphics[width=0.48\columnwidth]{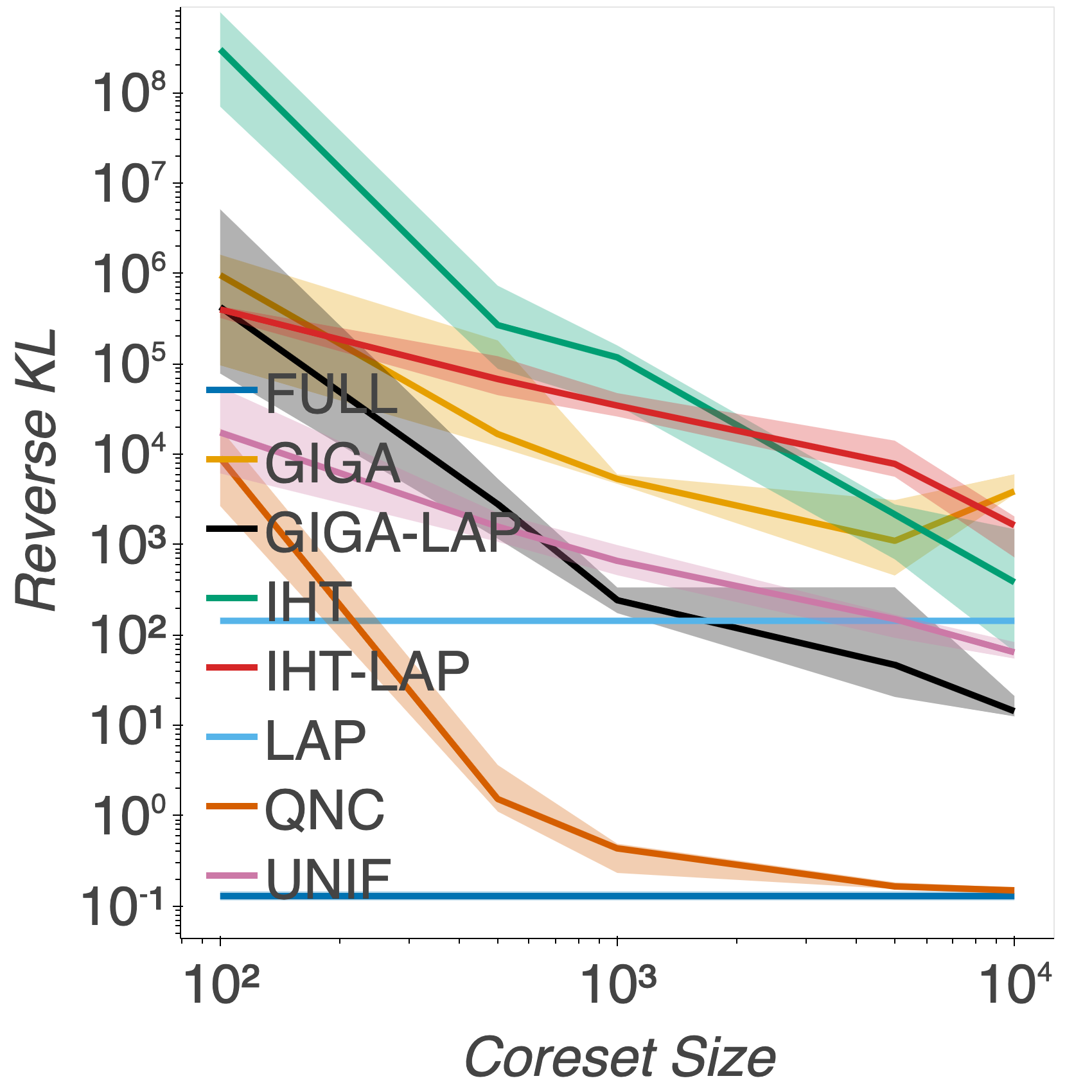}
\includegraphics[width=0.48\columnwidth]{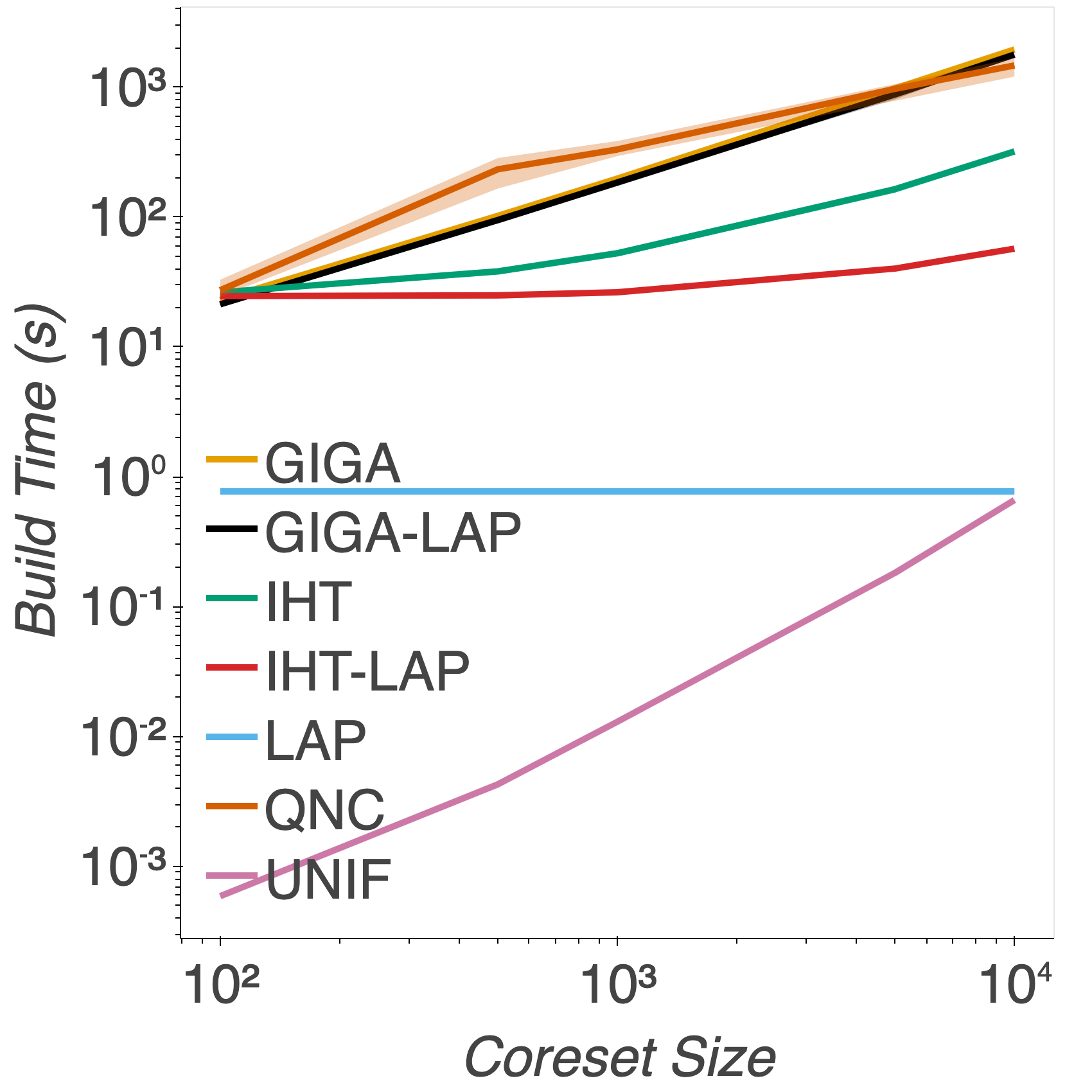}
\caption{Logistic regression}
\label{fig:delays_log_stein_kl_build_time_app}
\end{subfigure}
\begin{subfigure}{0.48\columnwidth}
\includegraphics[width=0.48\columnwidth]{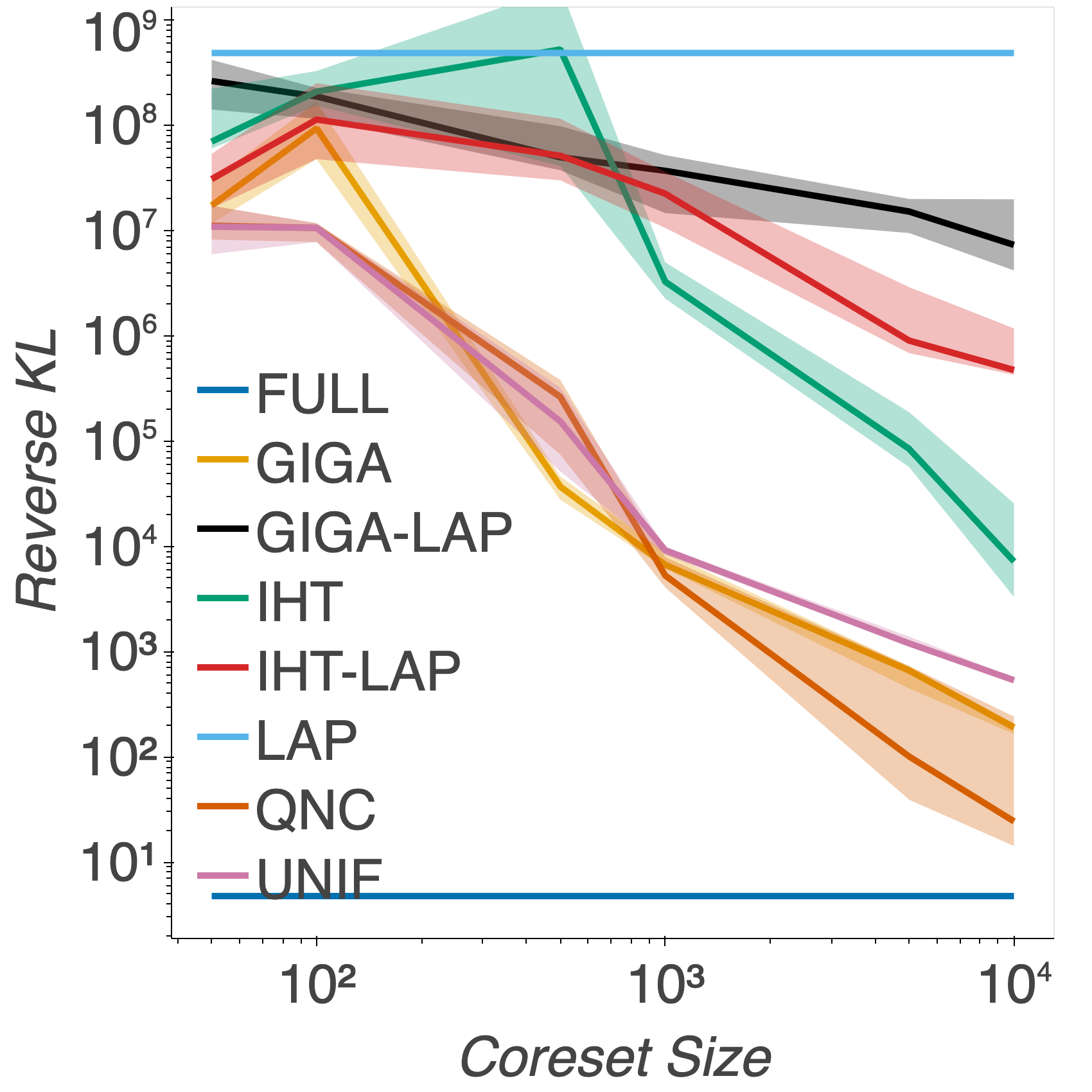}
\includegraphics[width=0.48\columnwidth]{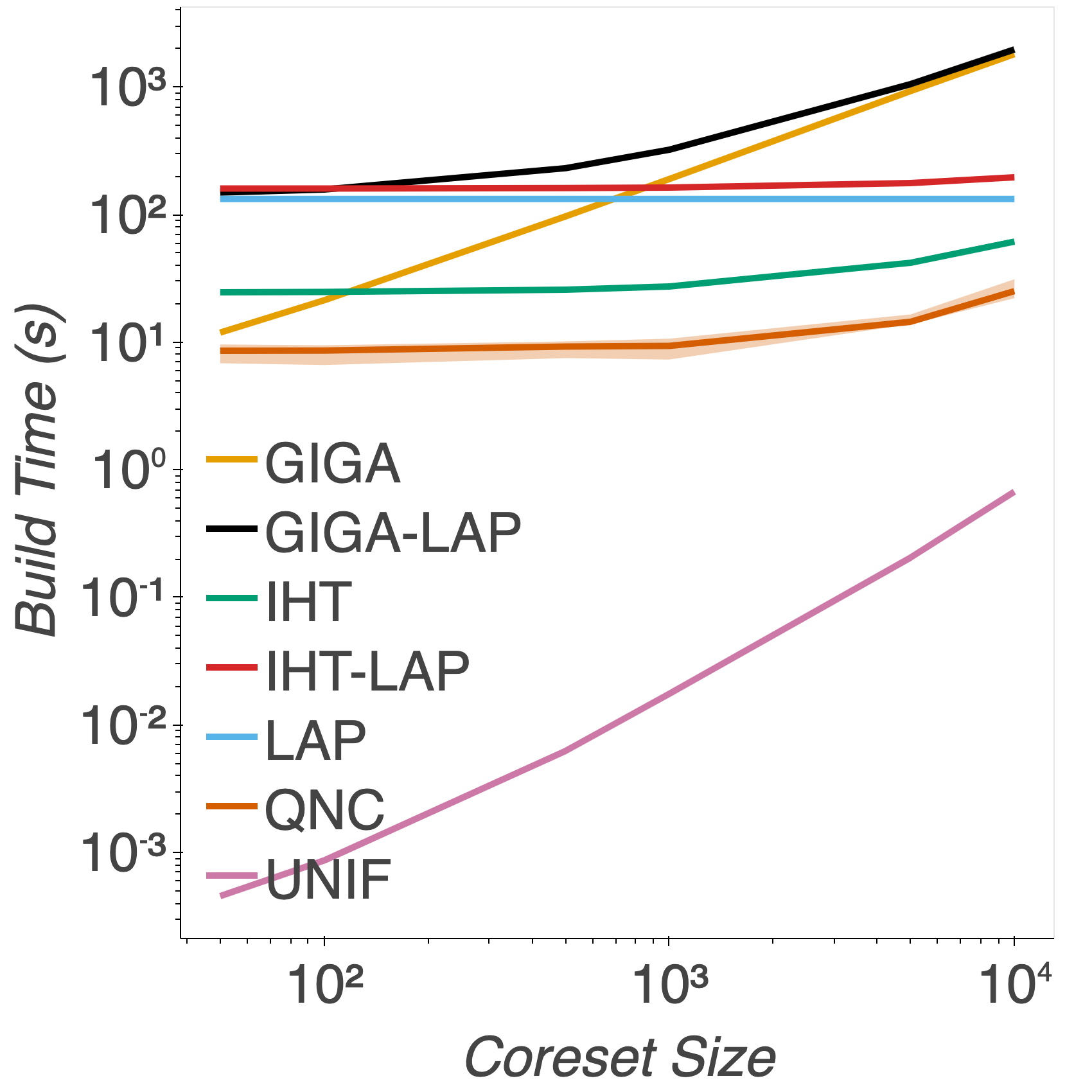}
\caption{Basis function regression}
\label{fig:housing_log_stein_kl_build_time_app}
\end{subfigure}
\caption{Reverse KL divergence (left) and build time 
in seconds (right) for each experiment, with added plots for 
GIGA and IHT with a Laplace approximation used for $\hat{\pi}$. 
We plot the median and a shaded area 
between the 25$^\text{th}$/75$^\text{th}$ percentiles 
over 10 random trials. Using a Laplace approximation used for $\hat{\pi}$ 
rarely provides an improvement in performance, and in some cases
performs significantly worse than with the uniform choice of $\hat{\pi}$.}
\label{fig:lap_kl_build_time}
\end{center}
\vskip -0.2in
\end{figure}
In \cref{fig:lap_kl_build_time}, we see that in the 
synthetic Gaussian experiment, the choice of $\hat{\pi}$ 
has very little effect, and both GIGA and IHT perform
quite badly in this large, high-dimensional example. 
In the sparse regression and basis function regression
experiments, using a Laplace approximation for $\hat{\pi}$
leads to a worse performance. This may be unsurprising
because the Laplace approximation (LAP) does not provide
a good approximation to the true posterior in these settings,
as detailed above and in \cref{sec:experiments}. Only in the 
logistic regression experiment do we see an improvement 
in performance from GIGA-LAP over GIGA.

One reason for this is that the Laplace approximation
provides limited help in coreset construction. Due to the 
large data set sizes used in our 
experiments, we only use a small number 
of samples ($S=500$) in the coreset construction process.
However, especially in high dimensions, we need a large
number of samples to obtain a good $L_2$ approximation
to the posterior from the Laplace approximation.
\subsection{Performance of sparse regression methods}
Throughout our experiments, we see that the sparse regression
methods GIGA and IHT perform poorly. One reason for this
is that these methods involve approximating a high
dimensional integral with a small, fixed, number of samples.
This problem persists no matter which choice of 
low-cost approximation $\hat{\pi}$ we make, and is
particularly problematic in high dimensions \citep{james2013introduction}.
As noted by \citep{Campbell19}, methods like these
are further limited by using a fixed approximation $\hat{\pi}$,
rather than iteratively updating it as in the sparse
variational inference approach.

\subsection{Sensitivity Analysis}\label{app:sub:sensitivity}
\begin{figure}[h]
\begin{center}
\begin{subfigure}{0.6\columnwidth}
\includegraphics[width=0.48\columnwidth]{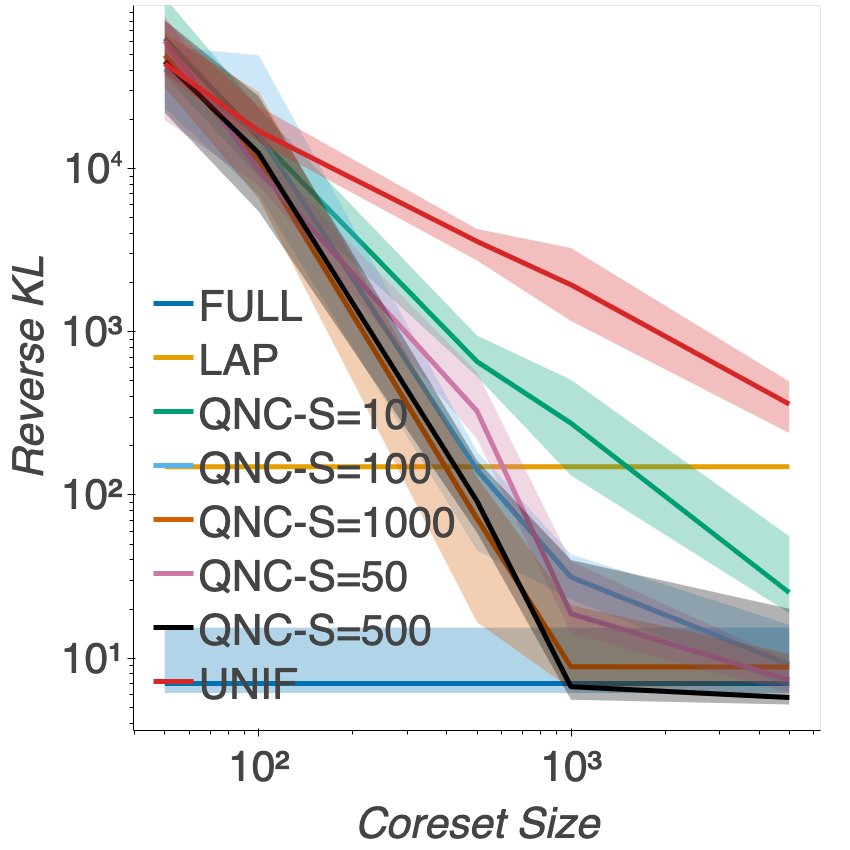}
\includegraphics[width=0.48\columnwidth]{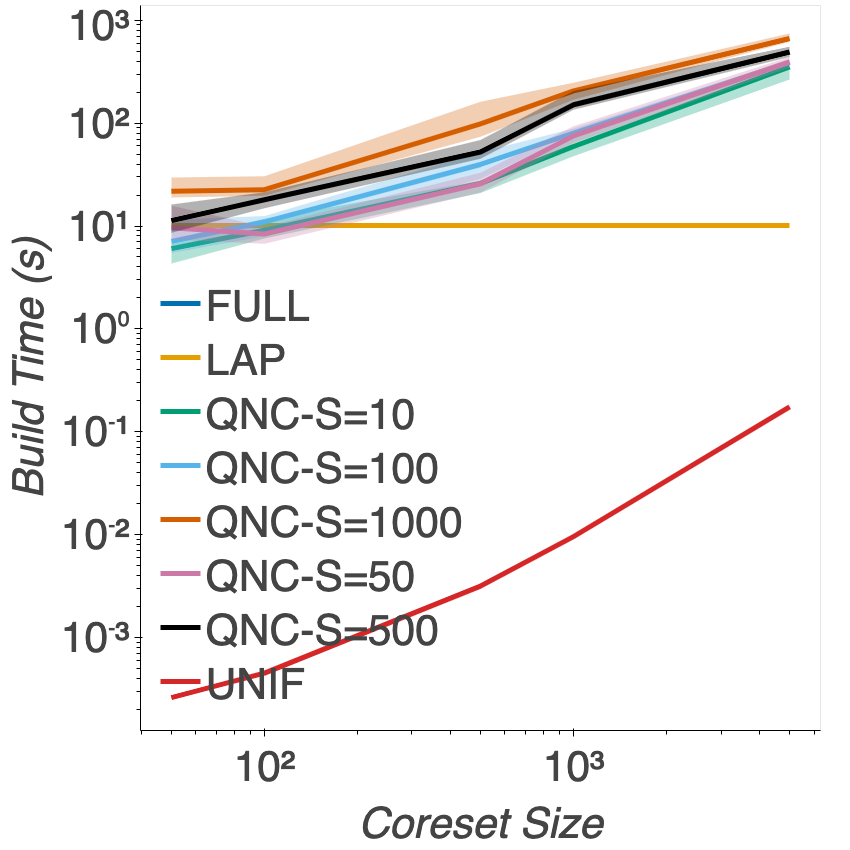}
\caption{S}
\label{fig:synth_gauss_kl_build_time}
\end{subfigure} \\
\begin{subfigure}{0.6\columnwidth}
\includegraphics[width=0.48\columnwidth]{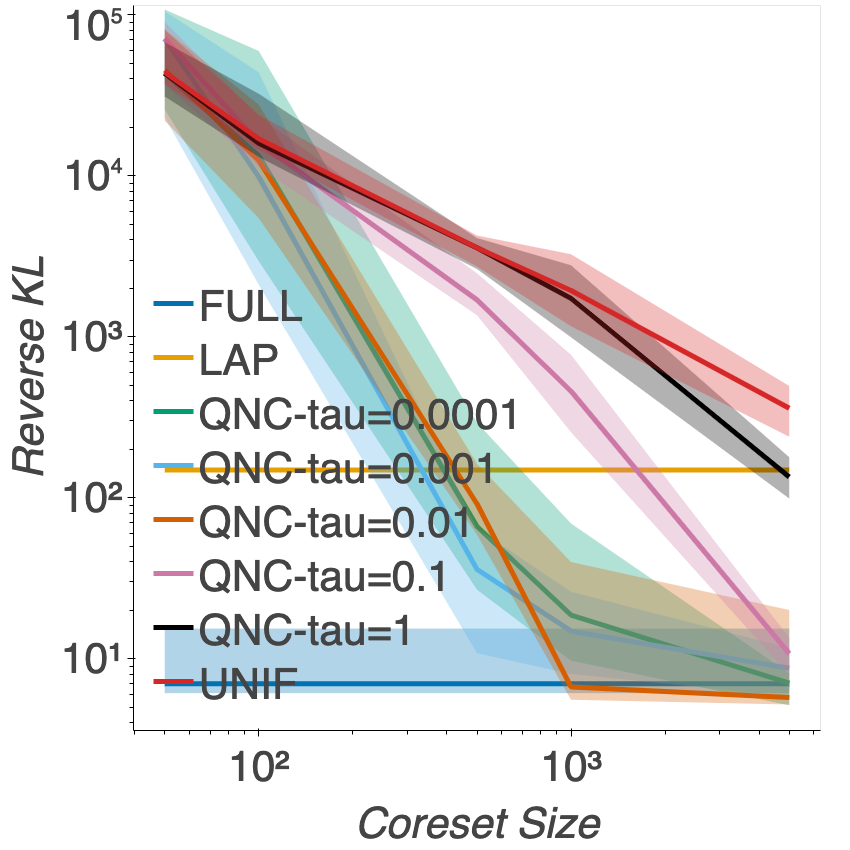}
\includegraphics[width=0.48\columnwidth]{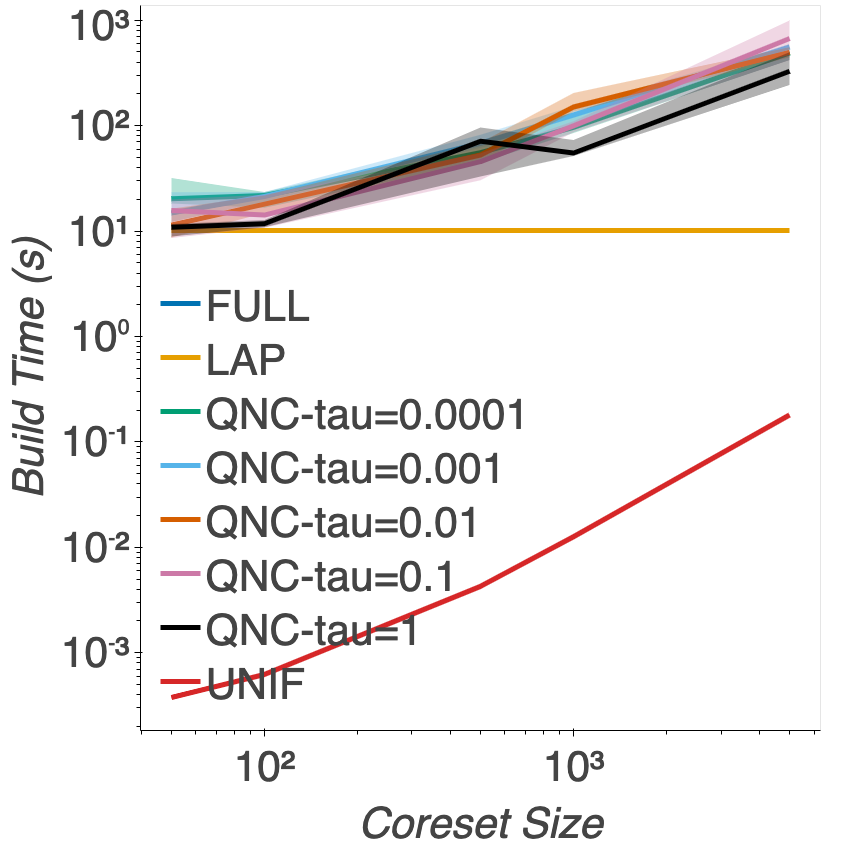}
\caption{$\tau$}
\label{fig:delays_sparsereg_stein_kl_build_time}
\end{subfigure}
\begin{subfigure}{0.6\columnwidth}
\includegraphics[width=0.48\columnwidth]{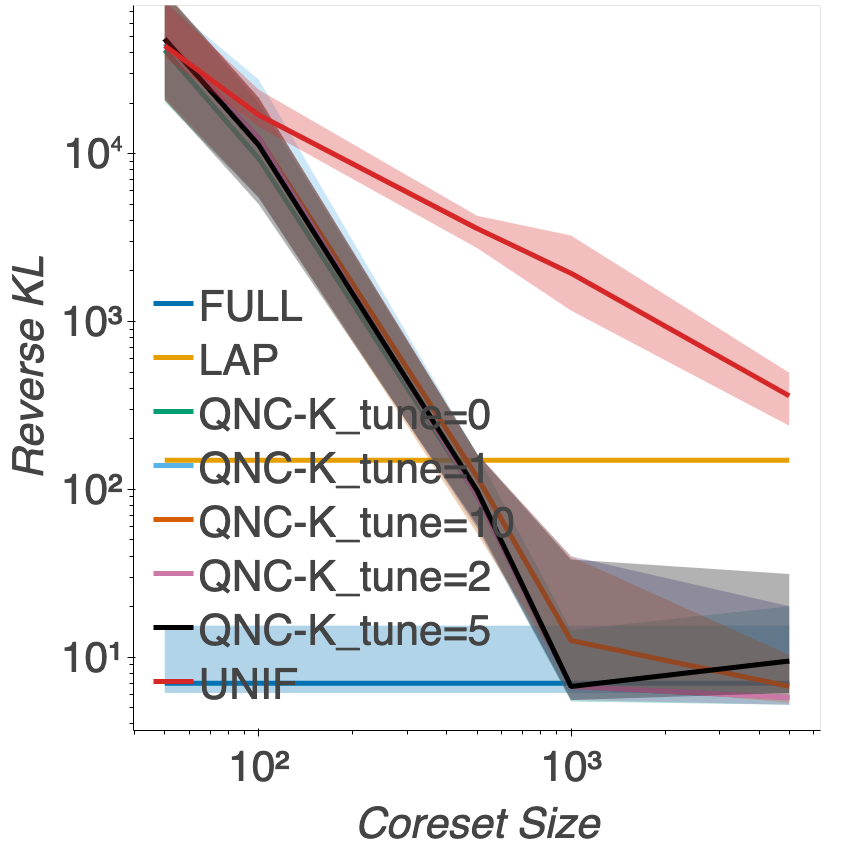}
\includegraphics[width=0.48\columnwidth]{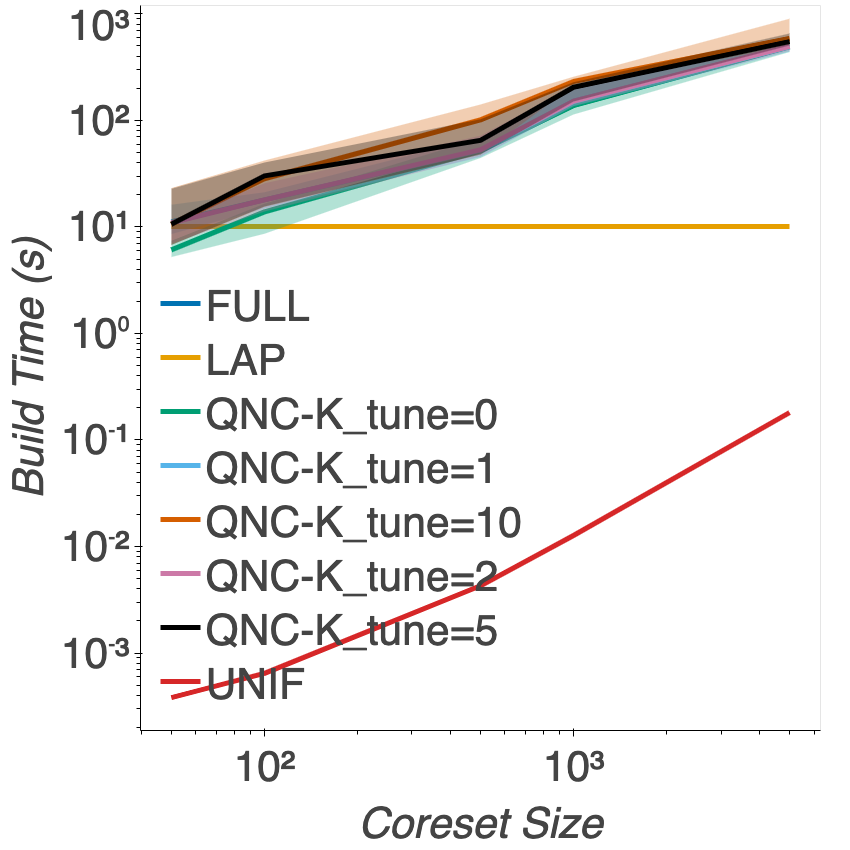}
\caption{$K_{tune}$}
\label{fig:delays_sparsereg_stein_kl_build_time}
\end{subfigure}
\caption{Sensitivity Analyses for the parameters 
$(a)$ $S$, $(b)$ $\tau$ and $(c)$ $K_{tune}$. 
In each case, we plot the Reverse KL divergence (left) and build time 
in seconds (right) for the sparse regression experiment.  }
\label{app:sensitivity}
\end{center}
\vskip -0.2in
\end{figure}

In this section we perform a sensitivity analysis for
the parameters $S$, $K_{tune}$ and $\tau$ that we use 
in \cref{alg:anc_coreset}. 
We do this by repeating the Bayesian sparse linear regression experiment detailed in \cref{sec:sub:sparsereg} for varying values of one of these parameters at a time (with all other parameters kept fixed). From \cref{app:sensitivity} we see the following:

\begin{itemize}
    \item $\textbf{S:}$
        In our original experiments, we use $S=500$ throughout. From this analysis, we see that the performance of our method holds up for substantially lower values of $S$ (though taking $S=10$ is too low and leads to a degradation in performance). Moreover, doubling the number of samples does not lead to significantly better performance. The build time generally increases for larger $S$, as we would expect.
    \item $\mathbf{ \tau :}$
        In our original experiments, we use $\tau = 0.01$ throughout. From this analysis, we see that taking too large a value of $\tau$ negatively affects the performance. This is in line with what we expect, since larger values skew the step away from a true Newton step. Our theory recommends taking $\tau$ as small as possible such that $G(w) + \tau I$ is still (numerically) invertible. Indeed, we see that decreasing $\tau$ can improve the performance of our model, though our choice still provides good results. Thus, we believe ours is a conservative choice that works well in practice. 
    \item $\mathbf{K_{tune}:}$
        In our original experiments, we use $K_{tune} = 1$ throughout. In this analysis, we see that the choice of this parameter has very little effect on the performance of our method for this experiment. In our original experiments, we take $\gamma_k = 1$ for $k > K_{tune}$. When we perform our line search, our starting value is also $\gamma_k = 1$. We find that the line search stops immediately, meaning that for every value of $K_{tune}$ we get essentially the same results.

        The purpose of this step in our algorithm is to guard against the case where our initial gradient and covariance estimates are very noisy, and we may want to take a smaller step initially. However, we see that this is in fact not needed for this experiment.
\end{itemize}

\end{document}